\documentclass[11pt]{article}

\usepackage[preprint]{acl}

\usepackage{graphicx}
\usepackage{amsmath}
\usepackage{multirow}
\usepackage{wrapfig}
\usepackage{xspace}
\usepackage{multirow} 
\usepackage{makecell}
\usepackage{array}
\usepackage{booktabs}
\usepackage{tabularx}
\usepackage{subcaption}
\usepackage{amssymb}
\usepackage{enumitem}

\usepackage{bbding}
\newcommand{\cmark}{\CheckmarkBold}
\newcommand{\xmark}{\XSolidBrush}

\newcommand{\eg}{\textit{e.g.,}\@\xspace}
\newcommand{\ie}{\textit{i.e.,}\@\xspace}

\usepackage{times}
\usepackage{latexsym}

\usepackage[T1]{fontenc}

\usepackage[utf8]{inputenc}

\usepackage{microtype}

\usepackage{inconsolata}

\usepackage{graphicx}

\title{CodeBind: Decoupled Representation Learning for Multimodal Alignment with Unified Compositional Codebook}

\author{
  \textbf{Zeyu Chen}$^{*}$\qquad
  \textbf{Jie Li}$^{*}$\qquad
  \textbf{Kai Han}$^{\dagger}$ \\[0.5em]
  Visual AI Lab, The University of Hong Kong \\
  \tt \small {zeyuc07@connect.hku.hk, jieli\_cn@163.com, kaihanx@hku.hk}
}

\begin{document}
\maketitle
\renewcommand{\thefootnote}{}
\footnotetext{$^{*}$Equal contribution. \quad $^{\dagger}$Corresponding author.}
\renewcommand{\thefootnote}{\arabic{footnote}}
\begin{abstract}
Multimodal representation alignment is pivotal for large language models and robotics. Traditional methods are often hindered by cross-modal information discrepancies and data scarcity, leading to suboptimal alignment spaces that overlook modality-unique features. We propose CodeBind, a framework that optimizes multimodal representation spaces through a modality-shared-specific codebook design.
By incrementally aligning target and bridging modalities, CodeBind bypasses the need for fully paired data.
Unlike traditional hard alignment, CodeBind decomposes features into shared components for semantic consistency and specific components for modality-unique details.
This design utilizes a compositional vector quantization scheme, where a shared codebook bridges modality gaps and modality-specific codebooks mitigate representation bias by preventing dominant modalities from overshadowing others. Validated across nine modalities (text, image, video, audio, depth, thermal, tactile, 3D point cloud, EEG), CodeBind achieves state-of-the-art performance in multimodal classification and retrieval tasks.
Project page: \url{https://visual-ai.github.io/codebind}
\end{abstract}    
\vspace{-0.3cm}
\section{Introduction}
\label{sec:intro}
\vspace{-0.1cm}
Multimodal representation alignment enables Large Language Models (LLMs) to interact with the physical world through diverse sensory inputs~\citep{alayrac2022flamingo, liu2023llava, li2022blip, li2023blip2}. In complex multi-sensor systems like robotic perception, aligning specialized modalities, including audio, depth, thermal, and point clouds, with the established vision-language space is essential for achieving advanced cross-modal intelligence.

Scaling alignment beyond the mature vision-language domain presents two critical challenges that hinder practical deployment. 
First, intrinsic information gaps exist between modalities~\cite{liang2022mind, shi2023Understanding, ramasinghe2024accept}. Compressing heterogeneous data into a shared space often results in a ``least common denominator'' effect. Such suboptimal alignment underperforms on datasets rich in modality-unique features, a frequent bottleneck in robotic manipulation tasks~\cite{tjandrasuwita2025understandalign, jiang2023understanding}.
Second, multimodal data imbalance induces significant representation bias. Unlike web-scale image-text pairs, specialized modalities often have limited data availability, causing the alignment process to bias heavily toward dominant modalities like vision~\citep{wang2024freebind,lyu2024omnibind}. This weakens inter-modal interaction and causes rare modalities to be overshadowed by dominant ones~\citep{lin2025t2dr}. Such bias is particularly detrimental in multi-sensor systems, where the unique contributions of rare sensors must be preserved rather than overwhelmed by visual sensors.

\begin{figure}[t]
\centering
\includegraphics[width=0.9\linewidth]{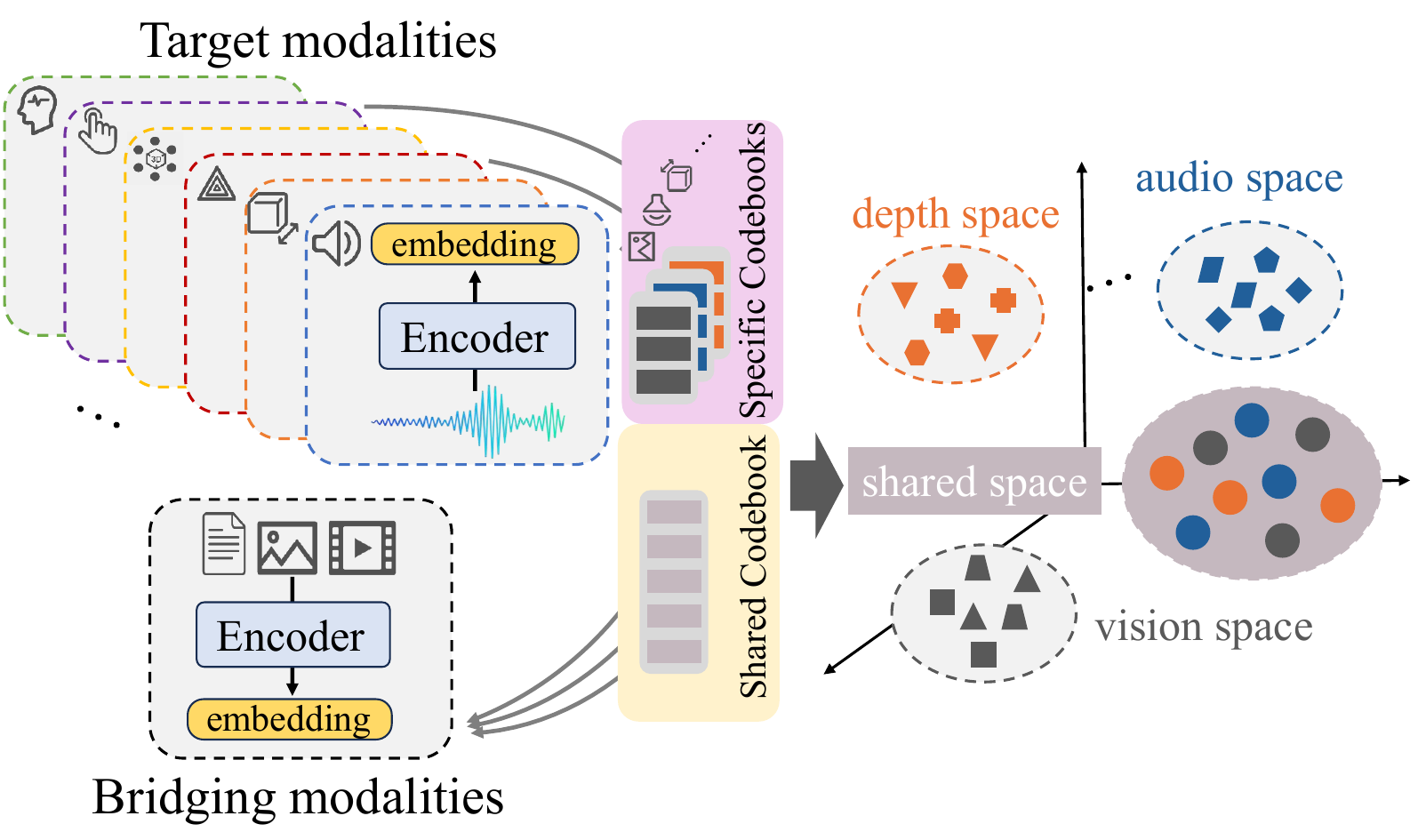}
\vspace{-0.3cm} 
\caption{
\textbf{Multi-modal alignment via codebook}. Target modalities are partially aligned with bridging modalities via codebooks, resulting in a shared space. Unique features from both bridging and target modalities are preserved in specific space.
}
\vspace{-0.7cm}
\label{fig_multimodal_bind_with_anchor}
\end{figure}

Current efforts to enhance multimodal alignment typically rely on training on massive paired datasets, including multimodal data crowdsourcing approaches~\citep {zhu2024languagebind,han2024onellm} or projecting multimodal data into a shared space anchored by vision-language alignment models~\citep{girdhar2023imagebind, lei2024vitlens}.  However, such hard alignment forces entire feature vectors into narrow subspaces and often fails to protect the distinctive richness of individual modalities, leading to degraded performance in multimodal understanding, especially when certain modalities are informatively weak or data-scarce~\citep{jiang2023understanding, tjandrasuwita2025understandalign}.

To address these challenges, we propose \textbf{CodeBind}, a framework that reframes multimodal alignment via discrete vector quantization (VQ). Our core innovation, the modality-shared-specific codebook design, optimizes the multimodal feature space and preserves representational fidelity.

To alleviate suboptimal alignment brought by intrinsic modality gap, we decompose representations into shared components (capturing modality-agnostic invariants) and specific components (preserving modality-unique details). Unlike conventional hard alignment approaches~\citep{radford2021learning,ilharco2021openclip, girdhar2023imagebind, lei2024vitlens, lyu2024unibind,guzhov2022audioclip}, this approach ensures semantic alignment without sacrificing unimodal representational fidelity, empowering intra-modality tasks like fine-grained retrieval.
To mitigate the representation bias, we employ VQ as a distribution-agnostic feature base. By mapping features from different modalities to the same set of discrete semantic centers, VQ provides a structure that prevents dominant modalities from overwhelming the latent space, thereby ensuring feature consistency across unbalanced datasets.

Building on this, our \textit{modality-shared-specific codebook} utilizes a shared codebook for cross-modal consistency and specific codebooks for modality-unique subspaces. By integrating \textit{compositional VQ}, we exponentially expand representational capacity using a compact set of codevectors, ensuring high efficiency and utilization without the computational overhead of traditional large-scale codebooks.

Our contributions are summarized as follows:
\begin{itemize}[leftmargin=*, nosep]
    \item We propose a novel representation decoupling approach that disentangles modality-shared and -specific components, preventing semantic interference and enhancing cross-modal alignment consistency.

    \item We introduce a modality-shared-specific codebook design using compositional vector quantization, resulting in a unified alignment space and maintaining high representational capacity.

    \item We demonstrate the scalability of our approach across nine modalities (text, image, video, audio, depth, thermal, tactile, 3D point cloud, and electroencephalogram (EEG)), achieving state-of-the-art (SOTA) results in cross-modal classification and retrieval while preserving fine-grained details.
\end{itemize}

\vspace{-0.2cm}
\section{Related Work}
\vspace{-0.2cm}

\paragraph{Multimodal alignment} 
Building on vision-language pretraining~\citep{radford2021learning, zhai2023siglip}, multimodal alignment has expanded to audio~\citep{guzhov2022audioclip, wu2022wav2clip}, 3D point clouds~\citep{zhang2022pointclip, dharmasiri2024cross, he2024unim}, and neural signals~\citep{chen2024visual, yang2024neurobind}. 
Current strategies follow three paradigms.
First, unified architectures employ a universal encoder with modality-specific projectors to create a shared space~\citep{zhang2023metatrans, han2024onellm, faye2024oneencoder}. These methods often lack scalability by requiring full-scale retraining when incorporating new modalities. 
Second, bridging methods align various modalities via bridging modalities (text and image)~\citep{girdhar2023imagebind, zhu2024languagebind, lei2024vitlens, lyu2024unibind, fu2024a} or by overlapping pretrained bi-modal models~\citep{wang2023connecting, wang2023extending}. 

Though they show strong scalability~\citep{han2023imagebindllm, su2023pandagpt}, forcing alignment of semantic-scarce modalities to semantic-rich anchors often suppresses modality-unique features, leading to sub-optimal representation. 
Third, Mixture-of-Experts (MoE) strategies dynamically fuse modality encoders through routers~\citep{lyu2024omnibind, zhou2025unialign}. While efficient, they are prone to modality collapse under multimodal data imbalance.
While recent works attempt to mitigate data scarcity by generating pseudo-paired multimodal data~\citep{wang2023extending, wang2024freebind, wang2024omnibind}, introducing synthetic datasets~\citep{zhu2024languagebind}, or designing regularization to maintain performance under limited paired data~\citep{groger2025limitedalign}, these approaches are often labor-intensive and susceptible to semantic noise.
Our approach bypasses these limitations by optimizing the bridging method via partial alignment. We utilize codebook as a universal feature base to accommodate the expression of individual modalities regardless of data imbalance.

\paragraph{Multimodal representation learning}
Recent studies highlight a persistent modality gap in contrastive learning~\citep{liang2022mind, zhang2024C3}, suggesting that pursuing hard alignment often degrades performance by neglecting modality-specific features~\citep{jiang2023understanding, tjandrasuwita2025understandalign}. However, such features remain vital for multimodal retrieval and tasks such as sentiment analysis~\citep{zeng2025ciea, yang2023confede}. To balance multimodal consistency with heterogeneity, existing methods utilize inter/intra-modality regularization or causal components decomposition to disentangle task-related information from redundant noise~\citep{jiang2023understanding, liu2025casualalign, yang2023confede}. Others, like InfoBridge~\citep{li2025infobridge}, modify the InfoNCE loss to provide a protective margin that prevents over-alignment.
However, these strategies are primarily validated on data-rich image-text benchmarks and often focus on modality fusion rather than alignment. Consequently, the efficacy of decoupled features remains under-explored, particularly the alignment robustness of shared components and the informativeness of specific components in data-scarce scenarios.
In contrast, CodeBind decouples embeddings into shared and specific components to achieve robust partial alignment while explicitly safeguarding intra-modal features, validated on diverse and uncommon modalities like tactile and EEG.

\paragraph{Codebook-based unified representation space}
VQ has demonstrated remarkable efficacy in both unimodal and cross-modal representation learning~\citep{van2017vqvae, esser2021vqgan, zheng2022movq, sargent2023vq3d, li2022unimo}. Recent studies have extended joint codebook architectures from vision-language alignment~\citep{zheng2024unicode, duan2022multi} to broader multimodal scenarios~\citep{liu2022cross, xia2024achieving}, with CMG~\citep{xia2024achieving} and FCID~\citep{huang2025fcid} notably employing dual-disentanglement for fine-grained decomposition and alignment across audio, vision, and text.
However, existing codebook designs prioritize unified representation spaces, leaving their effectiveness in modality-specific information preservation unexplored. Furthermore, traditional large-scale codebooks frequently suffer from modality collapse and training inefficiencies~\citep{zheng2024unicode, duan2022multi}. 
To bridge these gaps, we propose a modality-shared-specific codebook design powered by compositional VQ to effectively scale codebook capacity across modalities.

\section{Method}
CodeBind facilitates scalable multimodal alignment without exhaustive pairings by aligning text and vision as bridging modalities with diverse target modalities (Sec.~\ref{sec:method_architecture}). As shown in Fig.~\ref{fig_multimodal_bind_with_anchor}, our framework decouples representations from bridging and target modalities into shared semantic invariants and modality-unique features (Sec.~\ref{sec:method_embedding}), which are then discretized via modality-shared-specific codebook with compositional VQ for efficiency (Sec.~\ref {sec:method_vq}). 

\begin{figure}[t]
\centering
\includegraphics[width=0.9\linewidth]{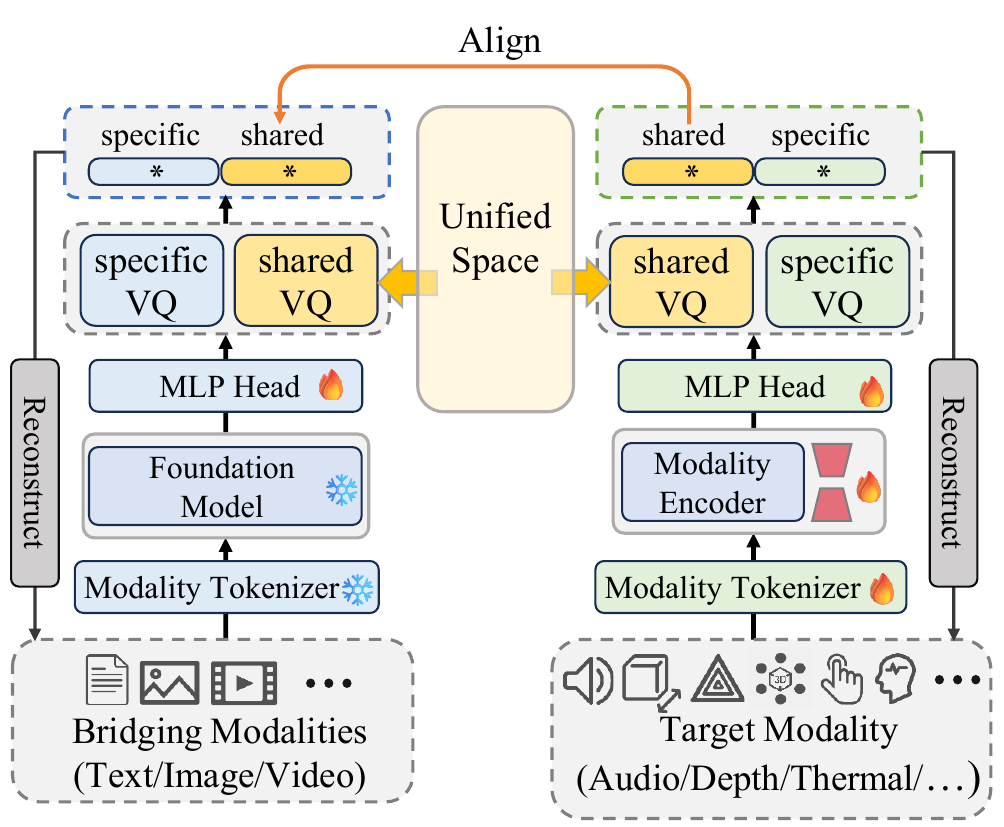}
\vspace{-0.3cm} 
\caption{
\textbf{Alignment across modalities}. Embeddings from bridging and target modalities are decoupled and quantized into shared and specific components, where shared ones are aligned within a unified space.
}
\vspace{-0.4cm} 
\label{fig_pipeline_align2modalities}
\end{figure}

\subsection{Architecture}
\label{sec:method_architecture}
Following ImageBind~\citep{girdhar2023imagebind} and ViT-Lens~\citep{lei2024vitlens}, we align target modalities to a unified space using a frozen vision-language foundation model (\eg OpenCLIP~\citep{radford2021learning,ilharco2021openclip}) as semantic bridge as shown in Fig.~\ref{fig_pipeline_align2modalities}. 
To decouple cross-modal semantics from modality-unique features, encoder outputs are projected into a shared embedding $z^\mathcal{M}_{\text{shared}}$ and a specific embedding $z^\mathcal{M}_{\text{spec}}$, where only the shared part is used for alignment.
To mitigate representation bias, we propose a modality-shared-specific codebook using compositional VQ. 
A universal shared codebook quantizes $z^\mathcal{M}_{\text{shared}}$ while specific codebooks handle modality-specific features. 
Finally, a Transformer decoder reconstructs the original data from the joint embedding $[z^\mathcal{M}_{\text{shared}}, z^\mathcal{M}_{\text{spec}}]$ to ensure information retention. Our plug-and-play design allows target modality encoders to be resumed from bridging methods like ImageBind and finetuned via LoRA~\citep{hu2022lora}.

\subsection{Decoupled Representations}
\label{sec:method_embedding}
Multimodal alignment traditionally maximizes $I(z^{\mathcal{M}_1}, z^{\mathcal{M}_2})$ to synchronize mutual information between shared semantics $y$ and modality embedding: $I(y; z^{\mathcal{M}_1})$ and $I(y; z^{\mathcal{M}_2})$. However, this often leads to over-alignment, where aligning modality-specific noise sacrifices unique semantic richness.
We posit that unimodal feature decoupling is a prerequisite for high-quality cross-modal alignment.
Each modality's embedding is decoupled into two orthogonal components. Shared components ($z^\mathcal{M}_{\text{shared}}$) capture cross-modal semantic invariants (\eg the concept of a ``cat'') to facilitate high-level tasks like cross-modal classification and retrieval. Specific components ($z^\mathcal{M}_{\text{spec}}$) encode modality-unique details (\eg cat color or fur texture, which may be absent in audio or depth). 
Enforcing orthogonality and constraining $z^\mathcal{M}_{\text{spec}}$ to a uniform distribution~\cite{jiang2023understanding} can satisfy $I(y; z^\mathcal{M}) \approx I(y; z^\mathcal{M}_{\text{shared}})$. Consequently, the objective shifts to maximizing $ I(z^{\mathcal{M}_1}_{\text{shared}}; z^{\mathcal{M}_2}_{\text{shared}})$.
This ensures robust semantic alignment while preserving the specific part for fine-grained retrieval and high-fidelity reconstruction.

\begin{figure}[t]
\centering
\includegraphics[width=0.99\linewidth]{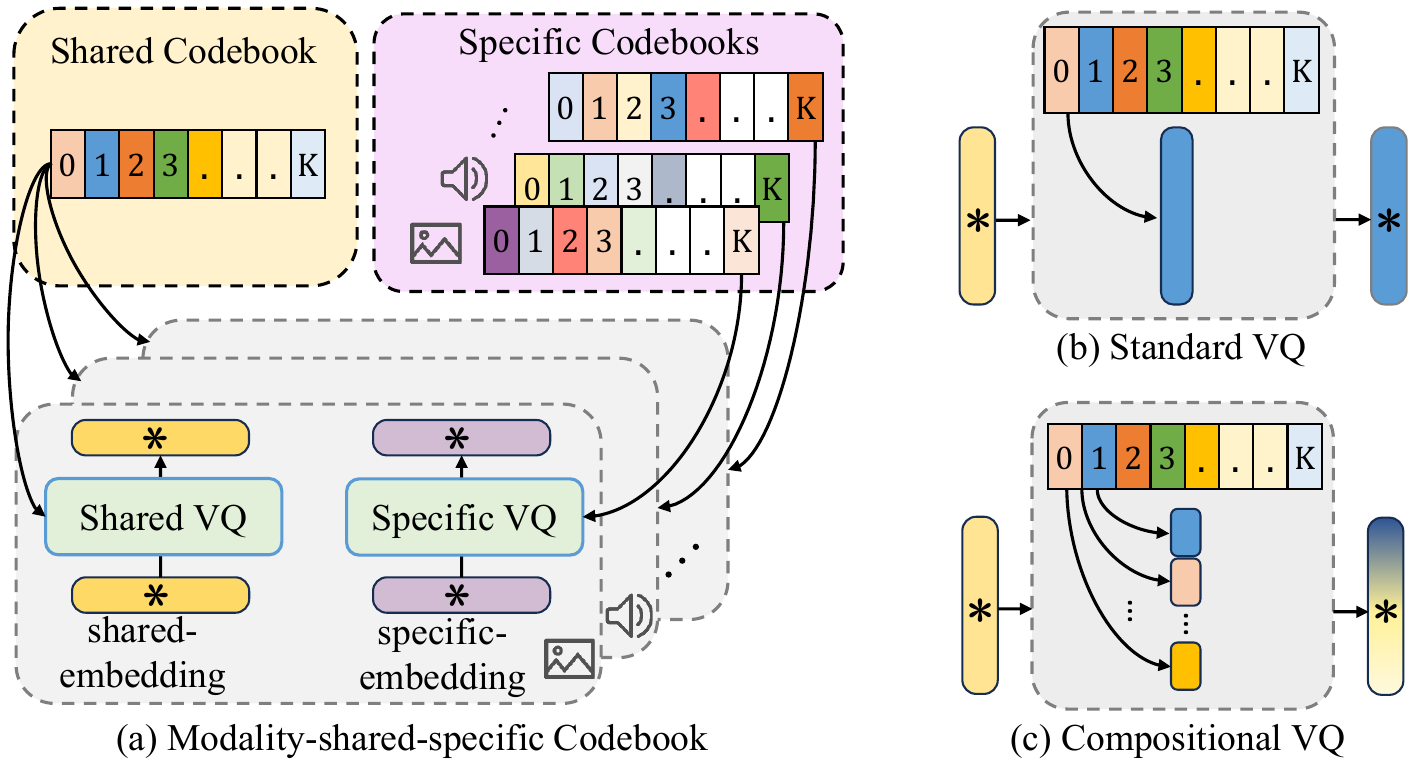}
\vspace{-0.3cm}  
\caption{
\textbf{Modality-shared-specific codebook for multi-modal alignment}. 
(a) The shared embeddings of different modalities use the same codebook for VQ, while the specific embeddings of each modality have their own specific codebooks.
(b) The standard VQ matches each input embedding to a single codevector. (c) Compositional VQ utilizes a combination of multiple low-dimensional codevectors to reconstruct a complete embedding.
}
\vspace{-0.3cm} 
\label{fig:modality-shared-specific_codebook}
\end{figure}

\subsection{Modality-Shared-Specific Codebook}
\label{sec:method_vq}

\paragraph{Preliminary: vector quantization}
In standard VQ, a quantizer
${Q}(\cdot)$ discretizes an input embedding $z\in \mathbb{R}^{D}$ by mapping it to the nearest codevector $c_k \in \mathbb{R}^{D}$ within a codebook $\mathcal{C}$,
\begin{equation*}
Q(z; \mathcal{C}) = \min_{k} \{ \| z - c_k \| \}, \  \mathcal{C} = \{ c_k \}_{k=1}^K
\label{eq:vq}
\end{equation*}
VQ provides two key advantages in multimodal alignment: (1) Mapping heterogeneous embeddings into a discrete, shared codebook space ensures uniform data treatment. This alleviates representation bias and prevents dominant modalities from overwhelming the alignment space. (2) Codevectors serve as learnable feature bases representing core semantic concepts, ensuring the aligned space is grounded in consistent semantic centers.

\paragraph{Compositional codebook}
Existing VQ-based multimodal alignment faces two critical limitations. (1) feature entanglement, where shared and specific information are mixed or inadequately quantized~\citep{zheng2024unicode, xia2024achieving}, leading to noise leakage and distribution bias; and (2) scalability issues, where increasing codebook size to capture rich details triggers high computational overhead and codebook collapse~\citep{zhu2024scaling, zheng2023online}.

To overcome these, we propose a modality-shared-specific codebook design (Fig.~\ref{fig:modality-shared-specific_codebook}) that utilizes a universal codebook $\mathcal{C}_\text{shared}$ to capture the semantic core for alignment and distinct codebooks $\mathcal{C}_\text{spec}^\mathcal{M}$ for modality-specific features. For instance, the concept 'striking' can be quantized in the shared space as the general act of hitting, while specific codebooks capture its modality-dependent meanings: the sharp crack in audio, the high-speed motion in video, or the physical pressure in touch.
To enhance the expressiveness of such features without expanding the codebook size, we implement \textit{compositional VQ}. In this design, a $d$-dimensional embedding is partitioned into $m$ sub-vectors, each of dimension $d^* = d/m$, which are then independently quantized.
This approach exponentially expands the representation space to $K^m$ combinations, forming high-fidelity representations from a compact set of codevectors while maintaining computational efficiency.

\subsection{Implementation}
\label{method:implementation}

\paragraph{Training objective}
The training of CodeBind is guided by a multi-task objective that harmonizes semantic alignment, feature decoupling, and codebook stability. 
We use InfoNCE loss ($\mathcal{L}_{align}$ ) for cross-modal alignment, complemented by orthogonal ($\mathcal{L}_{\text{orth}}$), reconstruction ($\mathcal{L}_{\text{recon}}$), and uniform ($\mathcal{L}_{\text{uni}}$) losses to ensure effective feature disentanglement and representational integrity.
To maintain codebook robustness, we utilize Exponential Moving Average (EMA) updates~\citep{van2017neural} with commitment loss ($\mathcal{L}_{\text{commit}}$), alongside dynamic reinitialization and specialized codevector-level regularizations ($\mathcal{L}_{\text{cctr}}$, $\mathcal{L}_{\text{cuni}}$) to prevent collapse and enhance semantic distinctiveness. Furthermore, a Cross-Modal Code Matching loss ($\mathcal{L}_{\text{cm}}$) is integrated to refine the shared codebook's alignment by synchronizing distance distributions across paired modalities~\citep{liu2022cross}. 
To eliminate manual tuning costs and stabilize multi-objective learning among different modalities, we design an adaptive loss balancing strategy. By maintaining a running estimate of each loss magnitude via  EMA, the weights of other loss functions are dynamically rescaled to match the scale of $\mathcal{L}_{align}$.To further smooth the optimization landscape, update intervals for these weights are increased linearly throughout the training process. See App.~\ref{supp: loss function} for details.

\paragraph{Scalable multi-path alignment}
Our framework employs a multi-path alignment strategy where each target modality (\eg audio) is concurrently aligned with multiple bridging modalities (\eg audio-text and audio-image pairs). To ensure a cohesive unified space, we further enforce internal alignment within the bridging modalities (\eg image-text). Unlike naive pairwise methods, this joint alignment paradigm utilizes distinct codebooks for different target-bridging paths to effectively decouple modality-specific features while maintaining global semantic consistency. This design also enables efficient scaling: new datasets for trained modalities benefit from transfer learning via pre-trained codebooks, while new modalities are integrated by training new codebooks. 
During inference, cross-modal alignment only requires shared embeddings, allowing specific parts to be discarded for a more lightweight pipeline. These components can be enabled to facilitate fine-grained intra-modal tasks. In both scenarios, the reconstruction module is omitted to ensure efficiency.

\begin{table}[tb]
\centering
\scalebox{0.65}
{
\begin{tabular}{c|l|r|c|r} \hline
\textbf{Modality} & \textbf{Dataset} & \textbf{\#Cls} & \textbf{Metric} & \textbf{\#Test} \\ \hline

\multirow{2}{*}{\makecell{\includegraphics[width=10pt]{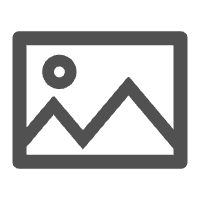} \\[-1ex] \textbf{Image}}}
& IN1K~\cite{russakovsky2015imagenet} & 1000 & Acc & 50000 \\
& P365~\cite{zhou2014learning} & 365 & Acc & 36500 \\ \hline

\multirow{2}{*}{\makecell{\includegraphics[width=10pt]{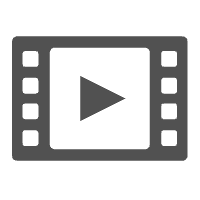} \\[-1ex] \textbf{Video}}}
& K400~\cite{kay2017kinetics} & 400 & Acc & 19759 \\
& MSR-VTT~\cite{xu2016msr} & - & Acc & 2990 \\ \hline

\multirow{2}{*}{\makecell{\includegraphics[width=10pt]{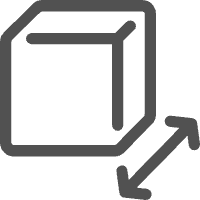} \\[-1ex] \textbf{Depth}}}
& NYU-D~\cite{silberman2012nyu} & 10 & Acc & 654 \\
& SUN-D~\cite{song2015sun} & 19 & Acc & 4660 \\ \hline

\multirow{5}{*}{\makecell{\includegraphics[width=10pt]{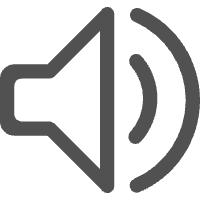} \\[-1ex] \textbf{Audio}}} 
& Audioset~\cite{gemmeke2017audioset} & 527 & Acc & 17132 \\
& ESC~\cite{piczak2015esc} & 50 & Acc & 2000 \\
& Clotho~\cite{drossos2020clotho} & - & Recall & 1046 \\
& AudioCaps~\cite{kim2019audiocaps} & - & Recall & 813 \\
& VGGS~\cite{chen2020vggsound} & 309 & Acc & 15434 \\ \hline

\multirow{2}{*}{\makecell{\includegraphics[width=10pt]{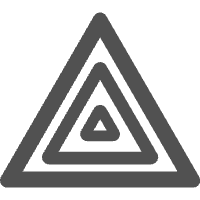} \\[-1ex] \textbf{Thermal}}}
& LLVIP~\cite{jia2021llvip} & 2 & Acc & 16604 \\
& FLIR\_v2~\cite{flirv2kaggle} & 10 & Acc & 1521 \\ \hline

\multirow{3}{*}{\makecell{\includegraphics[width=10pt]{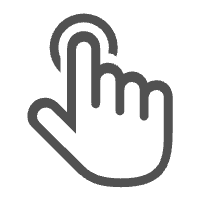} \\[-1ex] \textbf{Tactile}}}
& TAG-M~\cite{yang2022touch_and_go} & 20 & Acc & 29879 \\
& TAG-H/S~\cite{yang2022touch_and_go} & 2 & Acc & 29879 \\
& TAG-R/S~\cite{yang2022touch_and_go} & 2 & Acc & 8085 \\ \hline

\includegraphics[width=12pt]{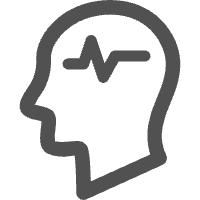} \textbf{EEG} 
& IN-EEG~\cite{spampinato2017eeg} & 40 & Acc & 1997 \\ \hline

\includegraphics[width=12pt]{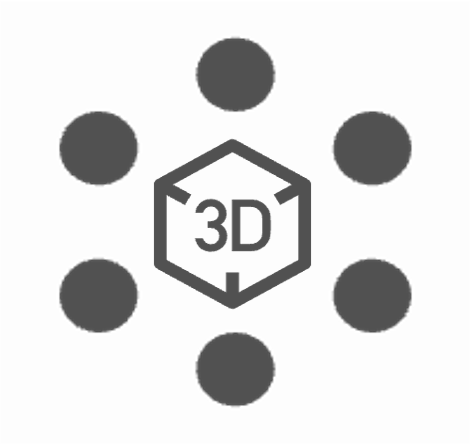} \textbf{3D} 
& ModelNet40~\cite{wu2015shapenets} & 40 & Acc & 2468 \\ \hline

\end{tabular}
}
\vspace{-0.2cm}
\caption{\textbf{Dataset Statistics} across nine aligned modalities.}
\vspace{-0.8cm}
\label{tab_datasets}
\end{table}

\begin{table*}[t]
\centering
\scriptsize

\begin{subtable}{\textwidth}
\centering
\newcolumntype{Y}{>{\centering\arraybackslash}X}

\setlength{\tabcolsep}{2.5pt}
\begin{tabularx}{\textwidth}{l|cc|cc|cc|ccccc|cc|ccc|c} 
\toprule
& \multicolumn{2}{c|}{\raisebox{-1pt}{\includegraphics[width=7.5pt]{fig/icon/image.png} \textbf{Image}}} 
& \multicolumn{2}{c|}{\raisebox{-1pt}{\includegraphics[width=7.5pt]{fig/icon/video.png} \textbf{Video}}} 
& \multicolumn{2}{c|}{\raisebox{-1pt}{\includegraphics[width=7.5pt]{fig/icon/depth.png} \textbf{Depth}}} 
& \multicolumn{5}{c|}{\raisebox{-1pt}{\includegraphics[width=7.5pt]{fig/icon/audio.png} \textbf{Audio}}} 
& \multicolumn{2}{c|}{\raisebox{-1pt}{\includegraphics[width=7.5pt]{fig/icon/thermal.png} \textbf{Thermal}}} 
& \multicolumn{3}{c|}{\raisebox{-1pt}{\includegraphics[width=7.5pt]{fig/icon/tactile.png} \textbf{Tactile}}} 
& \multicolumn{1}{c}{\raisebox{-1pt}{\includegraphics[width=7.5pt]{fig/icon/eeg.png} \textbf{EEG}}} \\ 
\midrule
& \tiny IN1K & \tiny P365 & \tiny K400 & \tiny MSR-VTT & \tiny NYU-D & \tiny SUN-D & \tiny Audioset & \tiny VGGS & \tiny ESC & \tiny Clotho & \tiny AudioCaps & \tiny LLVIP & \tiny FLIR\_v2 & \tiny TAG-M & \tiny TAG-H/S & \tiny TAG-R/S & \tiny IN-EEG \\ \midrule
ImageBind 
& 77.7 & 45.4 & 50.5 & 36.1 & 54.0 & 35.1 
& 17.6 & 27.8 & 66.9 & 6.0/28.4 & 9.3/42.3
& 63.4 & 46.6 & 24.2 & 65.7 & 69.8 & 18.4 \\ \midrule
CodeBind 
& \textbf{79.3} & \textbf{55.5} & \textbf{54.4} & \textbf{37.8} & \textbf{59.3} & \textbf{45.7} 
& \textbf{21.1} & \textbf{30.5} & \textbf{71.0} & \textbf{6.9/28.6} & \textbf{13.3/53.8}
& \textbf{95.5} & \textbf{97.2} & \textbf{42.6} & \textbf{83.9} & \textbf{78.2} & \textbf{33.1}\\ \bottomrule
\end{tabularx}

\caption{
CodeBind-IB integrates our method into ImageBind. For retrieval, Recall@1 is reported on MSR-VTT and ESC, and Recall@1/Recall@10 are reported on Clotho and AudioCaps.
For classification, Acc@1 is reported on all other datasets, except for AudioSet, where mAP is reported.
} 
\end{subtable}

\vspace{0.1cm}

\begin{subtable}{\textwidth}
\centering
\scriptsize
\setlength{\tabcolsep}{3.8pt} 
\newcolumntype{Y}{>{\centering\arraybackslash}X}

\begin{tabularx}{\textwidth}{l|cc|ccccc|ccc|c|Y} \toprule
    & \multicolumn{2}{c|}{\raisebox{-1pt}{\includegraphics[width=7.5pt]{fig/icon/depth.png} \textbf{Depth}}} 
    & \multicolumn{5}{c|}{\raisebox{-1pt}{\includegraphics[width=7.5pt]{fig/icon/audio.png} \textbf{Audio}}}  
    & \multicolumn{3}{c|}{\raisebox{-1pt}{\includegraphics[width=7.5pt]{fig/icon/tactile.png} \textbf{Tactile}}} 
    & \multicolumn{1}{c|}{\raisebox{-1pt}{\includegraphics[width=7.5pt]{fig/icon/eeg.png} \textbf{EEG}}} 
    & \multicolumn{1}{c}{\raisebox{-1pt}{\includegraphics[width=7.5pt]{fig/icon/pointcloud.png} \textbf{3D}}} \\ 
\midrule
  & NYU-D & SUN-D & Audioset & VGGS & ESC & Clotho & AudioCaps & TAG-M & TAG-H/S & TAG-R/S & IN-EEG & ModelNet40 \\ \midrule
ViT-Lens               
  & 68.5 & 52.2 & 26.7 & 31.7 & 75.9 & 8.1/31.2 & 14.4/54.9 
 & 65.8 & 74.7 & 63.8 & 41.8/42.7  & 70.6/94.4 \\ \midrule
CodeBind-VL  
 & \textbf{71.1} & \textbf{54.8} & \textbf{29.2} & \textbf{39.5} & \textbf{78.8} & \textbf{8.5/32.8} & \textbf{15.6/55.0} 
 & \textbf{67.6} & \textbf{76.1} & \textbf{72.8} & \textbf{54.5/54.1} & \textbf{78.3/96.5} \\ \bottomrule
\end{tabularx}

\caption{
CodeBind-VL integrates our method into ViT-Lens, and classification results from EEG and 3D point clouds are included.
}
\end{subtable}
\vspace{-0.25cm}
\caption{\textbf{Multi-modal classification and retrieval results on diverse benchmarks.} CodeBind-IB (a) and CodeBind-VL (b) maintain the dataset and modality setting with ImageBind and ViT-Lens respectively, for comparison fairness. The results highlight our model's superior alignment capabilities across nine modalities.
}
\vspace{-0.25cm}
\label{tab_combined}
\end{table*}

\begin{table*}[htb]
\centering
\scriptsize
\setlength{\tabcolsep}{5.8pt}

\begin{tabular*}{\textwidth}{l | c | cc | ccccc | cc | c}
\toprule
 & \multicolumn{1}{c|}{\raisebox{-1pt}{\includegraphics[height=7.5pt]{fig/icon/video.png}} \textbf{Video}} 
 & \multicolumn{2}{c|}{\raisebox{-1pt}{\includegraphics[height=7.5pt]{fig/icon/depth.png}} \textbf{Depth}} 
 & \multicolumn{5}{c|}{\raisebox{-1pt}{\includegraphics[height=7.5pt]{fig/icon/audio.png}} \textbf{Audio}} 
 & \multicolumn{2}{c|}{\raisebox{-1pt}{\includegraphics[height=7.5pt]{fig/icon/thermal.png}} \textbf{Thermal}} 
 & \multicolumn{1}{c}{\raisebox{-1pt}{\includegraphics[height=7.5pt]{fig/icon/pointcloud.png}} \textbf{3D}} \\
\midrule
Method & MSR-VTT & NYU-D & SUN-D & AudioSet & VGGS & ESC & Clotho & AudioCaps & LLVIP & FLIR\_v2 & ModelNet-40 \\ 
\midrule

FreeBind     & -- & -- & -- & 19.7 & -- & 93.6 & 13.7/-- & 29.2/-- & -- & -- & -- \\
OmniBind     & -- & -- & -- & 25.1 & -- & 93.5 & \textbf{23.3/49.5} & \textbf{46.7/79.7} & -- & -- & -- \\
LanguageBind & \textbf{44.8} & -- & -- & \textbf{30.0} & 38.6 & \textbf{94.0} & 16.7/52.0 & 19.7/67.6 & 87.2 & 48.0 & -- \\
OneLLM-D     & -- & 50.9 & 29.0 & -- & -- & -- & -- & -- & -- & -- & -- \\ 
OneEncoder   & -- & -- & 28.4 & -- & 80.1 & -- & -- & -- & -- & -- \\
\textsc{UniAlign}     & 37.0 & \textbf{71.4} & \textbf{58.2} & 23.8 & -- & 70.5 & -- & 11.7/49.3 & -- & -- & 55.9/84.3 \\ \midrule
CodeBind-IB  & 37.8 & 59.3 & 45.7 & 21.1 & 30.5 & 71.0 & 6.9/28.6 & 13.3/53.8 & \textbf{95.5} & \textbf{97.2} & -- \\
CodeBind-VL  & -- & 71.1 & 54.8 & 29.2 & \textbf{39.5} & 78.8 & 8.5/32.8 & 15.6/55.0 & -- & -- & \textbf{78.3/96.5} \\
\bottomrule
\end{tabular*}

\vspace{-0.2cm}
\caption{
Compared to other strong SOTA methods that use different data crowdsourcing approaches, our method is comparable without such approaches.
}
\label{tab_strong_sota}
\end{table*}

\section{Experiments}
\vspace{-0.1cm}

We evaluate CodeBind on 9 modalities, including text, image, video, audio, depth, thermal, 3D point cloud, tactile, and EEG. Tab.~\ref{tab_datasets} summarizes the datasets (App.~\ref{supp: dataset details} for details).
Our codebook design and training objectives are integrated into two SOTA baselines, ImageBind~\citep{girdhar2023imagebind} and ViT-Lens~\citep{lei2024vitlens}, resulting in CodeBind-IB and CodeBind-VL. 
We employ compositional VQ with 1024 shared and 256 specific codevectors, each with a dimensionality of 8. Training is initialized using pre-trained ImageBind and ViT-Lens, with a learning rate of $5 \times 10^{-4}$ on eight NVIDIA RTX 3090 GPUs (App.~\ref{supp: training details} for details).

\subsection{Main Comparison}
\label{exp: alignment comparison}
Similar to ImageBind and ViT-Lens, we evaluate our method on multimodal classification and retrieval tasks.
Tab.~\ref{tab_combined} demonstrates consistent improvements with our approach compared to baselines.

\paragraph{Multimodal classification}
We extract aligned feature of any modality and assign it to the class with the highest similarity to the textual prompt embeddings defined by ~\citet{radford2021learning}.
In \textit{image and video classification} (ImageNet~\citealp{russakovsky2015imagenet}, Placese365~\citealp{zhou2014learning}, Kinetics400~\citealp{kay2017kinetics}), ImageBind maintains parity with OpenCLIP~\citep{ilharco2021openclip} while our plug-in codebooks yield consistent gains. 
In \textit{depth scene classification}, CodeBind-IB achieves a notable improvement of +5.3\%/+10.6\% on NYU-D~\citep{silberman2012nyu} and SUN-D~\citep{song2015sun}.
In \textit{audio classification}, we outperform ImageBind by +3.5\%/+2.7\%/+4.1\%
across three audio datasets (AudioSet~\citealp{gemmeke2017audioset}, VGGSound~\citealp{chen2020vggsound}, ESC~\citealp{piczak2015esc}).
In \textit{thermal classification}, CodeBind-IB achieves substantial boosts (+32.1\% on LLVIP~\citep{jia2021llvip}, +50.6\% on FLIR\_v2~\citep{flirv2kaggle}). 
Furthermore, we observe remarkable improvement in Tough-and-go~\citep{yang2022touch_and_go} and ImageNet-EEG~\citep{spampinato2017eeg} for tactile and EEG visual concept classification.
In Tab.~\ref{tab_combined} (b), CodeBind-VL achieves consistent performance gains across tasks. In \textit{point cloud classification} using ShapeNets~\citep{wu2015shapenets} for training and evaluation on ModelNet40~\citep{wu2015modelnet40}, it improves Acc@1 by +7.7\% and Acc@5 by +2.1\% over ViT-Lens.

\paragraph{Multimodal retrieval}
In \textit{video and audio retrieval}, CodeBind-IB consistently outperforms the baseline, yielding a +1.7\% gain on MSR-VTT~\citep{xu2016msr} and notable improvements on Clotho~\citep{drossos2020clotho} (+0.9\% Recall@1, +0.2\% Recall@10). These gains are even more pronounced on AudioCaps, with substantial increases of +4.0\% and +11.5\%. CodeBind-VL further confirms the consistent enhancement of our method.

\paragraph{Comparison with other strong SOTA methods}
FreeBind~\citep{wang2024freebind} and OmniBind~\citep{wang2024omnibind} generate pseudo-data pairs retrieved from pre-trained bi-modal expert models to bridge disparate spaces. LanguageBind~\citep{zhu2024languagebind} scales up to 10M dataset, including synthetic data for scarce modalities like thermal.  
Unified architectures like OneLLM~\citep{han2024onellm} and OneEncoder~\citep{faye2024oneencoder} forcibly align all modalities into a single encoder, exacerbating the risk of catastrophic forgetting and sub-optimal alignment. 
\textsc{UniAlign}~\citep{zhou2025unialign} achieves extreme parameter efficiency (7.8M) through a modality-aware MoE schema but overlooks fine-grained, modality-specific details.
In contrast, our codebook-based alignment leverages naturally paired data rather than labor-intensive or synthetic data expansion. By utilizing compositional VQ to represent shared and specific features, our method achieves comparable performance with efficient parameterization ($\sim$50M), as shown in Table~\ref{tab_strong_sota}.

\vspace{-0.1cm}
\subsection{Embedding Space Optimization} 

\begin{figure}[tb]
\centering
\includegraphics[width=0.9\linewidth]{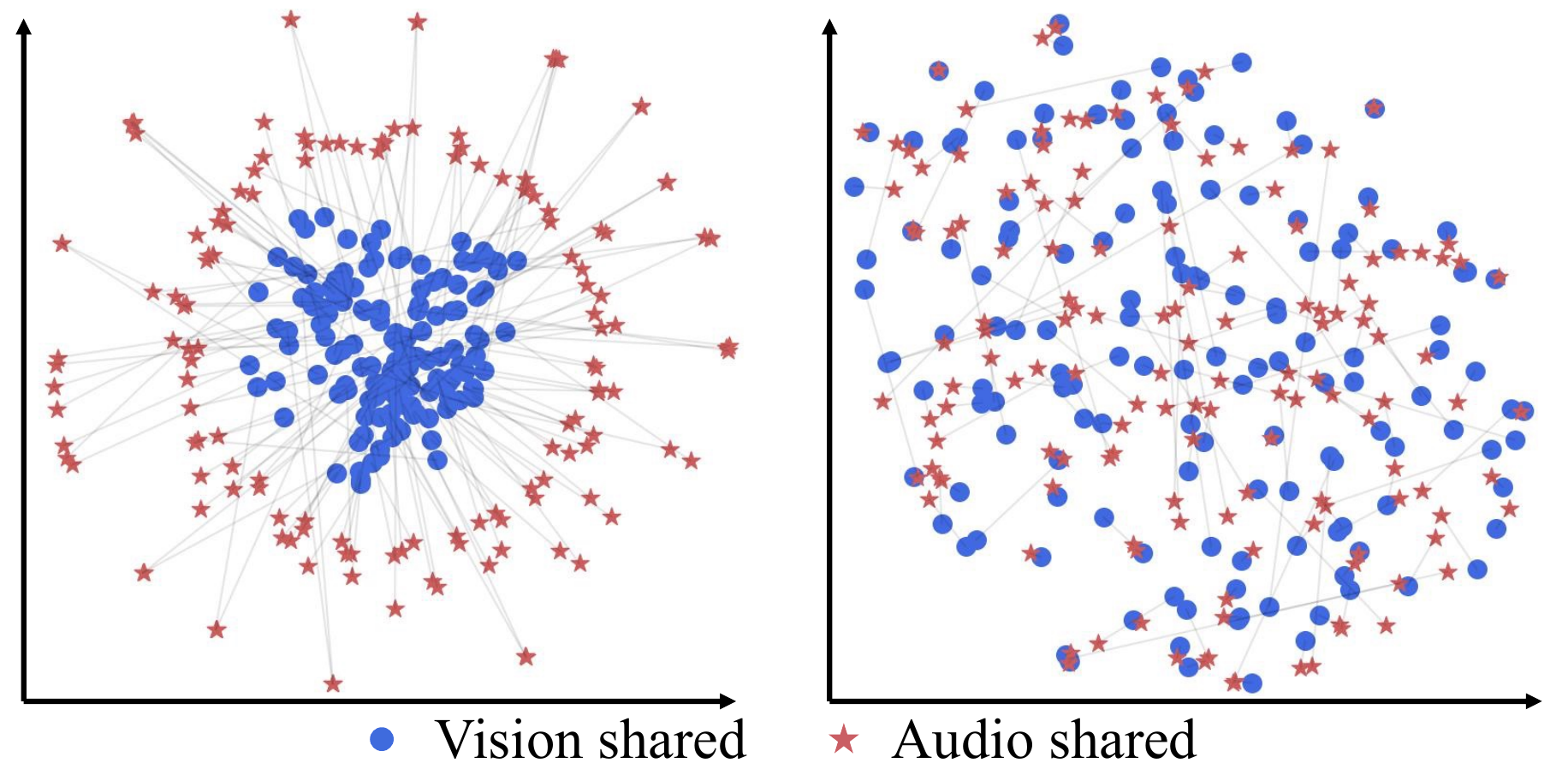}
\vspace{-0.1cm} 
\caption{
\textbf{Visualization of embeddings in unified space.} T-SNE visualization of sampled embeddings by ImageBind ({left}) and CodeBind-IB ({right}) using AudioSet~\cite{gemmeke2017audioset}. The paired embeddings are linked by a grey line.
}
\vspace{-0.4cm}
\label{fig_embedding_distribution}

\end{figure}

\begin{figure}[tb]
\centering
\includegraphics[width=0.98\linewidth]{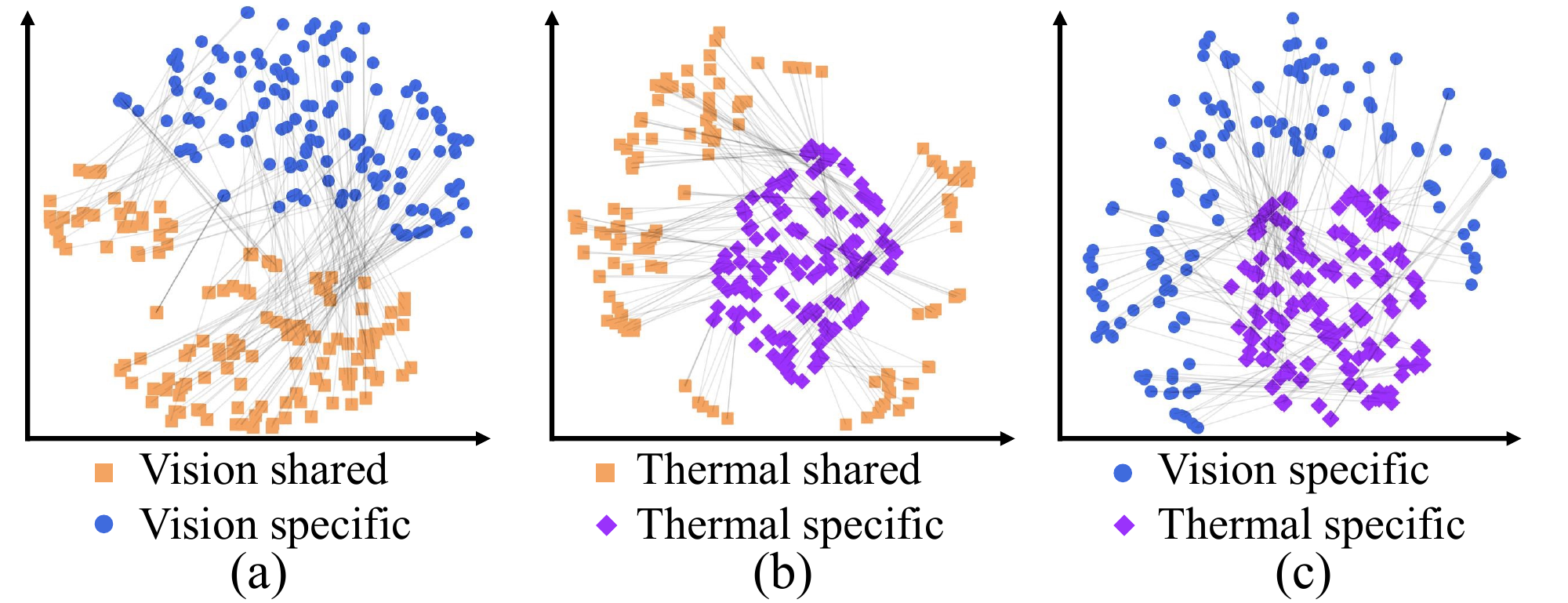}
\vspace{-0.1cm} 
\caption{
\textbf{Visualization of decoupled embeddings of image and thermal} from FLIR\_v2~\cite{flirv2kaggle}.
}
\vspace{-0.1cm}
\label{fig_embedding_info}
\end{figure}

\vspace{-0.1cm}
We qualitatively evaluate the reduction in modality gap through t-SNE visualization~\cite{van2008visualizing} of embedding distributions. In Fig.~\ref{fig_embedding_distribution}, while ImageBind exhibits a clear inter-modal gap, our shared embeddings are closely clustered and mixed, directly correlating with our improved alignment accuracy. Furthermore, we visualize the distribution of shared and specific components to validate their decoupled properties. As illustrated in Fig.~\ref{fig_embedding_info}, our method effectively separates shared semantics from modality-specific features in both thermal and vision domains. Detailed codevector-level analysis is available in App.~\ref{supp: codebook feature space analysis}.

\begin{table}[tb]
\centering
\scriptsize
\setlength{\tabcolsep}{2pt}
\newcolumntype{Y}{>{\centering\arraybackslash}X}

\begin{tabularx}{\columnwidth}{cc|Y|Y|Y}
\toprule
\multicolumn{2}{c|}{Dataset} & Stanford Dogs & Oxford Pet Cats & Oxford Pet Dogs \\ 
\multicolumn{2}{c|}{Sample Size} & 6000 & 2371 & 4978 \\ \midrule
\multicolumn{2}{c|}{ImageBind} & 50.4 & 87.0 & 92.2 \\ \midrule
\multirow{3}{*}{Ours} 
& Specific & 48.9 & 83.7 & 91.1 \\ 
& Shared & \textbf{63.5} & \underline{88.3} & \textbf{94.6} \\ 
& Concatenated & \underline{60.2} & \textbf{88.4} & \underline{94.4} \\ \bottomrule
\end{tabularx}

\vspace{-0.1cm}
\caption{\textbf{Fine-grained retrieval recall.} The best result is \textbf{bolded}, and the second best result is \underline{underlined}. Our decoupled components achieve higher accuracy on fine-grained tasks.}
\vspace{-0.3cm}
\label{tab_fine_retrieval}
\end{table}

\begin{figure}[tb]
\centering
\includegraphics[width=0.98\linewidth]{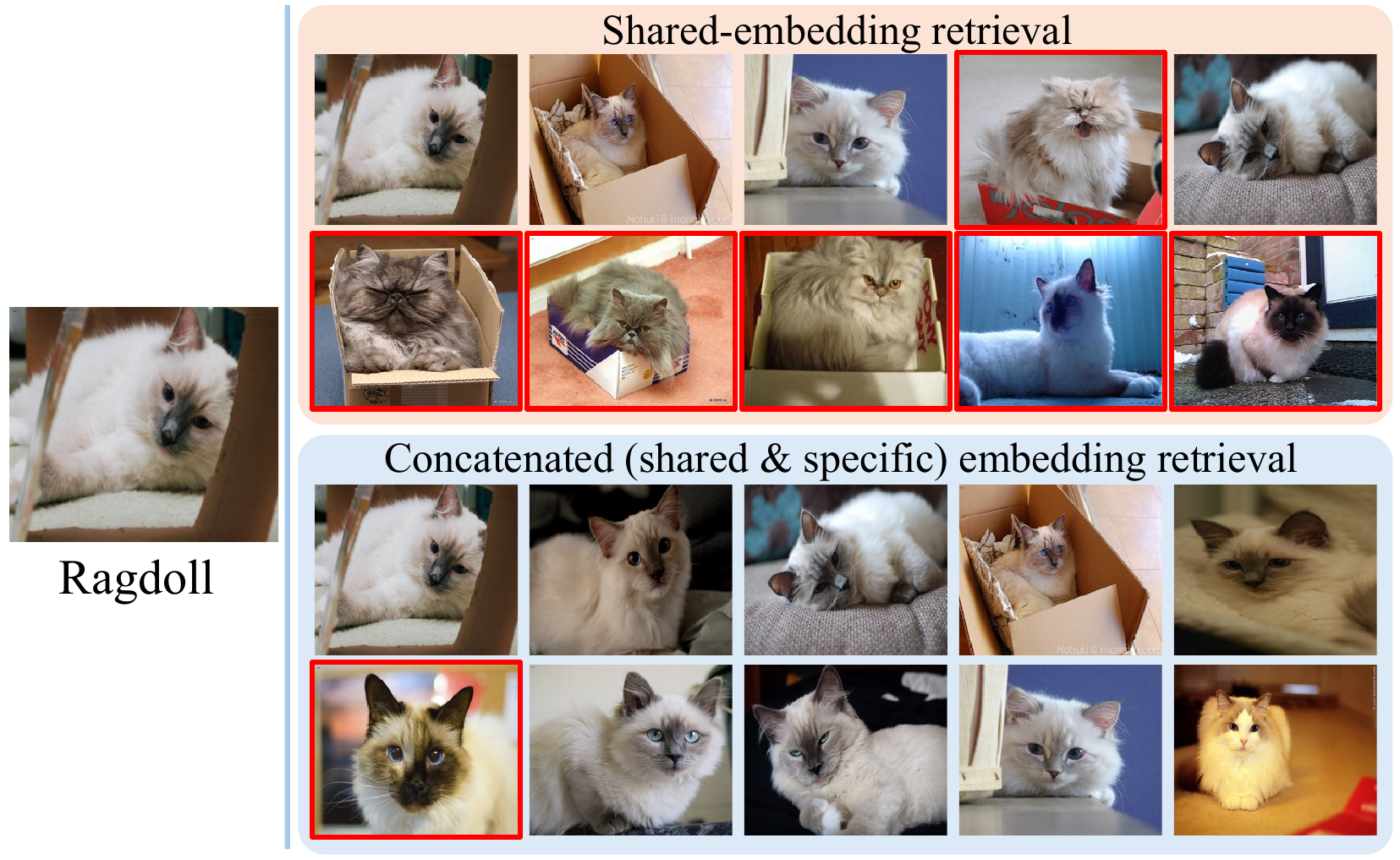}
\vspace{-0.3cm} 
\caption{
\textbf{Fine-grained retrieval results}. Source image `ragdoll' and retrieval results with incorrect matches in red boxes.
}
\vspace{-0.2cm}
\label{fig_fine_retrive}

\end{figure}

\begin{figure}[tb]
\centering
\includegraphics[width=0.9\linewidth]{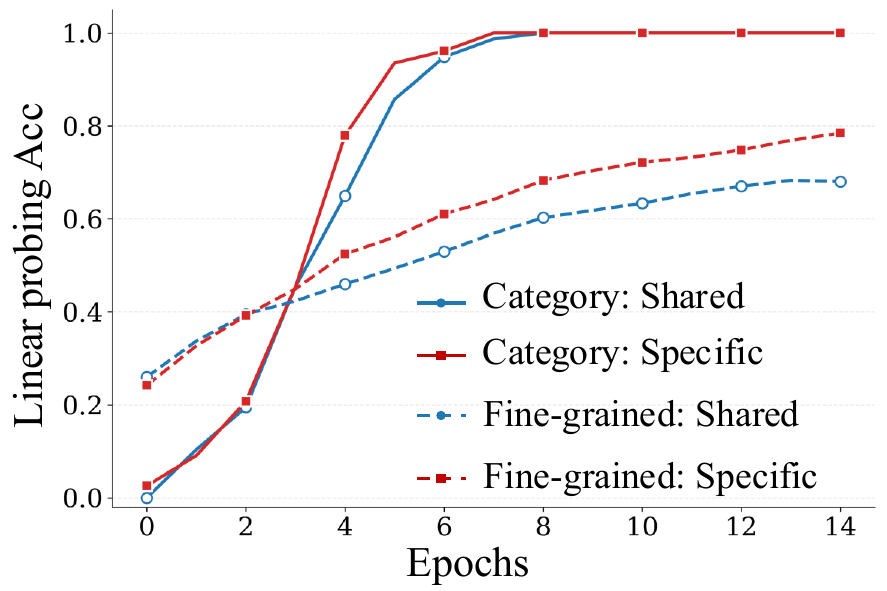}
\vspace{-0.2cm} 
\caption{
\textbf{Linear probing} for category names and fine-grained details using shared and specific embeddings from ImageNet~\cite{russakovsky2015imagenet} samples.
}
\vspace{-0.2cm}
\label{fig_linear_prob}

\end{figure}

\begin{table}[htbp]
\centering
\scriptsize
\setlength{\tabcolsep}{2pt}
\newcolumntype{Y}{>{\centering\arraybackslash}X}
\begin{tabularx}{\columnwidth}{lY|cc|YY}
\toprule
\multicolumn{2}{c|}{\textbf{Configuration}} & \multicolumn{2}{c|}{\textbf{Space}} & \multicolumn{2}{c}{\textbf{Accuracy (\%)}} \\ 
Method & Text Density & Shared & Spec. & Concat & Sum \\ \midrule
ImageBind & Category & --- & --- & 94.4 & 89.0 \\ \midrule
\multirow{2}{*}{CodeBind-IB} & \multirow{2}{*}{Category} & \cmark & \cmark & \underline{95.7} & \underline{91.6} \\
 & & \cmark & \xmark & 95.5 & 89.8 \\ \midrule
\multirow{2}{*}{\shortstack[l]{CodeBind-IB}} & \multirow{2}{*}{Dense} & \cmark & \cmark & \textbf{97.3} & \textbf{94.0} \\
 & & \cmark & \xmark & 96.0 & 92.8 \\
\bottomrule
\end{tabularx}
\vspace{-0.1cm}
\caption{\textbf{Multimodal fusion for event classification on AVE dataset~\citep{tian2018ave}} Event category classification accuracy is reported for shared embeddings alone versus fusion with specific embeddings (concatenation/summation). The impact of text density during the multimodal alignment training stage is further investigated by comparing concise event categories with VLM-generated dense descriptions as text embeddings.} 
\vspace{-0.2cm}
\label{tab_modal_fusion}
\end{table}

\vspace{-0.1cm}

\subsection{Retention of Modality-Unique Information}
To assess the retention of modality-unique information, we conduct fine-grained intra-modal retrieval and linear probing to distinguish shared semantics from fine-grained modality-unique details.

\vspace{-0.1cm}
\paragraph{Fine-grained intra-modal retrieval}
To evaluate fine-grained details within our decoupled embeddings, we perform an image-to-image retrieval on fine-grained datasets, Stanford Dogs~\citep{Khosla2011standforddog} and Oxford-IIIT Pet~\citep{parkhi2012oxfordpet}.
In Tab.~\ref{tab_fine_retrieval}, our concatenated embeddings surpass ImageBind in average Top-10 recall, validating that the fusion of shared and specific components enhances intra-modality information preservation. Notably, our specific embeddings alone achieve performance comparable to ImageBind, highlighting their representational utility. Qualitative results in Fig.~\ref{fig_fine_retrive} further confirm that specific embeddings refine visual similarity, effectively complementing shared semantics with fine-grained details.

\paragraph{Linear probing for fine-grained details}
We further conduct linear probing on both semantic and fine-grained attributes. 
Specifically, we leverage a vision language model (VLM), Qwen2.5-VL-72B~\citep{Qwen2.5VL}, to curate a fine-grained attribute taxonomy for ImageNet, extracting non-semantic properties, such as lighting, texture, and background color, that are decoupled from category labels.
We then train independent linear probes to classify the semantic labels and the fine-grained labels. A probing swap is then performed to assess cross-component predictive power. 
Fig.~\ref{fig_linear_prob} shows that specific embeddings achieve superior accuracy and faster convergence in fine-grained attribute prediction, validating their proficiency in encoding physical nuances. Interestingly, the comparable performance of specific embeddings in semantic classification suggests a natural coupling between category and fine-grained attributes, a phenomenon further corroborated by our mutual information analysis (in App.~\ref{supp: linear probing}).

\paragraph{Multimodal fusion}
To directly examine the utilization of complementary cross-modal cues, we conduct multimodal fusion experiments on the AVE dataset~\citep{tian2018ave} for audio-visual event classification. We compare CodeBind-IB against ImageBind using straightforward fusion strategies, namely concatenation and summation. To further explore the impact of language density, we leverage Qwen3-VL-4B~\citep{bai2025qwen3} to generate dense, descriptive captions for video frames, moving beyond coarse category labels. More details are in App.~\ref{supp: multimodal fusion}.
As shown in Tab.~\ref{tab_modal_fusion}, integrating the video and audio specific embeddings alongside the shared ones consistently outperforms the use of shared embeddings in isolation. This gain is further amplified when employing VLM-generated dense descriptions as text embeddings during alignment, achieving an accuracy of 97.3\%, a significant margin over ImageBind's 94.4\%.
This empirically validates that our specific embeddings successfully capture and supply non-redundant cross-modal cues that directly benefit multimodal fusion tasks.

\vspace{-0.05cm}
\subsection{Ablation Study}
Our ablation study examines: (1) the effects of the codebook modules, the decoupling strategy, and the reconstruction modules; (2) codebook settings, including the advantages of compositional VQ over standard VQ, codebook size, and dimension; (3) the contribution of various loss terms.  

\begin{table}[tb]
\centering
\scriptsize 
\setlength{\tabcolsep}{2pt} 

\newcolumntype{Y}{>{\centering\arraybackslash}X}

\begin{tabularx}{\columnwidth}{cYY|ccc}
\toprule
\multicolumn{3}{c|}{\textbf{Components}} & \multicolumn{3}{c}{\textbf{Datasets}} \\ 
Codebook & Decoupled & Reconstruction & NYU-D & SUN-D & FLIR\_v2 \\ \midrule
\xmark & \xmark & \xmark & 54.0 & 35.1 & 46.6 \\ 
\xmark & \cmark  & \cmark & 54.1 & 39.7 & 94.5 \\ 
\cmark & \xmark & \xmark & \underline{57.6} & \textbf{46.9} & 80.5 \\
\cmark & \cmark & \xmark & 56.7 & 45.3 & \textbf{97.7} \\
\cmark & \cmark & \cmark & \textbf{59.3} & \underline{45.7} & \underline{97.2} \\ 
\bottomrule
\end{tabularx}

\vspace{-0.1cm}
\caption{\textbf{Effectiveness of each component.} 
Our codebook enhances performance; decoupling and reconstruction provide extra improvement.
}
\vspace{-0.2cm}
\label{tab_modules}
\end{table}

\begin{table}[tb]
\centering
\scriptsize 
\setlength{\tabcolsep}{2pt}
\newcolumntype{Y}{>{\centering\arraybackslash}X}

\begin{tabularx}{\columnwidth}{YY|ccc}
\toprule
\multicolumn{2}{c|}{\textbf{Codebook Config.}} & \multicolumn{3}{c}{\textbf{Datasets}} \\ 
Shared & Compositional & NYU-D & SUN-D & FLIR\_v2 \\ \midrule
\cmark & \xmark     & 48.5 & 40.7 & 81.1 \\
\xmark     & \cmark & \underline{56.1} & \underline{45.7} & \underline{90.3} \\
\cmark & \cmark & \textbf{59.3} & \textbf{45.7} & \textbf{97.2} \\ 
\bottomrule
\end{tabularx}

\vspace{-0.2cm}
\caption{
\textbf{Different codebook settings} with shared/separated codebooks and compositional/conventional VQ.
}
\vspace{-0.3cm}
\label{tab_codebooks}
\end{table}

\begin{table}[tb]
\centering
\scriptsize
\setlength{\tabcolsep}{1pt}

\newcolumntype{Y}{>{\centering\arraybackslash}X}

\begin{tabularx}{\columnwidth}{YYYYY | ccc}
\toprule
$\mathcal{L}_{\mathrm{align}}$ & $\mathcal{L}_{\mathrm{cm}}$ & \makecell{$\mathcal{L}_{\mathrm{cctr}},$ \\ $\mathcal{L}_{\mathrm{cuni}}$} & \makecell{$\mathcal{L}_{\mathrm{orth}},$ \\ $\mathcal{L}_{\mathrm{uni}}$} & $\mathcal{L}_{\mathrm{recon}}$ & NYU-D & SUN-D & FLIR\_v2 \\ \midrule
\cmark & \xmark & \xmark & \xmark & \xmark & 54.0 & 35.1 & 46.6 \\
\cmark & \cmark & \xmark & \xmark & \xmark & 56.4 & \underline{46.0} & \textbf{98.7} \\
\cmark & \cmark & \cmark & \xmark & \xmark & 56.7 & 45.3 & \underline{97.7} \\
\cmark & \cmark & \cmark & \cmark & \xmark & \underline{57.6} & \textbf{46.9} & 80.6 \\
\cmark & \cmark & \cmark & \cmark & \cmark & \textbf{59.3} & 45.7 & 97.2 \\ \bottomrule
\end{tabularx}

\vspace{-0.1cm}
\caption{\textbf{Ablation on loss functions building upon one another.}}
\vspace{-0.2cm}
\label{tab_loss}
\end{table}

\begin{figure*}[htb]
\centering
\includegraphics[width=0.98\linewidth]{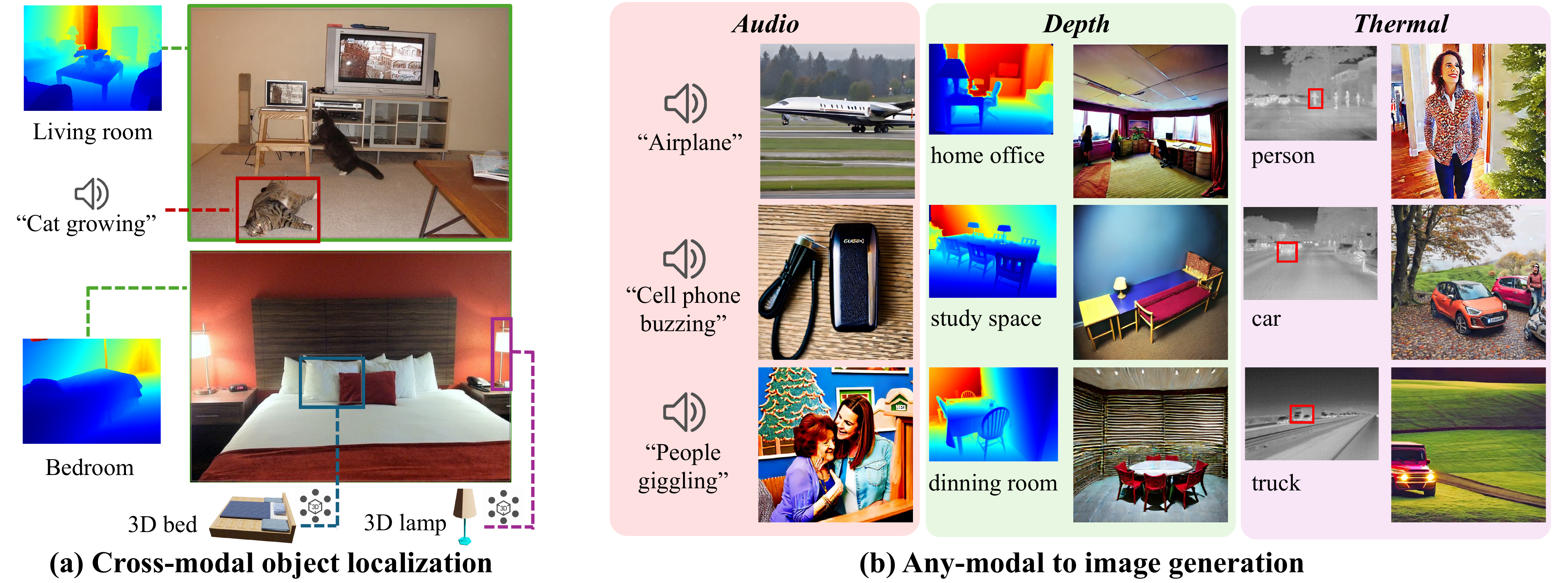}
\vspace{-0.1cm} 
\caption{
\textbf{Several applications utilizing alignment space of CodeBind.} 
(A) Semantic aligned embeddings from depth, audio, and 3D point cloud are retrieved using the image bounding box region as vision proposals.
(B) By replacing the image encoder with any modality encoder in Stable unCLIP, we achieve any-modal-to-image generation.
}
\vspace{-0.4cm}
\label{fig_downstream_application}

\end{figure*}

\begin{figure}[tb]
\centering
\includegraphics[width=0.98\linewidth]{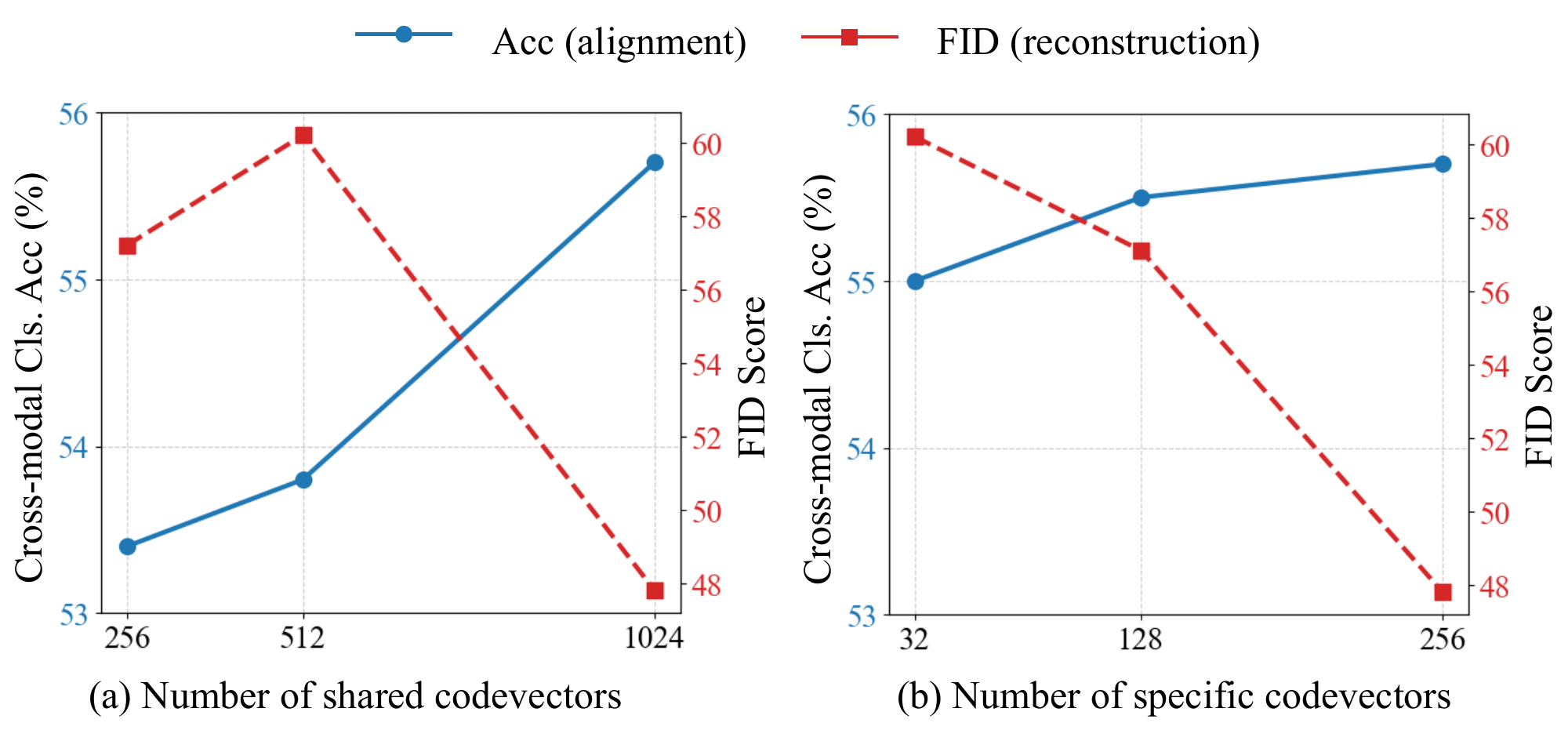}
\vspace{-0.1cm} 
\caption{
\textbf{Impact of codebook size on alignment and reconstruction.} 
(a) Varying shared codebook size when fixing specific codebook size at 256 (b) Varying specific codebook size when fixing shared codebook size at 1024. Cross-modal classification accuracy and FID of reconstructed images are compared using NYU-D~\citep{silberman2012nyu} with codevector dim=128.
}
\vspace{-0.3cm}
\label{fig_codebook_size}
\end{figure}

\paragraph{Codebook, decoupling and reconstruction}
Table~\ref{tab_modules} shows that incorporating codebook design improves modality alignment, yielding gains of +3.6\%/+11.8\%/+33.9\% on NYU-D, SUN-D and FLIR\_v2 over ImageBind. This confirms the shared codebook's efficacy in facilitating robust alignment in a unified space. Furthermore, representation decoupling into shared and specific streams delivers substantial performance leaps (up to +48.0\%), validating the necessity of isolating the semantic core from modality-specific nuances. Notably, the codebook and decoupling modules alone, before adding the reconstruction module, already deliver substantial gains.

\paragraph{Loss functions}
Tab.~\ref{tab_loss} shows training objectives building upon one another. Our regularization losses significantly enhance representation disentanglement and informativeness while achieving superior alignment performance (details in App.~\ref{sec: loss ablation supp}).

\paragraph{Codebook design and size}
In Tab.~\ref{tab_codebooks}, the shared codebook approach outperforms separate codebooks, showing improvements of +3.2\%/+6.9\% on NYU-D/FLIR\_v2. 
Compared to standard VQ, our compositional VQ significantly improves alignment by +10.8\%, +5.7\%, and 16.1\% on three datasets. 
The main advantage is its expanded capacity, which can accommodate more information without the need for codebook expansion.

Fig.~\ref{fig_codebook_size} shows that codebook size controls the trade-off between cross-modal alignment efficacy and intra-modal reconstruction fidelity. Expanding the shared codebook boosts cross-modal alignment. The specific codebook size primarily modulates reconstruction quality without notable impacts on alignment. These results confirm our codebooks' role as information moderators: the shared one extracts cross-modal invariants, while the specific ones retain modal-unique features. Furthermore, while the reconstruction module adds training overhead, it is discarded during inference without compromising alignment (Tab.~\ref{tab_modules}). Its inclusion remains vital, however, to ensure the representational integrity of modality-specific information.
(See App.~\ref{sec: codebook ablation supp} for more analysis.)

\subsection{More Applications}

CodeBind enables zero-shot cross-modal object localization and any-modal-to-image generation by seamlessly integrating diverse modalities into established vision-language and generative frameworks (\eg GroundingDINO~\citep{liu2024groundingdino} and Stable unCLIP~\citep{rombach2022stablediffusion}) without additional training.

\section{Conclusion}
\vspace{-0.1cm}
This paper proposes CodeBind, a novel multi-modal alignment framework that effectively aligns diverse modalities without requiring large-scale fully-paired datasets, thereby delivering enhanced practicality and scalability.
Through decoupled representations and a shared-specific codebook mechanism, CodeBind significantly enhances cross-modal compatibility while preserving distinct modality characteristics.  
Extensive experiments consistently demonstrate the superiority of CodeBind in cross-modal classification and retrieval tasks. It optimizes the representation space by reducing modality gaps while preserving informative modality-unique details, revealing its great potential in various downstream tasks. 

\section*{Limitations}
\vspace{-0.05cm}

In our experiments, while modality-specific information in visual embeddings is validated via text-anchored interpretability provided by VLM, extending this to uncommon modalities (which lack strong foundational spaces) is currently challenging. Nevertheless, our decoupled architecture provides the structural prerequisite for future explorations of the modality-specific information in complex multimodal reasoning tasks like sentiment analysis.
In addition, though we primarily utilize category names rather than detailed descriptions during alignment for fair comparison, incorporating LLM-generated dense semantic descriptions, as suggested by recent findings \citep{fan2024denseinfo}, could further unlock the potential of our decoupled space.
Finally, our plug-and-play design enables integration into MLLMs for ``on-demand'' fusion. Future work can employ gated mechanisms to dynamically switch between cross-modal consistent concepts and modality-unique cues, adapting to varying scenario requirements.
Moving forward, our shared-specific decoupling fosters safer and more interpretable multimodal systems. In robotics, isolating modality-specific signals shields the shared semantic core from sensor interference. In medical diagnostics, this transparency enables clinicians to trace decisions to specific modality cues, ensuring clinical accountability.

\section*{Acknowledgments}
This work is supported by the Hong Kong Research Grants Council - General Research Fund (Grant No.: $17211024$) and HKU Seed Fund for PI Research.

\bibliography{main}

@String(IJCV = {Int. J. Comput. Vis.})

@String(CVPR= {IEEE Conf. Comput. Vis. Pattern Recog.})

@String(ICCV= {Int. Conf. Comput. Vis.})

@String(ECCV= {Eur. Conf. Comput. Vis.})

@String(NIPS= {Adv. Neural Inform. Process. Syst.})

@String(ACMMM= {ACM Int. Conf. Multimedia})

@String(ICASSP=	{IEEE International Conference on Acoustics, Speech and Signal Processing})

@String(ICLR = {Int. Conf. Learn. Represent.})

@String(IJCAI = {IJCAI})

@String(ACL = {Annual Meeting of the Association for Computational Linguistics})

@String(NAACL = {North American Chapter of the Association for Computational Linguistics})

@String(IJCV  = {IJCV})

@String(CVPR  = {CVPR})

@String(ICCV  = {ICCV})

@String(ECCV  = {ECCV})

@String(NIPS  = {NeurIPS})

@String(ACMMM = {ACM MM})

@String(ICASSP=	{ICASSP})

@String(ICLR  = {ICLR})

@String(ACL = {ACL})

@String(NAACL = {NAACL})

@String(ICML = {ICML})

@string{ACLfindings = {ACL(Findings)} }

@inproceedings{jiang2023understanding,
  title={Understanding and constructing latent modality structures in multi-modal representation learning},
  author={Jiang, Qian and Chen, Changyou and Zhao, Han and Chen, Liqun and Ping, Qing and Tran, Son Dinh and Xu, Yi and Zeng, Belinda and Chilimbi, Trishul},
  booktitle=CVPR,
  year={2023}
}

@article{chen2024visual,
  title={Visual Neural Decoding via Improved Visual-EEG Semantic Consistency},
  author={Chen, Hongzhou and He, Lianghua and Liu, Yihang and Yang, Longzhen},
  journal={arXiv preprint arXiv:2408.06788},
  year={2024}
}

@inproceedings{xia2024achieving,
  title={Achieving cross modal generalization with multimodal unified representation},
  author={Xia, Yan and Huang, Hai and Zhu, Jieming and Zhao, Zhou},
  booktitle=NIPS,
  year={2024}
}

@inproceedings{liu2022cross,
  title={Cross-Modal Discrete Representation Learning},
  author={Liu, Alexander H and Jin, SouYoung and Lai, Cheng-I Jeff and Rouditchenko, Andrew and Oliva, Aude and Glass, James},
  booktitle=ACL, 
  year={2022}
}

@inproceedings{zheng2024unicode,
  title={Unicode: Learning a unified codebook for multimodal large language models},
  author={Zheng, Sipeng and Zhou, Bohan and Feng, Yicheng and Wang, Ye and Lu, Zongqing},
  booktitle=ECCV,
  year={2024}
}

@inproceedings{girdhar2023imagebind,
  title={Imagebind: One embedding space to bind them all},
  author={Girdhar, Rohit and El-Nouby, Alaaeldin and Liu, Zhuang and Singh, Mannat and Alwala, Kalyan Vasudev and Joulin, Armand and Misra, Ishan},
  booktitle=CVPR,
  year={2023}
}

@inproceedings{guzhov2022audioclip,
  title={Audioclip: Extending clip to image, text and audio},
  author={Guzhov, Andrey and Raue, Federico and Hees, J{\"o}rn and Dengel, Andreas},
  booktitle=ICASSP,
  year={2022},
}

@inproceedings{lei2024vitlens,
  title={Vit-lens: Towards omni-modal representations},
  author={Lei, Weixian and Ge, Yixiao and Yi, Kun and Zhang, Jianfeng and Gao, Difei and Sun, Dylan and Ge, Yuying and Shan, Ying and Shou, Mike Zheng},
  booktitle=CVPR,
  year={2024}
}

@article{lyu2024omnibind,
  title={OmniBind: Teach to Build Unequal-Scale Modality Interaction for Omni-Bind of All},
  author={Lyu, Yuanhuiyi and Zheng, Xu and Kim, Dahun and Wang, Lin},
  journal={arXiv preprint arXiv:2405.16108},
  year={2024}
}

@inproceedings{duan2022multi,
  title={Multi-modal alignment using representation codebook},
  author={Duan, Jiali and Chen, Liqun and Tran, Son and Yang, Jinyu and Xu, Yi and Zeng, Belinda and Chilimbi, Trishul},
  booktitle=CVPR,
  year={2022}
}

@inproceedings{dharmasiri2024cross,
  title={Cross-Modal Self-Training: Aligning Images and Pointclouds to learn Classification without Labels},
  author={Dharmasiri, Amaya and Naseer, Muzammal and Khan, Salman and Khan, Fahad Shahbaz},
  booktitle=CVPR,
  year={2024}
}

@inproceedings{lyu2024unibind,
  title={UniBind: LLM-Augmented Unified and Balanced Representation Space to Bind Them All},
  author={Lyu, Yuanhuiyi and Zheng, Xu and Zhou, Jiazhou and Wang, Lin},
  booktitle=CVPR,
  year={2024}
}

@inproceedings{zhu2024languagebind,
  title={LanguageBind: Extending Video-Language Pretraining to N-modality by Language-based Semantic Alignment},
  author={Bin Zhu and Bin Lin and Munan Ning and Yang Yan and Jiaxi Cui and WANG HongFa and Yatian Pang and Wenhao Jiang and Junwu Zhang and Zongwei Li and Cai Wan Zhang and Zhifeng Li and Wei Liu and Li Yuan},
  booktitle=ICLR,
  year={2024},
}

@inproceedings{fu2024a,
title={A Touch, Vision, and Language Dataset for Multimodal Alignment},
author={Letian Fu and Gaurav Datta and Huang Huang and William Chung-Ho Panitch and Jaimyn Drake and Joseph Ortiz and Mustafa Mukadam and Mike Lambeta and Roberto Calandra and Ken Goldberg},
booktitle=ICML,
year={2024},
}

@inproceedings{su2023pandagpt,
  title={PandaGPT: One Model To Instruction-Follow Them All},
  author={Su, Yixuan and Lan, Tian and Li, Huayang and Xu, Jialu and Wang, Yan and Cai, Deng},
  booktitle={Proceedings of the 1st Workshop on Taming Large Language Models: Controllability in the era of Interactive Assistants!},
  year={2023}
}

@article{zhang2023metatrans,
  title={Meta-transformer: A unified framework for multimodal learning},
  author={Zhang, Yiyuan and Gong, Kaixiong and Zhang, Kaipeng and Li, Hongsheng and Qiao, Yu and Ouyang, Wanli and Yue, Xiangyu},
  journal={arXiv preprint arXiv:2307.10802},
  year={2023}
}

@inproceedings{zhu2024scaling,
title={Scaling the Codebook Size of {VQ}-{GAN} to 100,000 with a Utilization Rate of 99\%},
author={Lei Zhu and Fangyun Wei and Yanye Lu and Dong Chen},
booktitle=NIPS,
year={2024},
}

@inproceedings{radford2021learning,
  title={Learning transferable visual models from natural language supervision},
  author={Radford, Alec and Kim, Jong Wook and Hallacy, Chris and Ramesh, Aditya and Goh, Gabriel and Agarwal, Sandhini and Sastry, Girish and Askell, Amanda and Mishkin, Pamela and Clark, Jack and others},
  booktitle=ICML,
  year={2021},
}

@inproceedings{cherti2023reproducible,
  title={Reproducible scaling laws for contrastive language-image learning},
  author={Cherti, Mehdi and Beaumont, Romain and Wightman, Ross and Wortsman, Mitchell and Ilharco, Gabriel and Gordon, Cade and Schuhmann, Christoph and Schmidt, Ludwig and Jitsev, Jenia},
  booktitle=CVPR,
  year={2023}
}

@misc{ilharco2021openclip,
  author       = {Ilharco, Gabriel and
                  Wortsman, Mitchell and
                  Wightman, Ross and
                  Gordon, Cade and
                  Carlini, Nicholas and
                  Taori, Rohan and
                  Dave, Achal and
                  Shankar, Vaishaal and
                  Namkoong, Hongseok and
                  Miller, John and
                  Hajishirzi, Hannaneh and
                  Farhadi, Ali and
                  Schmidt, Ludwig},
  title        = {OpenCLIP},
  month        = jul,
  year         = 2021,
  note         = {If you use this software, please cite it as below.},
  publisher    = {Zenodo},
  version      = {0.1},
  doi          = {10.5281/zenodo.5143773},
  url          = {https://doi.org/10.5281/zenodo.5143773}
}

@inproceedings{hu2022lora,
  title={Lo{RA}: Low-Rank Adaptation of Large Language Models},
  author={Edward J Hu and yelong shen and Phillip Wallis and Zeyuan Allen-Zhu and Yuanzhi Li and Shean Wang and Lu Wang and Weizhu Chen},
  booktitle=ICLR,
  year={2022},
}

@inproceedings{song2015sun,
  title={Sun rgb-d: A rgb-d scene understanding benchmark suite},
  author={Song, Shuran and Lichtenberg, Samuel P and Xiao, Jianxiong},
  booktitle=CVPR,
  year={2015}
}

@inproceedings{silberman2012nyu,
  title={Indoor segmentation and support inference from rgbd images},
  author={Silberman, Nathan and Hoiem, Derek and Kohli, Pushmeet and Fergus, Rob},
  booktitle=ECCV,
  year={2012},
}

@inproceedings{gemmeke2017audioset,
  title={Audio set: An ontology and human-labeled dataset for audio events},
  author={Gemmeke, Jort F and Ellis, Daniel PW and Freedman, Dylan and Jansen, Aren and Lawrence, Wade and Moore, R Channing and Plakal, Manoj and Ritter, Marvin},
  booktitle=ICASSP,
  year={2017},
}

@inproceedings{piczak2015esc,
  title={ESC: Dataset for environmental sound classification},
  author={Piczak, Karol J},
  booktitle=ACMMM,
  pages={1015--1018},
  year={2015}
}

@inproceedings{drossos2020clotho,
  title={Clotho: An audio captioning dataset},
  author={Drossos, Konstantinos and Lipping, Samuel and Virtanen, Tuomas},
  booktitle=ICASSP,
  year={2020},
}

@inproceedings{kim2019audiocaps,
  title={Audiocaps: Generating captions for audios in the wild},
  author={Kim, Chris Dongjoo and Kim, Byeongchang and Lee, Hyunmin and Kim, Gunhee},
  booktitle=NAACL,
  year={2019}
}

@inproceedings{chen2020vggsound,
  title={Vggsound: A large-scale audio-visual dataset},
  author={Chen, Honglie and Xie, Weidi and Vedaldi, Andrea and Zisserman, Andrew},
  booktitle=ICASSP,
  year={2020},
}

@inproceedings{yang2022touch_and_go,
  title={Touch and go: Learning from human-collected vision and touch},
  author={Yang, Fengyu and Ma, Chenyang and Zhang, Jiacheng and Zhu, Jing and Yuan, Wenzhen and Owens, Andrew},
  booktitle=NIPS,
  year={2022}
}

@inproceedings{spampinato2017eeg,
  title={Deep learning human mind for automated visual classification},
  author={Spampinato, Concetto and Palazzo, Simone and Kavasidis, Isaak and Giordano, Daniela and Souly, Nasim and Shah, Mubarak},
  booktitle=CVPR,
  year={2017}
}

@inproceedings{wu2015shapenets,
  title={3d shapenets: A deep representation for volumetric shapes},
  author={Wu, Zhirong and Song, Shuran and Khosla, Aditya and Yu, Fisher and Zhang, Linguang and Tang, Xiaoou and Xiao, Jianxiong},
  booktitle=CVPR,
  year={2015}
}

@inproceedings{han2024onellm,
  title={Onellm: One framework to align all modalities with language},
  author={Han, Jiaming and Gong, Kaixiong and Zhang, Yiyuan and Wang, Jiaqi and Zhang, Kaipeng and Lin, Dahua and Qiao, Yu and Gao, Peng and Yue, Xiangyu},
  booktitle=CVPR,
  year={2024}
}

@article{yang2024neurobind,
  title={Neurobind: Towards unified multimodal representations for neural signals},
  author={Yang, Fengyu and Feng, Chao and Wang, Daniel and Wang, Tianye and Zeng, Ziyao and Xu, Zhiyang and Park, Hyoungseob and Ji, Pengliang and Zhao, Hanbin and Li, Yuanning and others},
  journal={arXiv preprint arXiv:2407.14020},
  year={2024}
}

@inproceedings{he2024unim,
  title={UniM-OV3D: Uni-Modality Open-Vocabulary 3D Scene Understanding with Fine-Grained Feature Representation},
  author={He, Qingdong and Peng, Jinlong and Jiang, Zhengkai and Wu, Kai and Ji, Xiaozhong and Zhang, Jiangning and Wang, Yabiao and Wang, Chengjie and Chen, Mingang and Wu, Yunsheng},
  booktitle=IJCAI,
  year={2024}
}

@inproceedings{wang2024omnibind,
  title={Omnibind: Large-scale omni multimodal representation via binding spaces},
  author={Wang, Zehan and Zhang, Ziang and Zhang, Hang and Liu, Luping and Huang, Rongjie and Cheng, Xize and Zhao, Hengshuang and Zhao, Zhou},
  booktitle=ICLR,
  year={2025}
}

@inproceedings{wang2023connecting,
title={Connecting Multi-modal Contrastive Representations},
author={Zehan Wang and Yang Zhao and Xize Cheng and Haifeng Huang and Jiageng Liu and Aoxiong Yin and Li Tang and Linjun Li and Yongqi Wang and Ziang Zhang and Zhou Zhao},
booktitle=NIPS,
year={2023},
}

@inproceedings{wang2023extending,
  title={Extending multi-modal contrastive representations},
  author={Wang, Zehan and Zhang, Ziang and Liu, Luping and Zhao, Yang and Huang, Haifeng and Jin, Tao and Zhao, Zhou},
  booktitle=NIPS,
  year={2025}
}

@inproceedings{liang2022mind,
  title={Mind the gap: Understanding the modality gap in multi-modal contrastive representation learning},
  author={Liang, Victor Weixin and Zhang, Yuhui and Kwon, Yongchan and Yeung, Serena and Zou, James Y},
  booktitle=NIPS,
  year={2022}
}

@inproceedings{shi2023Understanding,
title={Understanding the modality gap in {CLIP}},
author={Peiyang Shi and Michael C. Welle and M{\r{a}}rten Bj{\"o}rkman and Danica Kragic},
booktitle={ICLR 2023 Workshop on Multimodal Representation Learning: Perks and Pitfalls},
year={2023},
}

@inproceedings{ramasinghe2024accept,
  title={Accept the modality gap: An exploration in the hyperbolic space},
  author={Ramasinghe, Sameera and Shevchenko, Violetta and Avraham, Gil and Thalaiyasingam, Ajanthan},
  booktitle=CVPR,
  year={2024}
}

@article{russakovsky2015imagenet,
  title={Imagenet large scale visual recognition challenge},
  author={Russakovsky, Olga and Deng, Jia and Su, Hao and Krause, Jonathan and Satheesh, Sanjeev and Ma, Sean and Huang, Zhiheng and Karpathy, Andrej and Khosla, Aditya and Bernstein, Michael and others},
  journal=IJCV,
  year={2015},
}

@inproceedings{zhou2014learning,
  title={Learning deep features for scene recognition using places database},
  author={Zhou, Bolei and Lapedriza, Agata and Xiao, Jianxiong and Torralba, Antonio and Oliva, Aude},
  booktitle=NIPS,
  year={2014}
}

@article{kay2017kinetics,
  title={The kinetics human action video dataset},
  author={Kay, Will and Carreira, Joao and Simonyan, Karen and Zhang, Brian and Hillier, Chloe and Vijayanarasimhan, Sudheendra and Viola, Fabio and Green, Tim and Back, Trevor and Natsev, Paul and others},
  journal={arXiv preprint arXiv:1705.06950},
  year={2017}
}

@inproceedings{xu2016msr,
  title={Msr-vtt: A large video description dataset for bridging video and language},
  author={Xu, Jun and Mei, Tao and Yao, Ting and Rui, Yong},
  booktitle=CVPR,
  year={2016}
}

@inproceedings{jia2021llvip,
  title={LLVIP: A visible-infrared paired dataset for low-light vision},
  author={Jia, Xinyu and Zhu, Chuang and Li, Minzhen and Tang, Wenqi and Zhou, Wenli},
  booktitle=ICCV,
  year={2021}
}

@misc{flirv2kaggle,
  author={Teledyne FLIR},
  title={Teledyne FLIR ADAS Thermal Dataset v2},
  howpublished={\url{https://www.kaggle.com/datasets/samdazel/teledyne-flir-adas-thermal-dataset-v2/}},
  year={2018}
}

@inproceedings{alayrac2022flamingo,
  title={Flamingo: a visual language model for few-shot learning},
  author={Alayrac, Jean-Baptiste and Donahue, Jeff and Luc, Pauline and Miech, Antoine and Barr, Iain and Hasson, Yana and Lenc, Karel and Mensch, Arthur and Millican, Katherine and Reynolds, Malcolm and others},
  booktitle=NIPS,
  year={2022}
}

@inproceedings{li2022blip,
  title={Blip: Bootstrapping language-image pre-training for unified vision-language understanding and generation},
  author={Li, Junnan and Li, Dongxu and Xiong, Caiming and Hoi, Steven},
  booktitle=ICML,
  year={2022},
}

@inproceedings{li2023blip2,
  title={Blip-2: Bootstrapping language-image pre-training with frozen image encoders and large language models},
  author={Li, Junnan and Li, Dongxu and Savarese, Silvio and Hoi, Steven},
  booktitle=ICML,
  year={2023},
}

@inproceedings{liu2023llava,
      title={Visual Instruction Tuning}, 
      author={Liu, Haotian and Li, Chunyuan and Wu, Qingyang and Lee, Yong Jae},
      booktitle=NIPS,
      year={2023},
}

@inproceedings{wang2024freebind,
  title={FreeBind: Free Lunch in Unified Multimodal Space via Knowledge Fusion},
  author={Wang, Zehan and Zhang, Ziang and Cheng, Xize and Huang, Rongjie and Liu, Luping and Ye, Zhenhui and Huang, Haifeng and Zhao, Yang and Jin, Tao and Gao, Peng and others},
  booktitle=ICML,
  year={2024},
}

@inproceedings{rombach2022stablediffusion,
  title={High-resolution image synthesis with latent diffusion models},
  author={Rombach, Robin and Blattmann, Andreas and Lorenz, Dominik and Esser, Patrick and Ommer, Bj{\"o}rn},
  booktitle=CVPR,
  year={2022}
}

@inproceedings{zheng2023online,
  title={Online clustered codebook},
  author={Zheng, Chuanxia and Vedaldi, Andrea},
  booktitle=ICCV,
  year={2023}
}

@inproceedings{van2017neural,
  title={Neural discrete representation learning},
  author={Van Den Oord, Aaron and Vinyals, Oriol and others},
  booktitle=NIPS,
  year={2017}
}

@article{van2008visualizing,
  title={Visualizing data using t-SNE.},
  author={Van der Maaten, Laurens and Hinton, Geoffrey},
  journal={Journal of machine learning research},
  year={2008}
}

@InProceedings{parkhi2012oxfordpet,
  title={Cats and Dogs},
  author={Omkar M. Parkhi and Andrea Vedaldi and Andrew Zisserman and C. V. Jawahar},
  booktitle=CVPR,
  year={2012}
}

@inproceedings{Khosla2011standforddog,
  author = {Aditya Khosla and Nityananda Jayadevaprakash and Bangpeng Yao and Li Fei-Fei},
  title = {Novel Dataset for Fine-Grained Image Categorization},
  booktitle = {First Workshop on Fine-Grained Visual Categorization, CVPR},
  year = {2011},
}

@inproceedings{zhang2024ram,
  title={Recognize anything: A strong image tagging model},
  author={Zhang, Youcai and Huang, Xinyu and Ma, Jinyu and Li, Zhaoyang and Luo, Zhaochuan and Xie, Yanchun and Qin, Yuzhuo and Luo, Tong and Li, Yaqian and Liu, Shilong and others},
  booktitle=CVPR,
  year={2024}
}

@inproceedings{liu2024groundingdino,
  title={Grounding dino: Marrying dino with grounded pre-training for open-set object detection},
  author={Liu, Shilong and Zeng, Zhaoyang and Ren, Tianhe and Li, Feng and Zhang, Hao and Yang, Jie and Jiang, Qing and Li, Chunyuan and Yang, Jianwei and Su, Hang and others},
  booktitle=ECCV,
  year={2024}
}

@inproceedings{lin2014mscoco,
  title={Microsoft coco: Common objects in context},
  author={Lin, Tsung-Yi and Maire, Michael and Belongie, Serge and Hays, James and Perona, Pietro and Ramanan, Deva and Doll{\'a}r, Piotr and Zitnick, C Lawrence},
  booktitle=ECCV,
  year={2014},
}

@inproceedings{wu2015modelnet40,
  title={3d shapenets: A deep representation for volumetric shapes},
  author={Wu, Zhirong and Song, Shuran and Khosla, Aditya and Yu, Fisher and Zhang, Linguang and Tang, Xiaoou and Xiao, Jianxiong},
  booktitle=CVPR,
  year={2015}
}

@article{han2023imagebindllm,
  title={Imagebind-llm: Multi-modality instruction tuning},
  author={Han, Jiaming and Zhang, Renrui and Shao, Wenqi and Gao, Peng and Xu, Peng and Xiao, Han and Zhang, Kaipeng and Liu, Chris and Wen, Song and Guo, Ziyu and others},
  journal={arXiv preprint arXiv:2309.03905},
  year={2023}
}

@inproceedings{zhai2023siglip,
  title={Sigmoid loss for language image pre-training},
  author={Zhai, Xiaohua and Mustafa, Basil and Kolesnikov, Alexander and Beyer, Lucas},
  booktitle=ICCV,
  year={2023}
}

@inproceedings{wu2022wav2clip,
  title={Wav2clip: Learning robust audio representations from clip},
  author={Wu, Ho-Hsiang and Seetharaman, Prem and Kumar, Kundan and Bello, Juan Pablo},
  booktitle=ICASSP,
  year={2022},
}

@inproceedings{zhang2022pointclip,
  title={Pointclip: Point cloud understanding by clip},
  author={Zhang, Renrui and Guo, Ziyu and Zhang, Wei and Li, Kunchang and Miao, Xupeng and Cui, Bin and Qiao, Yu and Gao, Peng and Li, Hongsheng},
  booktitle=CVPR,
  year={2022}
}

@inproceedings{van2017vqvae,
  title={Neural discrete representation learning},
  author={Van Den Oord, Aaron and Vinyals, Oriol and others},
  booktitle=NIPS,
  year={2017}
}

@inproceedings{esser2021vqgan,
  title={Taming transformers for high-resolution image synthesis},
  author={Esser, Patrick and Rombach, Robin and Ommer, Bjorn},
  booktitle=CVPR,
  year={2021}
}

@inproceedings{sargent2023vq3d,
  title={Vq3d: Learning a 3d-aware generative model on imagenet},
  author={Sargent, Kyle and Koh, Jing Yu and Zhang, Han and Chang, Huiwen and Herrmann, Charles and Srinivasan, Pratul and Wu, Jiajun and Sun, Deqing},
  booktitle=ICCV,
  year={2023}
}

@inproceedings{zheng2022movq,
  title={Movq: Modulating quantized vectors for high-fidelity image generation},
  author={Zheng, Chuanxia and Vuong, Tung-Long and Cai, Jianfei and Phung, Dinh},
  booktitle=NIPS,
  year={2022}
}

@article{li2022unimo,
  title={UNIMO-2: End-to-end unified vision-language grounded learning},
  author={Li, Wei and Gao, Can and Niu, Guocheng and Xiao, Xinyan and Liu, Hao and Liu, Jiachen and Wu, Hua and Wang, Haifeng},
  journal={arXiv preprint arXiv:2203.09067},
  year={2022}
}

@inproceedings{zhang2024C3,
  title={Connect, Collapse, Corrupt: Learning Cross-Modal Tasks with Uni-Modal Data},
  author={Zhang, Yuhui and Sui, Elaine and Yeung-Levy, Serena},
  booktitle=ICLR,
  year={2024}
}

@inproceedings{li2025infobridge,
  title={InfoBridge: Balanced Multimodal Integration through Conditional Dependency Modeling},
  author={Li, Chenxin and Liu, Yifan and Pan, Panwang and Liu, Hengyu and Liu, Xinyu and Li, Wuyang and Wang, Cheng and Yu, Weihao and Lin, Yiyang and Yuan, Yixuan},
  booktitle=ICCV,
  year={2025}
}

@inproceedings{tjandrasuwita2025understandalign,
  title={Understanding the emergence of multimodal representation alignment},
  author={Tjandrasuwita, Megan and Ekbote, Chanakya and Ziyin, Liu and Liang, Paul Pu},
  booktitle=ICML,
  year={2025}
}

@article{faye2024oneencoder,
  title={OneEncoder: A Lightweight Framework for Progressive Alignment of Modalities},
  author={Faye, Bilal and Azzag, Hanane and Lebbah, Mustapha},
  journal={arXiv preprint arXiv:2409.11059},
  year={2024}
}

@inproceedings{zhou2025unialign,
  title={UNIALIGN: Scaling Multimodal Alignment within One Unified Model},
  author={Zhou, Bo and Li, Liulei and Wang, Yujia and Liu, Huafeng and Yao, Yazhou and Wang, Wenguan},
  booktitle=CVPR,
  year={2025}
}

@inproceedings{liu2025casualalign,
  title={Plug-and-play Feature Causality Decomposition for Multimodal Representation Learning},
  author={Liu, Ye and Ji, Zihan and Cai, Hongmin},
  booktitle=ICCV,
  year={2025}
}

@inproceedings{groger2025limitedalign,
  title={With Limited Data for Multimodal Alignment, Let the STRUCTURE Guide You},
  author={Gr{\"o}ger, Fabian and Wen, Shuo and Le, Huyen and Brbi{\'c}, Maria},
  booktitle=NIPS,
  year={2025}
}

@inproceedings{fan2024denseinfo,
    title = {Exploring the Potential of Dense Information in Multimodal Alignment},
    author = {Fan, Zhiyuan  and Chen, Zhihong  and Wang, Benyou},
    booktitle = ACL,
    year = {2024},
}

@article{Qwen2.5VL,
  title={Qwen2.5-VL Technical Report},
  author={Bai, Shuai and Chen, Keqin and Liu, Xuejing and Wang, Jialin and Ge, Wenbin and Song, Sibo and Dang, Kai and Wang, Peng and Wang, Shijie and Tang, Jun and Zhong, Humen and Zhu, Yuanzhi and Yang, Mingkun and Li, Zhaohai and Wan, Jianqiang and Wang, Pengfei and Ding, Wei and Fu, Zheren and Xu, Yiheng and Ye, Jiabo and Zhang, Xi and Xie, Tianbao and Cheng, Zesen and Zhang, Hang and Yang, Zhibo and Xu, Haiyang and Lin, Junyang},
  journal={arXiv preprint arXiv:2502.13923},
  year={2025}
}

@inproceedings{lin2025t2dr,
  title={T2DR: A Two-Tier Deficiency-Resistant Framework for Incomplete Multimodal Learning},
  author={Lin, Han and Tang, Xiu and Li, Huan and Cao, Wenxue and Wu, Sai and Yao, Chang and Shou, Lidan and Chen, Gang},
  booktitle = ACLfindings,
  year={2025}
}

@inproceedings{huang2025fcid,
  title={Enhancing multimodal unified representations for cross modal generalization},
  author={Huang, Hai and Xia, Yan and Ji, Shengpeng and Wang, Shulei and Wang, Hanting and Fang, Minghui and Zhu, Jieming and Dong, Zhenhua and Zhou, Sashuai and Zhao, Zhou},
  booktitle=ACLfindings,
  year={2025}
}

@inproceedings{yang2023confede,
  title={Confede: Contrastive feature decomposition for multimodal sentiment analysis},
  author={Yang, Jiuding and Yu, Yakun and Niu, Di and Guo, Weidong and Xu, Yu},
  booktitle=ACL,
  year={2023}
}

@inproceedings{zeng2025ciea,
  title={Enhancing Multimodal Retrieval via Complementary Information Extraction and Alignment},
  author={Zeng, Delong and Xie, Yuexiang and Li, Yaliang and Shen, Ying},
  booktitle=ACL,
  year={2025}
}

@article{bai2025qwen3,
  title={Qwen3-vl technical report},
  author={Bai, Shuai and Cai, Yuxuan and Chen, Ruizhe and Chen, Keqin and Chen, Xionghui and Cheng, Zesen and Deng, Lianghao and Ding, Wei and Gao, Chang and Ge, Chunjiang and others},
  journal={arXiv preprint arXiv:2511.21631},
  year={2025}
}

@inproceedings{tian2018ave,
  title={Audio-visual event localization in unconstrained videos},
  author={Tian, Yapeng and Shi, Jing and Li, Bochen and Duan, Zhiyao and Xu, Chenliang},
  booktitle=ECCV,
  year={2018}
}

\appendix

\clearpage
\setcounter{page}{1}
\setcounter{section}{0}
\setcounter{table}{0}
\setcounter{figure}{0}
\setcounter{equation}{0}

\section{More Details of CodeBind framework}
\label{sec: framework_supp}

\subsection{Loss Function} 
\label{supp: loss function}

In Sec.~\ref{method:implementation} of the main paper, we present an overview of our training objective, which incorporates multiple loss functions to optimize multimodal representation for alignment. 
This section provides details of each loss function and its functionality when applied to decoupled embeddings and the modality-shared-specific codebook.

\paragraph{InfoNCE loss}  
The InfoNCE loss aligns shared embeddings between target modalities $\mathcal{T}$ and bridging modalities $\mathcal{A}$. 
The target modality $\mathcal{T}$ is characterized by data 
$X^{\mathcal{T}} = \{x^{\mathcal{T}}_1, \cdots, x^{\mathcal{T}}_N\}$ with shared embeddings 
$\{ z^{\mathcal{T}}_{\mathrm{shared},1}, \cdots, z^{\mathcal{T}}_{\mathrm{shared},N} \}$, whereas the bridging modality $\mathcal{A}$ is defined by data 
$X^{\mathcal{A}} = \{x^{\mathcal{A}}_1, \cdots, x^{\mathcal{A}}_N\}$ with shared embeddings 
$\{ z^{\mathcal{A}}_{\mathrm{shared},1}, \cdots, z^{\mathcal{A}}_{\mathrm{shared},N} \}$. 
$\hat{z}^{\mathcal{T}}_{\mathrm{shared},i} = Q(E(x^{\mathcal{T}}_{i}))$ and 
$\hat{z}^{\mathcal{A}}_{\mathrm{shared},i} = Q(E(x^{\mathcal{A}}_{i}))$ refer to the quantized shared embeddings of data 
$x^{\mathcal{T}}_i$ and 
$x^{\mathcal{A}}_i$.
The InfoNCE loss is defined as, 
\begin{equation}
\mathcal{L}_{\mathrm{X}^{\mathcal{A}},\mathrm{X}^{\mathcal{T}} } = -\log
{
\frac{\exp(\hat{z}^{\mathcal{T}}_{\mathrm{shared},i} \cdot \hat{z}^{\mathcal{A}}_{\mathrm{shared},i} / \eta ) }
     {
     \textstyle 
       \sum_{j} {\exp(\hat{z}^{\mathcal{T}}_{\mathrm{shared},i} \cdot \hat{z}^{\mathcal{A}}_{\mathrm{shared},j} / \eta )}
     }
} 
\label{eq:loss_infoNCE}
\end{equation}
where $\eta$ is a scalar temperature parameter, and $j$ refers to negative observations.
Following \citealp{girdhar2023imagebind}, we consider each example $j \neq i$ in a batch as a negative instance.
This loss function aims to bring embeddings $\hat{z}^{\mathcal{T}}_{\mathrm{shared},i}$ and $\hat{z}^{\mathcal{A}}_{\mathrm{shared},i}$ closer in the unified space, promoting multimodal alignment. 
Practically, we apply the alignment to shared embeddings of [CLS] token from encoded data, and employ a symmetric loss to enhance the alignment:
$\mathcal{L}_{align} = \mathcal{L}_{\mathrm{X}^{\mathcal{A}},\mathrm{X}^{\mathcal{T}} } + \mathcal{L}_{\mathrm{X}^{\mathcal{T}},\mathrm{X}^{\mathcal{A}} } $.
When multiple bridging modalities are involved, we calculate the InfoNCE loss separately for each bridging modality.

\paragraph{Reconstruction loss}
The shared embedding and the specific embedding are concatenated in the modality decoder to reconstruct the original input for each modality.
The reconstruction retains the embedding information, including both the shared information across modalities and the unique information within each modality.
Both [CLS] and patch token embeddings are quantized and input into the reconstruction module.
Specifically, the vector quantized concatenated embeddings of the input data $x_i^{\mathcal{M}}$, $\mathrm{concat}(\hat{z}_{\mathrm{shared}, i}^{\mathcal{M}}, \hat{z}_{\mathrm{spec}, i}^{\mathcal{M}})$ are passed through a decoder network $D(\cdot)$ to reconstruct $\hat{x}_i^\mathcal{M}$. 
Reconstruction is applied both to the bridging modalities and target modalities.
Note that in this paper, the text modality only has shared embedding without additional specific embedding and is therefore not reconstructed.
The reconstruction loss is defined as,
\begin{equation}
\mathcal{L}_{\mathrm{recon} } = \left \| x_i^{\mathcal{M}} - \hat{x}_i^{\mathcal{M}}  \right \| ^{2},\  \hat{x}_i^{\mathcal{M}} = D(Q(E(x_i^{\mathcal{M}})))
\label{eq:loss_recon}
\end{equation}

\paragraph{Orthogonal loss}
We use orthogonal loss $\mathcal{L}_{\mathrm{orth}}$ to encourage the shared and specific embeddings of the data $x_i^\mathcal{M}$ from any modality $\mathcal{M} \in \{ \mathcal{A}, \mathcal{T} \}$ to capture disentangled information. Given the quantized embeddings, $\hat{z}_{\mathrm{shared}, i}^\mathcal{M}$ and $\hat{z}_{\mathrm{spec}, i}^\mathcal{M}$, we minimize their inner product, encouraging them to contain complementary information. $\mathcal{L}_{\mathrm{orth}}$ is further averaged among bridging modalities and target modality.
The orthogonal loss is defined as,
\begin{equation}
\mathcal{L}_{\mathrm{orth} } = \frac{1}{N} \sum_i^N \langle \hat{z}_{\mathrm{shared},i}^\mathcal{M}, \hat{z}_{\mathrm{spec},i}^\mathcal{M} \rangle ^2 
\label{eq:loss_orth}
\end{equation}

\paragraph{Uniform loss}
Uniform loss is used $\mathcal{L}_{\mathrm{uni}}$ to ensure that specific embeddings are evenly distributed in the feature space, enabling them to differentiate between samples effectively. This loss complements the reconstruction loss by encouraging specific embeddings to preserve essential information.
Following~\citealp{jiang2023understanding}, the uniform loss is defined as the logarithm of the averaged Gaussian potential among all pairwise combinations of quantized specific embeddings $(\hat{z}_{\mathrm{spec}, i}^{\mathcal{M}}, \hat{z}_{\mathrm{spec},j}^{\mathcal{M}})$ within a modality $\mathcal{M}$, and is further averaged across all modalities. 
We apply uniform loss only to the specific embeddings of [CLS] tokens encoded from pairwise data for computational efficiency. Also, the loss is averaged among bridging modalities and target modality.
The uniform loss is defined as,
\begin{equation}
\mathcal{L}_{\mathrm{uni} } = \log \frac{1}{N} \sum_{i,j}^N \exp{(-||\hat{z}_{\mathrm{spec},i}^{\mathcal{M}} - z_{\mathrm{spec},j}^{\mathcal{M}}||^2)}
\label{eq:loss_uni}
\end{equation}

\paragraph{Codevector commitment loss}
Commitment loss is used during codebook learning for the encoder embeddings $z_i^\mathcal{M} = E(x_i^\mathcal{M})$ to be similar to its quantized embeddings $\hat{z}_i^\mathcal{M} = Q(z_i^\mathcal{M})$
The commitment loss is,
\begin{equation}
\mathcal{L}_{\mathrm{vq} } = \beta  \left \|  z_i^\mathcal{M} - \texttt{sg}\left [\hat{z}_i^\mathcal{M} \right ] \right \| _{2}^{2}
\end{equation}
where $\texttt{sg}(\cdot)$ denotes the stop-gradient operation, $\beta$ is the weighting factor.

\paragraph{Codevector cross-modal code matching (CMCM) loss}

We also adopt CMCM loss~\citep{liu2022cross} to further enhance the codevectors alignment across paired modalities.
Given paired data from bridging and target modalities, $(x_i^{\mathcal{A}}, x_i^{\mathcal{T}})$, the calculation is based on the probability distribution of codevector usage within the shared embeddings of [CLS] tokens of relevant paired data.
Here, the probability distribution of a codevector $e_k$ given a sub-vector $z_{i, h}^\mathcal{M} \in [z_{i,1}^\mathcal{M}, \cdots, z_{i,m}^\mathcal{M}]$ segmented by our compositional codebook approach from the original embedding $z_i^\mathcal{M}$, is defined as the softmin function over their Euclidean distance, \ie $P(e_k|z_{i,h}^\mathcal{M})=
\frac{\exp{(-||z_{i,h}^\mathcal{M}-e_k||^2)}}{\sum_l \exp{(-||z_{i,h}^\mathcal{M}-e_l||^2)}}$.
This loss ensures that the codevector distributions of data from the bridging modality and target modality in positive pairs are similar, while distributions from negative pairs are distinct. Positive pairs are defined as $(z_{i,h}^\mathcal{A}, z_{i,h}^\mathcal{T})$ and negative pairs as $(z_{i,h}^\mathcal{A}, z_{j,h}^\mathcal{T})$, with the sub-vectors corresponding to the same segmented position of the original embeddings $z_i^\mathcal{A}$ and $z_i^\mathcal{T}$.
The codevector CMCM loss is defined as,
\begin{equation}
\mathcal{L}_{\mathrm{cm}} = -\frac{1}{N+m} \sum_{i,h}^{N,m} \log \frac{\exp{S_{code}(z_{i,h}^\mathcal{A}, z_{i,h}^\mathcal{T})}}{\sum_{j}\exp{S_{code}(z_{i,h}^\mathcal{A},z_{j,h}^\mathcal{T})}}
\label{eq:loss_cm}
\end{equation}

\begin{align*}
 S_{code}(z_{i,h}^\mathcal{A}, z_{i,h}^\mathcal{T}) = &\sum_{k}P(e_k|z_{i,h}^\mathcal{A})\log P(e_k|z_{i,h}^\mathcal{T}) + \\
& \sum_{k}P(e_k|z_{i,h}^\mathcal{T})\log P(e_k|z_{i,h}^\mathcal{A})
\end{align*}

\paragraph{Codevector regularization loss} 
The codevector regularization loss, comprising contrastive ($\mathcal{L}_{\mathrm{cctr}}$) and uniform ($\mathcal{L}_{\mathrm{cuni}}$) components, ensures codevectors remain semantically distinguishable. Specifically, the codevector contrastive loss $\mathcal{L}_{\mathrm{cctr}}$ fosters a non-uniform distance distribution by widening the gap between sub-vectors and their corresponding codebook entries. Given a sub-vector $z_{i,h}$, we define positive distance as the mean distance to the top 10\% closest codevectors ($\mathcal{C}_{\mathrm{pos}}$) and negative distance as the distance to the furthest 50\% ($\mathcal{C}_{\mathrm{neg}}$). By averaging this contrastive objective across all partitioned sub-vectors and modalities, codevectors are encouraged to encode diverse, meaningful semantics. This loss is formulated as follows:
\begin{equation}
\mathcal{L}_{\mathrm{cctr}} = \frac{\frac{1}{|\mathcal{C}(\mathrm{pos})|} \sum_{e_{k_1} \in \mathcal{C}(\mathrm{pos})} d(e_{k_1}, z_{i, h})}{\sum_{e_{k_2} \in \mathcal{C}(\mathrm{neg})} d(e_{k_2}, z_{i,h})}
\label{eq:loss_cctr}
\end{equation}

The codevector uniform loss, $\mathcal{L}_{\mathrm{cuni}}$, mirrors the $\mathcal{L}_{\mathrm{uni}}$ term used for specific embeddings, aiming to promote balanced codevector utilization and prevent individual entries from being overused or underrepresented. This loss is applied independently to both shared and specific codebooks. By encouraging an even distribution of codevector usage, $\mathcal{L}_{\mathrm{cuni}}$ maintains semantic diversity and mitigates redundancy within the learned representation space. This loss is formulated as:
\begin{equation}
\mathcal{L}_{\mathrm{cuni}} = \log \frac{1}{|\mathcal{C}|} \sum_{e_m, e_n \in \mathcal{C}} \exp{(-||e_m - e_n||^2)}
\label{eq:loss_cuni}
\end{equation}

\subsection{Codebook Update Policy}
\label{supp: codebook update}  

The shared codebook is initialized by K-Means on text sub-vectors and updated via batched shared embeddings from bridging and target modalities. Specific codebooks undergo a similar K-Means initialization and refinement process using their respective target embeddings. For numerical stability and faster convergence, all sub-vectors and codevectors are normalized to unit magnitude to ensure consistent scaling.

\paragraph{Update by EMA}
We update the codebook using Exponential Moving Average (EMA)~\citep{van2017neural}. 
For a codevector $e_k \in \mathcal{C}$ and its associated feature cluster $Z = \{z_i | min_v\ \mathrm{dist}(z_i, e_v) = e_k\}$, the update process at training step $t$ is as follows with a decay factor $\gamma = 0.99$:

\begin{align}
&N_k^{(t)} = \gamma N_k^{(t-1)} + (1-\gamma) |Z| \notag \\
&f_k^{(t)} = \gamma f_k^{(t-1)} + (1-\gamma) \sum_{z\in Z} z \notag \\
&e_k^{(t)} = \frac{f_k^{(t)}}{N_k^{(t)}}
\end{align}
$N_k^0 = 0$, and $f_k^0$ are initialized per cluster as described previously. This approach ensures that the codebook evolves adaptively with the data distribution during training.

\paragraph{Reinitialization} 

We follow the approach in \citealp{zheng2023online} to adjust less-used or unused codevectors when learning the codebook.
We start by accumulatively counting the average usage of codevectors in each training mini-batch.
At each training step $t$, a decay value $\alpha_{k}^{(t)}$ is calculated for each codevector entry $e_{k}^{(t)}$ based on the accumulated average usage.
Then, feature anchors are randomly selected from the encoded features in a mini-batch.
Unused or low-used codevectors are then reinitialized as,

\begin{equation}
e_{k}^{(t)}  = e_{k}^{(t-1)} \cdot (1- \alpha _{k}^{(t)}) + \hat{z} _{k}^{(t)}  \cdot \alpha _{k}^{(t)}
\label{eq:codebook_reinit}
\end{equation}
where $\hat{z} _{k}^{(t)}$ is the sampled feature anchor.
Notably, this is specifically applied to reinitialize unused or low-used codevectors, rather than updating the active ones.

\subsection{Architecture}
\label{supp: architecture}
We utilize a two-tower Transformer to align bridging modalities (text, vision) with target modalities (audio, depth, thermal, etc.) pair-wise. Modality embeddings are decoupled into shared and specific components via trainable projection heads and quantized using respective codebooks. Cross-modal alignment relies on shared embeddings, while the full concatenated representation is used for reconstruction. Notably, for text embeddings, we do not perform this decomposition and assume that they only contain shared components. This is because text modalities generally capture high-level, coarse semantic meanings, and the shared components are sufficient to represent such information in a compressed form.

\paragraph{Modality encoder}
Our main experiments use ImageBind~\citep{girdhar2023imagebind} and ViT-Lens~\citep{lei2024vitlens} as baselines, employing a Transformer architecture for all modality encoders. Following the configurations used in ImageBind and ViTLens, in CodeBind-IB and CodeBind-VL, the encoders of bridging modalities are extracted and frozen using OpenCLIP ViT-H and Vit-B/16 respectively~\citep{cherti2023reproducible,ilharco2021openclip}.
For other modalities, the checkpoints of trained modality-specific encoders in ImageBind and ViT-Lens are resumed to develop our CodeBind-IB and CodeBind-VL. To improve the alignment of each target modality and facilitate the learning of our integrated modality-shared-specific codebooks, the encoders of target modalities are finetuned using LoRA~\citep{hu2022lora} with rank $4$ on the final $6$ layers of the Transformer structure, along with trainable projection heads.

\paragraph{Reconstruction decoder}
The reconstruction is used to impose constraints on the specific embeddings to preserve the comprehensive information of the initial data, in conjunction with the shared embeddings.
The reconstruction decoder consists of a ViT structure with 8 transformer layers. They are applied to each trained modality, except for text, which does not have specific embeddings.

\section{Implementation Details}
\label{sec: implete details}

\subsection{Datasets}
\label{supp: dataset details}

\paragraph{Audio datasets:}
\textbf{AudioSet}~\citep{gemmeke2017audioset} is used for both training and evaluation, including 10-second videos sourced from YouTube that have been annotated into 527 classes. 
We do not use the unbalanced training split with 2M clips. 
Instead, we employ the balanced training split, which includes about 20K videos.
And we use the test split of around 18K videos for evaluation. 
The prepared data from \citealp{lei2024vitlens} is used instead of processing it from scratch.
The number of datasets utilized is slightly lower than the original due to some unavailable data.
Textual descriptions for each class are generated using the class names and predefined templates from CLIP~\citep{radford2021learning}.
\textbf{VGGSound(VGGS)}~\citep{chen2020vggsound} consists of about 200K video clips, with about 15K in the test split and others in the training split. 
These clips are 10 seconds in length and are labeled with 309 sound classes, including human actions, sound-emitting objects, and human-object interactions.
\textbf{AudioCaps}~\citep{kim2019audiocaps} includes about 46K audio clips to human-written text pairs collected via crowdsourcing on the AudioSet dataset. We use it for both training and evaluation. The test split is form \citep{lei2024vitlens}, which has 813 clips for evaluation.
\textbf{ESC}~\citep{piczak2015esc} is for the Environmental Sound Classification (ESC) task, which consists of 2K 5s audio clips classified into 50 classes.
It is only used for evaluation, employing a predefined 5-fold evaluation, each with 400 test audio clips.
We compute predictions on the evaluation set for each fold and report the 5-fold average performance.

\paragraph{Depth datasets:}
\textbf{SUN-D}~\citep{song2015sun} and \textbf{NYU-D}~\citep{silberman2012nyu} are RGB-D datasets used for scene classification, which contain registered RGB and Depth maps.
SUN-D is used for both training and testing, while NYU-D is used only for testing.
For depth map preprocessing, we follow ImageBind and Vit-Lens, utilizing in-filled depth values and converting them to disparity for scale normalization.
In our experiments, we also follow ImageBind and Vit-Lens to map scene class names to 10 categories and employ the same evaluation method.

\paragraph{Thermal datasets:}
\textbf{LLVIP}~\citep{jia2021llvip} dataset consists of RGB image and thermal (infrared low-light) image pairs captured with fixed cameras observing outdoor street scenes.
The train set comprises 12,025 RGB and thermal pairs, while the test set contains 3,463 pairs
We follow ImageBind to convert the original detection dataset into a binary classification dataset. 
This process involves cropping out pedestrians from bounding boxes and creating random bounding boxes to ensure a balanced distribution of samples for the categories `person' and `background', each containing about 8K samples.
For evaluation, we use the following class names for the `person' class - [`person', `man', `woman', `people'], and [`street', `road, `car', `light', `tree'] for the `background' class.
\textbf{FLIR\_v2}~\citep{flirv2kaggle} dataset consists of a total of 9,711 thermal and 9,233 RGB images for object detection, with a 90\%/10\% train/validation split. 
We utilize paired RGB-thermal images and crop images of each category based on bounding boxes to prepare them for a classification task. Categories with very few instances are removed, and the dataset is filtered to retain the following 10 categories: 
[`car', `person', `sign', `motor', `truck', `light', `bike', `hydrant', `dog', `other vehicle'].

\paragraph{Tactile datasets:}
The tactile data is sourced from \textbf{Touch-and-go}~\citep{yang2022touch_and_go} dataset, which includes tactile data collected by human data collectors probing objects in natural environments using tactile sensors. Simultaneously, egocentric videos are recorded to obtain the corresponding images.
The dataset includes annotations for 20 different material classes and offers labels for hard/soft (H/S) and rough/smooth (R/S) characteristics. This results in three versions of the tactile dataset: one for material classification (\textbf{TAG-M}), one for hard/soft classification (\textbf{TAG-H/S}), and one for rough/smooth classification (\textbf{TAG-R/S}).
Following ViT-Lens, our model is trained using the TAG-M dataset and evaluated on all three datasets.

\paragraph{Electroencephalography (EEG) datasets:}
For the visual concept classification task, the \textbf{ImageNet-EEG (IN-EEG)}~\citep{spampinato2017eeg} dataset is utilized, consisting of EEG recordings obtained from six human subjects using a 128-channel human brain activity recording system. 
Each subject is exposed to 2,000 images from 40 categories sourced from the ImageNet~\citep{russakovsky2015imagenet} dataset. With each category comprising 50 unique images, a total of 12,000 EEG sequences are recorded.
We conducted experiments on a test set of 1,997 samples to compare with ImageBind. Additionally, we utilized both the validation set (1,998 samples) and the test set to compare with ViT-Lens.

\paragraph{3D Point cloud datasets:}
For the 3D classification task, we use \textbf{ShapeNet55}~\citep{wu2015shapenets} for training and \textbf{ModelNet40}~\citep{wu2015modelnet40} for evaluation. 
Following VitLens~\citep{lei2024vitlens}, we also employ the processed ShapeNet55 dataset. 
This dataset consists of approximately 52.5K 3D point clouds generated from CAD models, along with corresponding images created using virtual cameras and text data obtained by filling metadata into a predefined template.
The ModelNet40 dataset is a popular benchmark for 3D shape classification.
This dataset contains 12,311 CAD models across 40 categories, with 9,843 training samples and 2,468 testing samples. 
Each object is represented as a 3D point cloud and manually annotated with its category, including everyday items like chairs, tables, desks, and household objects. 
We evaluate our model's performance in 3D classification using only the test set in our experiments.

\subsection{Training}
\label{supp: training details}

\begin{table*}[htb]
\centering
\scriptsize
\setlength{\tabcolsep}{3pt} 

\newcolumntype{Y}{>{\centering\arraybackslash}X}

\begin{tabularx}{\textwidth}{ll|Y Y Y Y Y Y Y} \toprule 
\multicolumn{2}{l|}{Modality} 
& \makecell{\includegraphics[width=0.3\linewidth]{fig/icon/image.png} \\ \textbf{Image}} 
    & \makecell{\includegraphics[width=0.3\linewidth]{fig/icon/video.png} \\ \textbf{Video}} 
    & \makecell{\includegraphics[width=0.3\linewidth]{fig/icon/depth.png} \\ \textbf{Depth}} 
    & \makecell{\includegraphics[width=0.3\linewidth]{fig/icon/audio.png} \\ \textbf{Audio}} 
    & \makecell{\includegraphics[width=0.3\linewidth]{fig/icon/thermal.png} \\ \textbf{Thermal}} 
    & \makecell{\includegraphics[width=0.3\linewidth]{fig/icon/tactile.png} \\ \textbf{Tactile}} 
    & \makecell{\includegraphics[width=0.3\linewidth]{fig/icon/eeg.png} \\ \textbf{EEG}} \\ \midrule

\multicolumn{2}{l|}{Batch Size} & \makecell{IN1K: 16 \\ p365: 10} & 6 & 8 & 6 & 8 & 6 & 6 \\ \midrule

\multirow{6}{*}{Loss $\lambda$} 
    & $\mathcal{L}_{\text{vq}}$    & 10 & 10 & 1000 & 50 & 1000 & 1000 & 10 \\ 
    & $\mathcal{L}_{\text{cctr}}, \mathcal{L}_{\text{cuni}}$  & 0.01 & 0.01 & 0.01 & 0.01 & 0.05 & 0.01 & 0.01 \\ 
    & $\mathcal{L}_{\text{cm}}$  & 0.05 & 0.05 & 0.1 & 1.0 & 1.0 & 0.1 & 0.1 \\ 
    & $\mathcal{L}_{\text{orth}}$  & 0.1 & 0.1 & 0.1 & 0.1 & 1.0 & 0.05 & 0.05 \\ 
    & $\mathcal{L}_{\text{uni}}$   & 1.0 & 1.0 & 1.0 & 1.0 & 1.0 & 0.1 & 0.1 \\ 
    & $\mathcal{L}_{\text{recon}}$   & 1.0 & 1.0 & 1.0 & 1.0 & 1.0 & 1.0 & 1.0 \\ 
    \hline

\multicolumn{2}{l|}{LoRA (Layers)} & -- & -- & 7--12 & 7--12 & 7--12 & 7--12 & 4--6 \\

\multicolumn{2}{l|}{\# Trainable Params.} & 32.4M & 29.9M & 59.6M & 56.7M & 61.6M & 58.5M & 34.1M \\ \bottomrule
\end{tabularx}

\vspace{-0.1cm}
\caption{\textbf{Training Hyper-parameters across Modalities in CodeBind-IB.} This table details the batch size per GPU, loss function weights ($\lambda$), LoRA inserted modality encoder layers, and the number of trainable parameters for aligning bridging modality to each target modality. Modality encoders of depth, audio, thermal, and tactile have 12 transformer blocks in total, while modality encoder of EEG has 6 blocks in total. Modality encoder of image and video is frozen and LoRA is disabled.
}
\vspace{-0.3cm}
\label{tab_supp_hyperparams}
\end{table*}

\begin{table}[htbp]
\centering
\small
\begin{tabular}{lccc}
\toprule
 & SUN-D & NYU-D & FLIR\_v2 \\ \midrule
manual loss weights & 45.7 & 59.3 & 97.2 \\ \midrule
adaptive loss weights & 46.0 & 57.5 & 97.2 \\ \bottomrule
\end{tabular}
\caption{\textbf{Comparison between manual and adaptive loss weights across different datasets.}}
\vspace{-0.2cm}
\label{tab_supp_ada_loss}
\end{table}

CodeBind-IB and CodeBind-VL share similar training hyperparameters with ImageBind and Vit-Lens, respectively. 
Bridging modalities are paired with one target modality at a time to train a unique set of shared and specific codebooks. Each bridging–target modality pair uses a distinct set of codebooks (\eg image-text-audio vs. image-text-depth), and codebooks are not unified or fixed across different target modalities.
Different batch sizes are used for different modalities to optimize GPU memory utilization.
Additionally, different loss functions are assigned varying weights depending on the modalities to be aligned to maintain a balanced loss composition. 
Details of these hyperparameters and the number of trainable parameters for CodeBind-IB are in Tab \ref{tab_supp_hyperparams}.
We further implement an adaptive loss balancing strategy to dynamically assign weights for $\mathcal{L}_{vq}, \mathcal{L}_{cctr}, \mathcal{L}_{cuni}, \mathcal{L}_{cm}, \mathcal{L}_{orth}, \mathcal{L}_{uni}$ based on their relative magnitudes with respect to $\mathcal{L}_{align}$. By default, the weights for $\mathcal{L}_{align}$ and $\mathcal{L}_{recon}$ are fixed at 1. Table~\ref{tab_supp_ada_loss} presents an ablation study comparing this adaptive approach against the predefined hyperparameter configurations detailed in Table~\ref{tab_supp_hyperparams}. Notably, the adaptive strategy yields performance comparable to our manually tuned settings across both depth and thermal modalities. These results demonstrate the robustness and scalability of our framework, as it effectively accommodates low-resource modalities while eliminating the burden of exhaustive hyperparameter searching.

\section{Additional Experiment Results}
\label{sec: exp_supp}

\subsection{Analysis of Codebook's Feature Space}
\label{supp: codebook feature space analysis}

\begin{figure}[tb]
\centering
\includegraphics[width=0.98\linewidth]{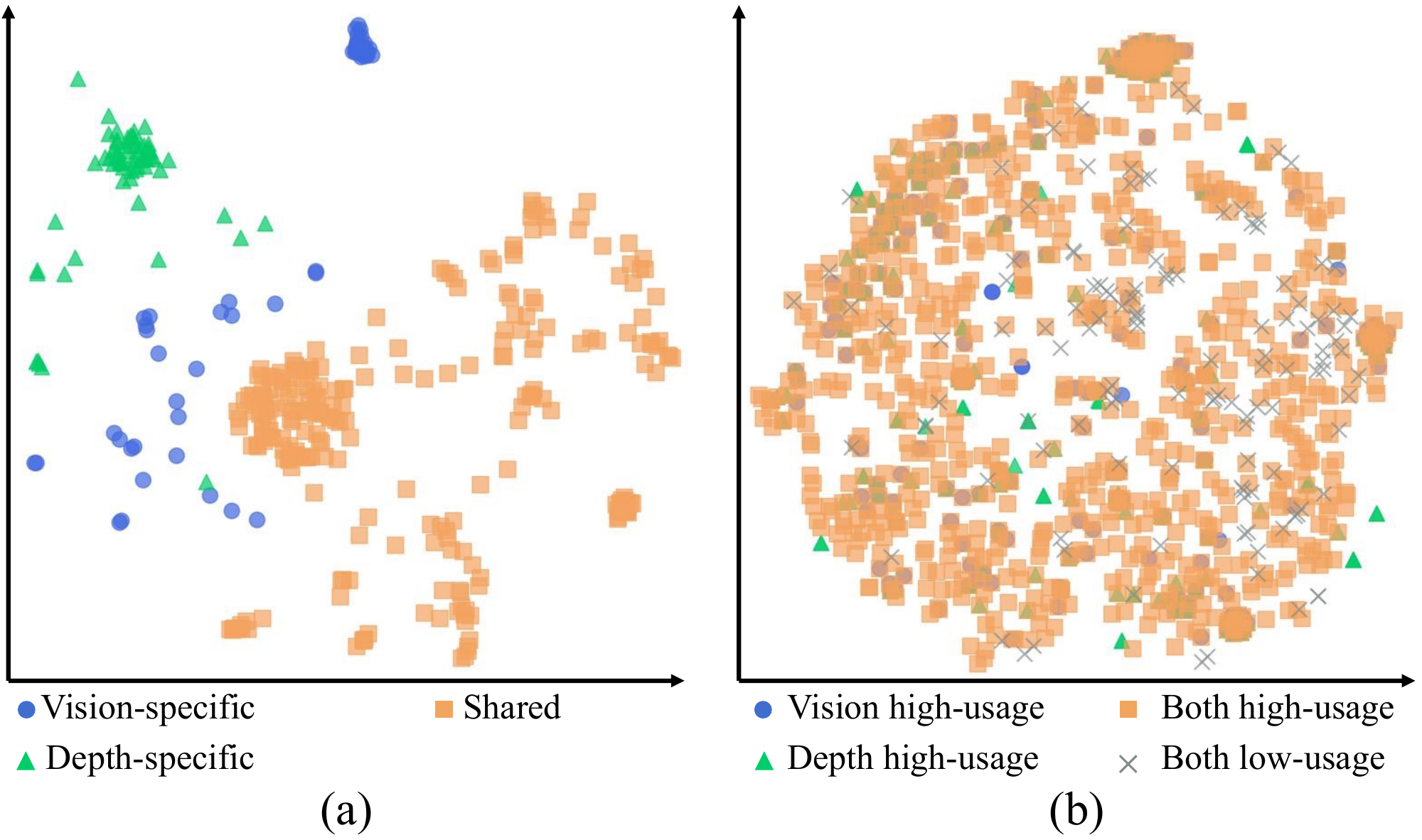}
\caption{
(a) \textbf{Distribution of codevectors} from the shared, vision-specific, and depth-specific codebooks.
(b) \textbf{Distribution and usage rates of codevectors} in the shared codebook for shared embeddings from vision and depth modalities.
}
\vspace{-0.6cm}
\label{fig_supp_codebook_distribution}
\end{figure}

\begin{figure}[tb]
\centering
\includegraphics[width=0.98\linewidth]{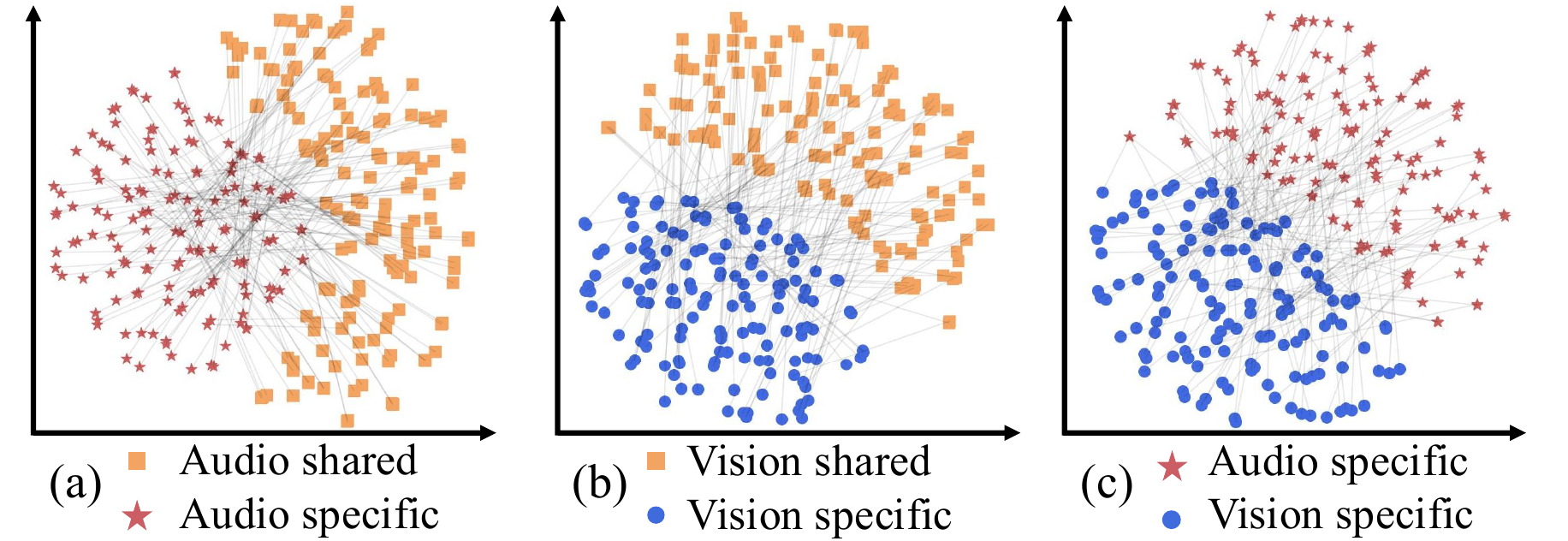}
\vspace{-0.1cm} 
\caption{
\textbf{2D t-SNE visualization of sampled embeddings from AudioSet~\cite{gemmeke2017audioset}.} 
}
\vspace{-0.2cm}
\label{fig_supp_embedding_info}

\end{figure}

\begin{figure}[tb]
\centering
\includegraphics[width=0.99\linewidth]{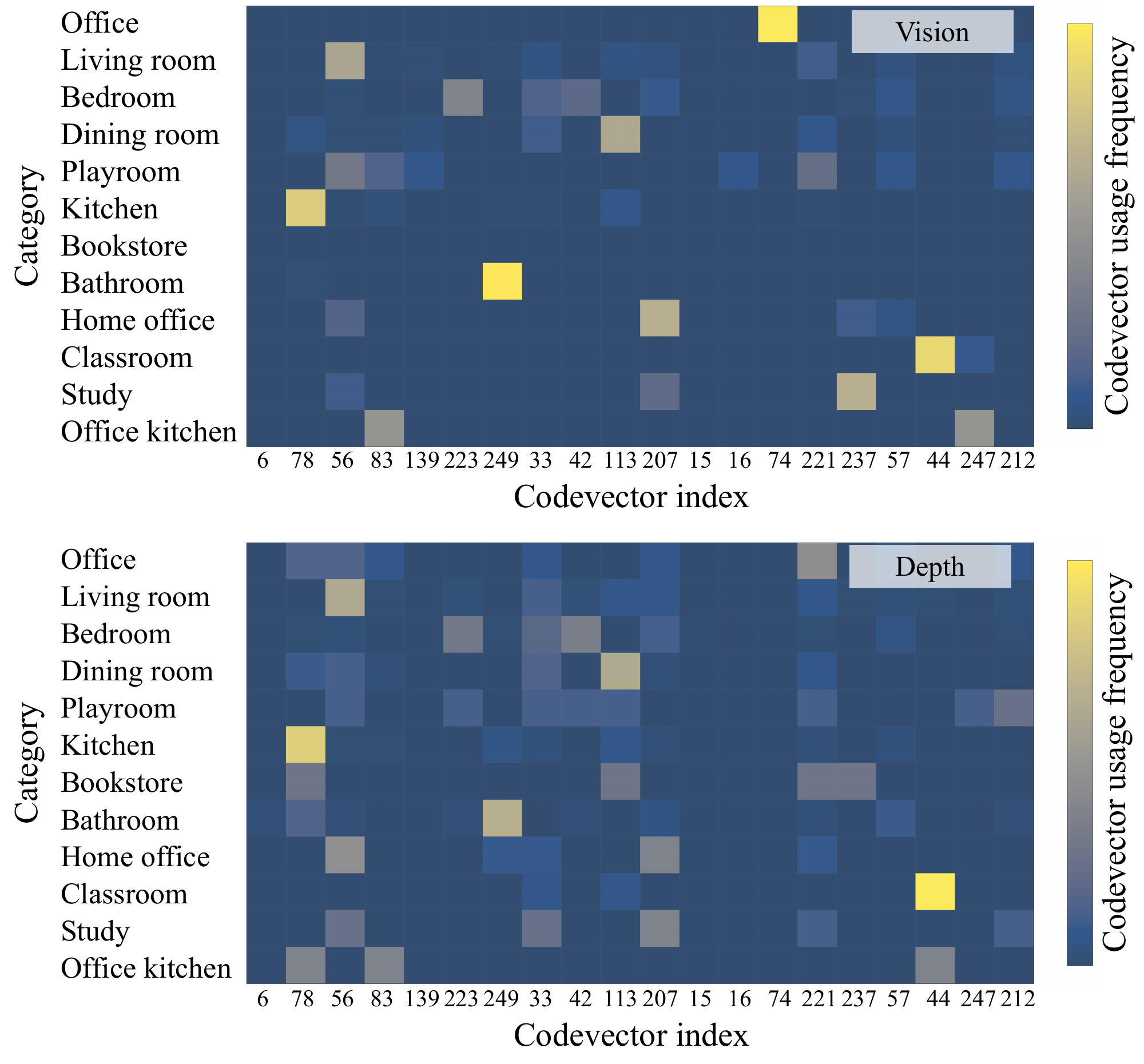}
\caption{
\textbf{Visualization of codevectors usage frequency distributions} of image-depth pairs in NYU-D dataset among various categories. 
Similar distribution patterns across two modalities indicate semantic consistency in our shared codebook.
}
\vspace{-0.2cm}
\label{fig_supp_codebook_class_wise_usage}
\end{figure}

\paragraph{Codebook distribution} 
We utilize t-SNE~\citep{van2008visualizing} to visualize the shared codebook and the specific codebooks from different modalities.
We use a conventional codebook design to facilitate observation, where each codevector has the same dimension as the input embedding. 
Specifically, the shared codebook contains 256 codevectors, while specific codebook for each modality contains 64 codevectors.
As shown in Fig.~\ref{fig_supp_codebook_distribution}(a), the varying distributions of specific and shared codebooks highlight modality-specific distinctions and their complementary roles in the feature space. This is consistent with observations in Fig.~\ref{fig_supp_embedding_info}. Furthermore, t-SNE visualizations (Fig.~\ref{fig_supp_embedding_info}) of 384 samples from AudioSet~\citep{gemmeke2017audioset}, extracted via random batch sampling, reaffirm the pattern of decoupled representations. Consistent with previous analyses, these 2D projections contrast our compositional codebook with ImageBind~\cite{girdhar2023imagebind} to demonstrate superior modal separation.

\paragraph{Shared codebook usage across modalities}
We visualize the distribution of codevectors within the shared codebook, color them by modality, and mark them based on their usage rate. The usage rate of codevectors is calculated in the validation set of the depth-image paired dataset NYU-D~\citep{silberman2012nyu}, categorizing those with usage rates above 75\% as high-usage and below 10\% as low-usage. The usage rates of all codevectors in the shared codebook are then aggregated across different modalities (\ie vision and depth).
In Fig.~\ref{fig_supp_codebook_distribution}(b), it is evident that most codevectors are used by both modalities, indicating effective utilization of the shared codebook. A small proportion of codevectors, however, are primarily used by only one modality, suggesting some remaining differences in representations between the modalities. Additionally, a few codevectors exhibit low usage rates in either modality, reflecting the presence of less common information due to uneven sample distributions.

\paragraph{Codevector-level alignment} 
We further investigate the alignment of shared embeddings from different modalities (\eg image and depth) at the codevector level. 
First, the usage rate of codevectors in the shared codebook is calculated based on the shared embeddings from paired data samples within each category. Then, the top-20 most frequently used codevectors are selected, and their usage rates are standardized across categories, with their distribution visualized using different colors in Fig.~\ref{fig_supp_codebook_class_wise_usage}.
The paired rows, where distributions from two modalities within the same category exhibit similar patterns of highlighted grids, demonstrate the fine-grained alignment between the shared embeddings from two modalities.
In paired columns representing different modalities, similar patterns of highlighted grids can be noticed. This suggests that our learned shared codevectors capture modality-invariant semantic information. 
Distinct patterns are evident between columns at different positions, indicating that each codevector captures unique semantic information.

\subsection{Analysis of Specific Embeddings} 
\label{supp: specific embedding analysis}

\begin{figure}
\begin{subfigure}{1.0\linewidth}
  \centering
  \includegraphics[width=.98\linewidth]{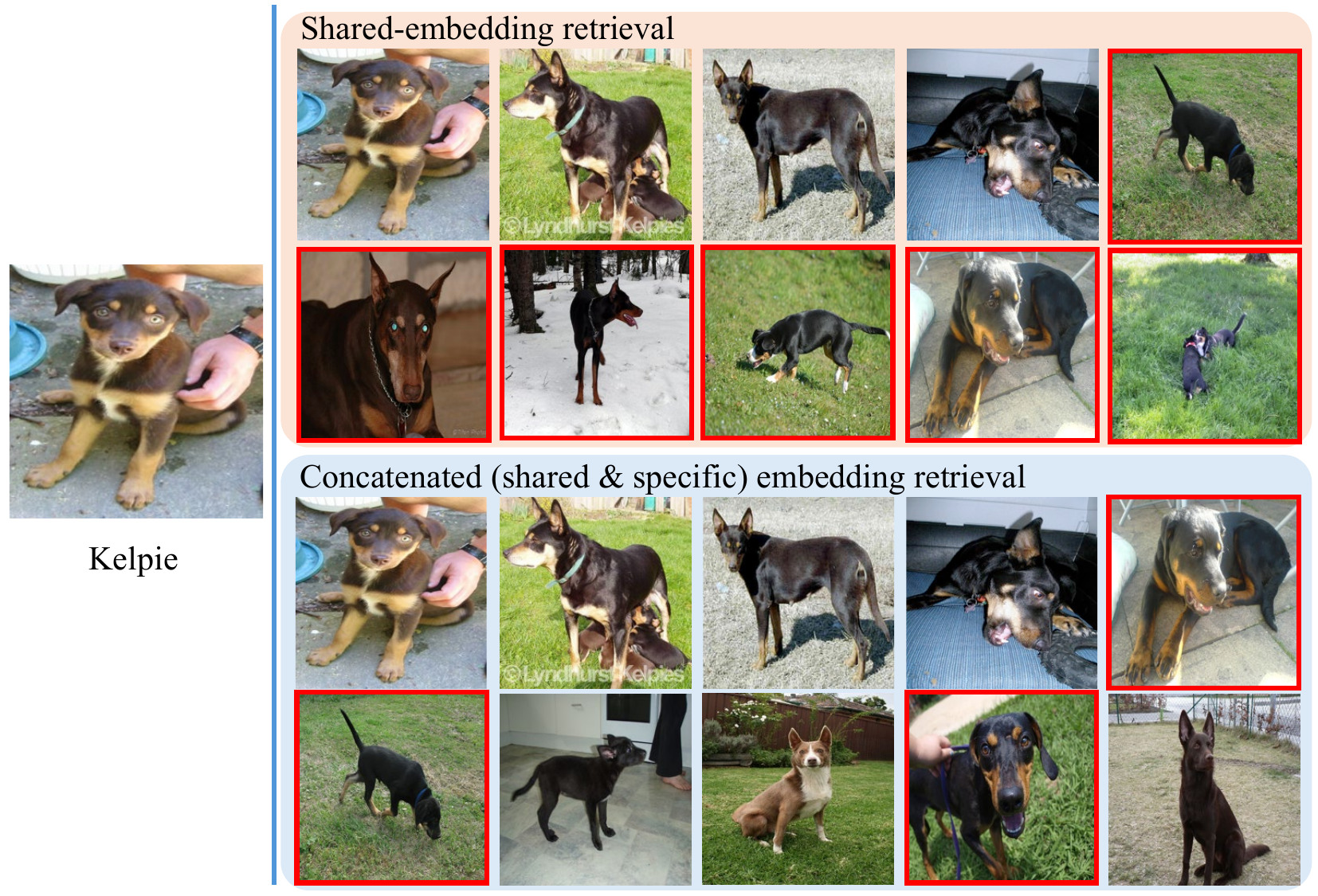}  
  \caption{Examples in Stanford Dogs dataset.}
\end{subfigure}

\begin{subfigure}{1.0\linewidth}
  \centering
  \includegraphics[width=.98\linewidth]{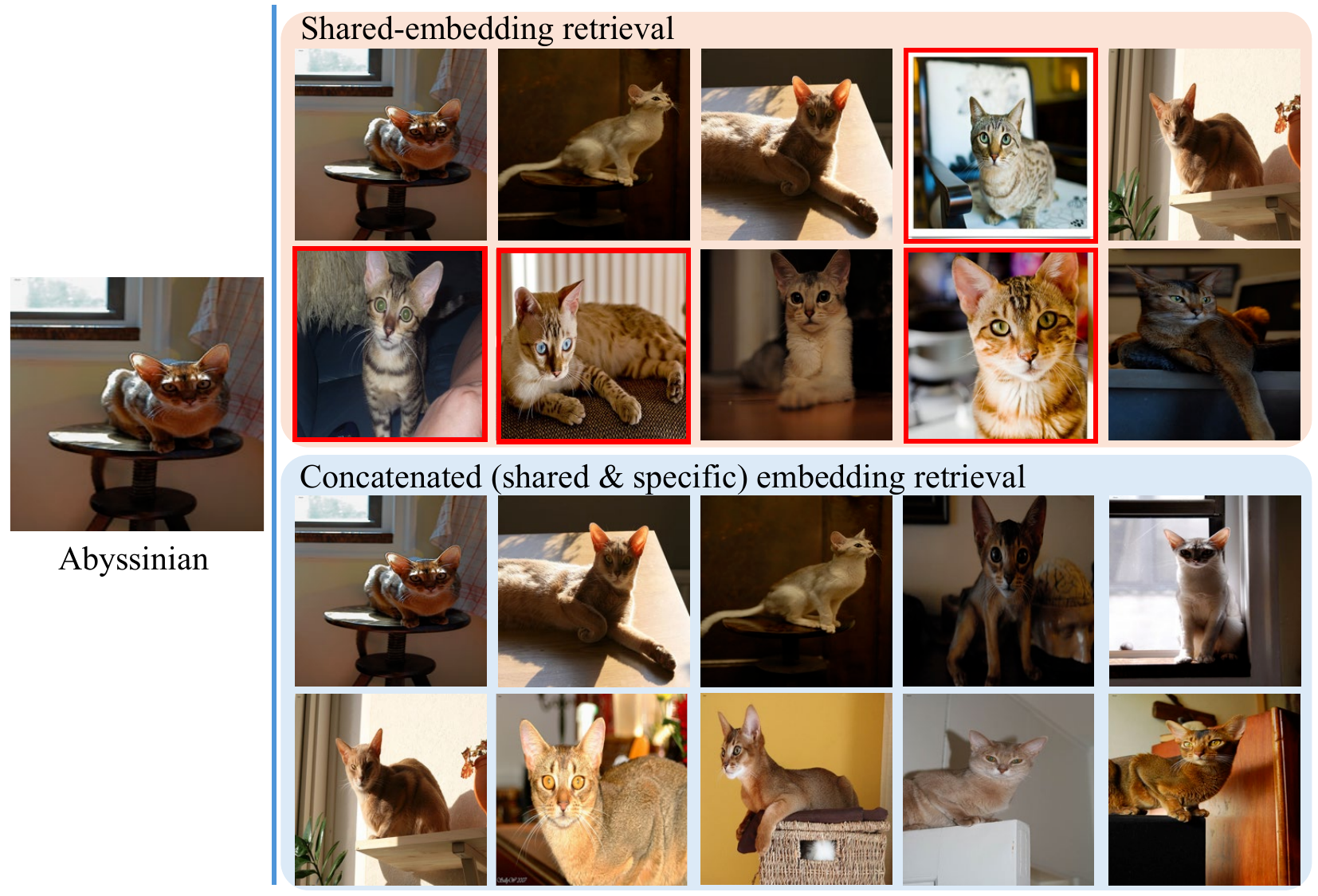}  
  \caption{Examples in cat images in Oxford-IIIT Pet dataset.}
\end{subfigure}

\begin{subfigure}{1.0\linewidth}
  \centering
  \includegraphics[width=.98\linewidth]{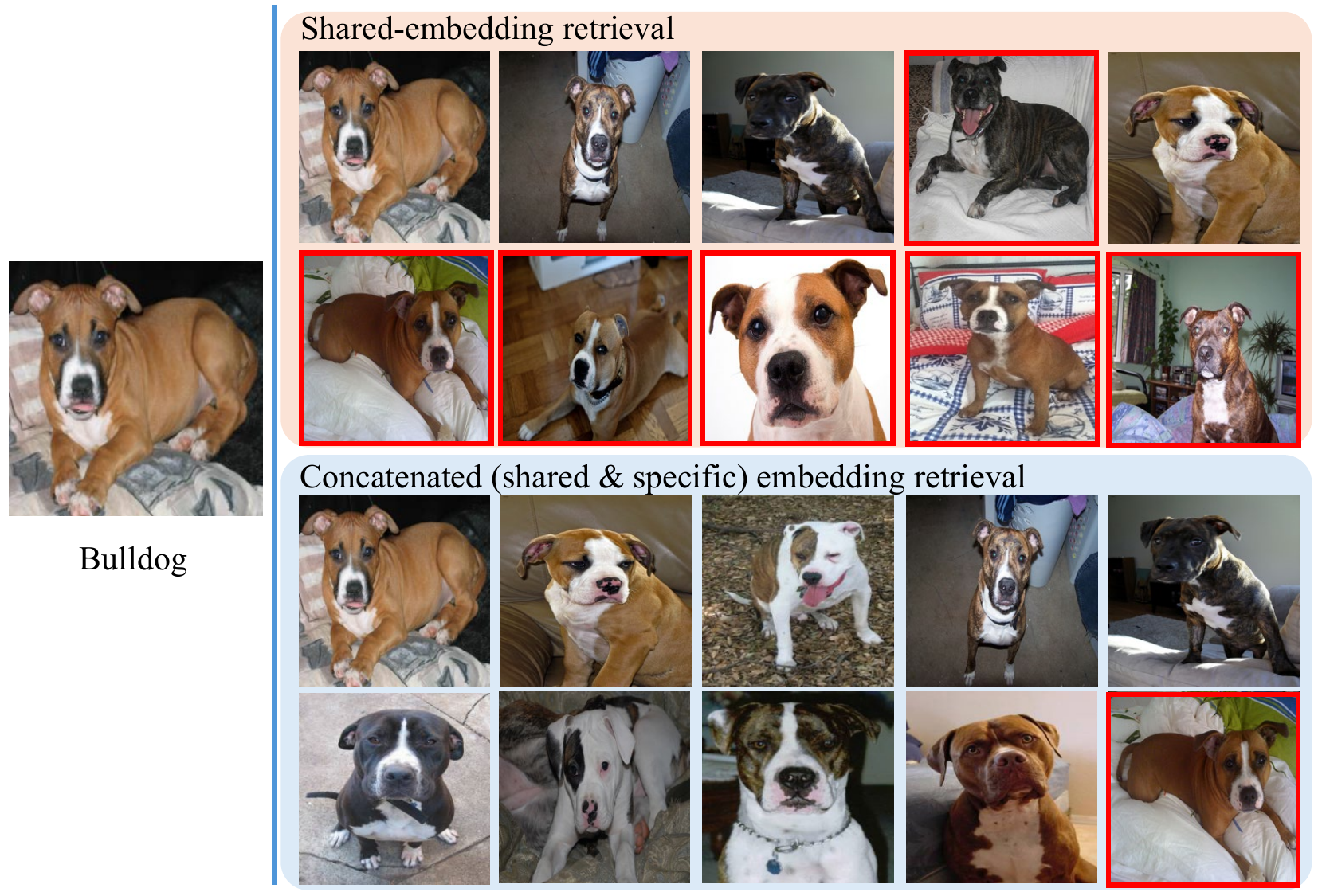}  
  \caption{Examples in dog images in Oxford-IIIT Pet dataset.}
\end{subfigure}

\caption{\textbf{Additional results for fine-grained retrieval.} By utilizing the concatenation of shared and specific embeddings, our method retrieves more correct images featuring the same cat or dog breed, outperforming scenarios that rely solely on shared embeddings.}
\vspace{-0.2cm} 
\label{fig_supp_fine_retrive}
\end{figure}

\newcolumntype{Y}{>{\centering\arraybackslash}X}
\begin{table}[htbp]
\centering
\small 
\begin{tabularx}{\columnwidth}{c Y}
\toprule
\multicolumn{1}{c}{\makecell{Fine-grained \\ Attribute}} & \multicolumn{1}{c}{Specific Categories} \\ \midrule
\multirow{2}{*}{\makecell{Lighting \\ illumination}} & natural diffused, artificial directional \\
 & ambient even, harsh shadowed \\ \midrule
\multirow{2}{*}{\makecell{Camera \\ geometry}} & eye level close up, elevated wide angle, \\
 & ground level perspective, top down view \\ \midrule
\multirow{2}{*}{\makecell{Texture \\ surface}} & smooth fine, coarse grainy, \\
 & patterned structured, mixed textures \\ \midrule
\multirow{2}{*}{\makecell{Scene \\ environment}} & blurred background, structured setting, \\
 & open space, indoor context \\ \midrule
\multirow{2}{*}{\makecell{Color \\ tone}} & warm natural, cool balanced, \\
 & neutral muted, vibrant saturated \\ \bottomrule
\end{tabularx}
\caption{
\textbf{Fine-grained labels derived by VLM for linear probing.}
}
\vspace{-0.2cm}
\label{tab_supp_prob_class}
\end{table}

\newcolumntype{Y}{>{\centering\arraybackslash}X}
\begin{table}[htb]
\centering
\small
\begin{tabularx}{\columnwidth}{c Y}
\toprule
Attribute & NMI with category \\ \midrule
lighting illumination & 0.2093 \\
camera geometry       & 0.1907 \\
texture surface       & 0.3145 \\
scene environment     & 0.3303 \\
color tone            & 0.2759 \\ \midrule
Average NMI   & 0.2641 \\ \bottomrule
\end{tabularx}
\vspace{-0.1cm}
\caption{\textbf{Mutual information analysis between category labels and fine-grained labels.} Normalized Mutual Information (NMI) between the VLM-curated fine-grained attributes and semantic class names on samples from ImageNet~\citep{russakovsky2015imagenet}. The low average NMI (0.2641) indicates that the extracted physical nuances are significantly disentangled from category-level semantics.}
\vspace{-0.2cm}
\label{tab_supp_prob_nmi}
\end{table}

\begin{figure}[t]
\centering
\includegraphics[width=1.0\linewidth]{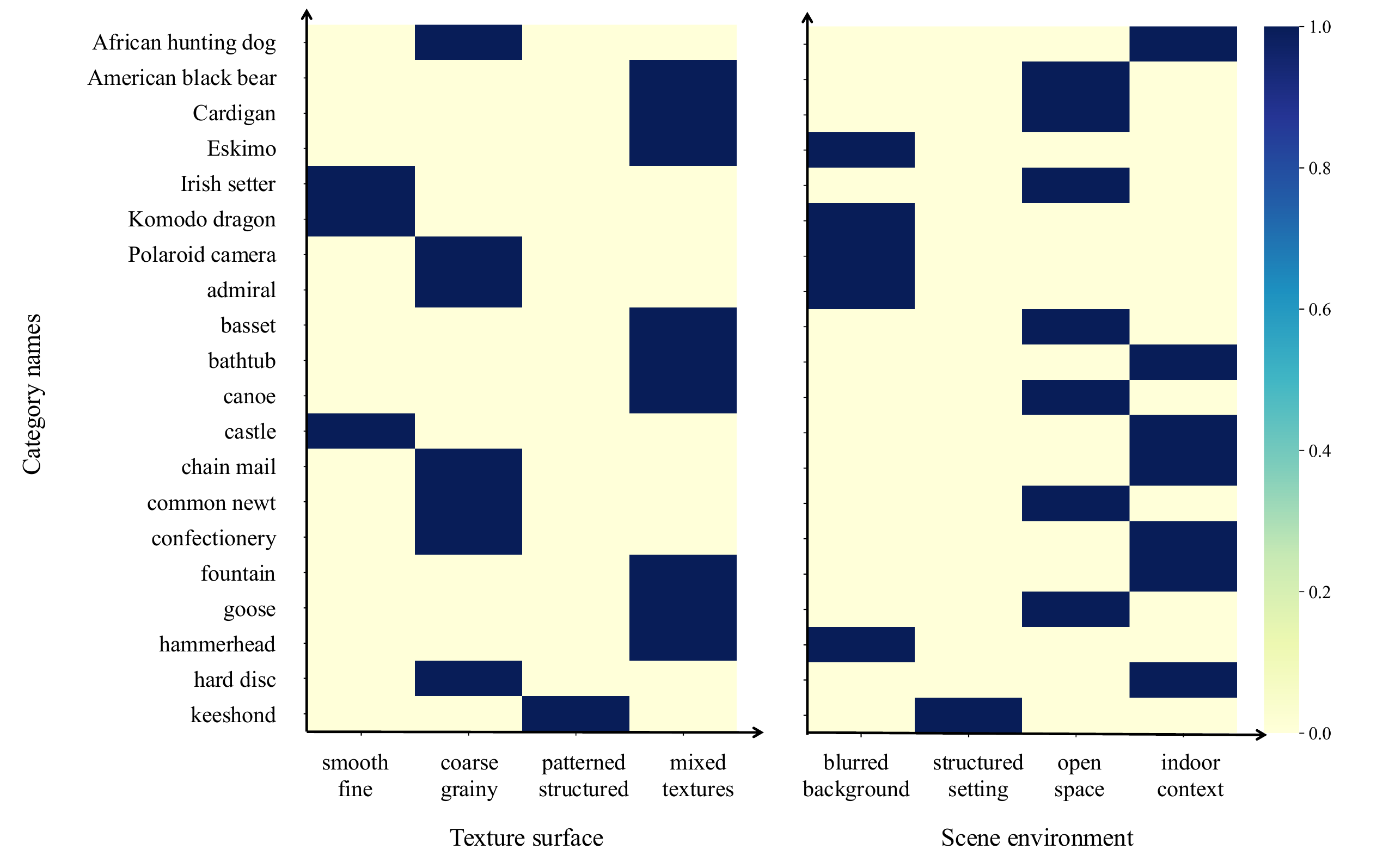}
\vspace{-0.5cm} 
\caption{
\textbf{Heatmap between fine-grained attributes and category names.} Texture surface and scene environment are selected fine-grained attributes summarised by VLM. 20 categories are selected from ImageNet1K~\cite{russakovsky2015imagenet} for display.
}
\vspace{-0.3cm}
\label{fig_supp_prob_nmi}
\end{figure}

\paragraph{Fine-grained intra-modality retrieval}
\label{supp: fine-grained intra-modality retrieval}

We observe that 61.7\% (3699/6000), 84.6\% (2005/2371), and 90.3\% (4496/4978) of retrieval samples achieved higher or equivalent top-10 recall using concatenated embeddings rather than shared embeddings on the Stanford Dogs and Oxford-IIIT Pet datasets (cats and dogs, respectively). Fig.~\ref{fig_supp_fine_retrive} provides additional examples with corresponding retrieved images where this improvement is evident.

\paragraph{Linear probing for shared and specific embeddings}
\label{supp: linear probing}
We use Qwen2.5-VL~\citep{Qwen2.5VL}, a vision-language model (VLM), to annotate samples from 1000 images from ImageNet through a systematic three-stage approach. (1) Detailed description generation: The VLM generates exhaustive descriptions of fine-grained features for sampled images. The prompts are specifically engineered to isolate distinctive details (as shown in Tab.~\ref{tab_supp_linear_prob}), ensuring these features are decoupled from the common category-level information used in cross-modal sharing. 
(2) Keyword extraction: The VLM analyzes the detailed descriptions from Stage 1 to extract four key terms across five major fine-grained dimensions. This stage distills raw text into structured keywords optimized for classification. The keywords for each fine-grained attribute are shown in Fig.~\ref{tab_supp_prob_class}
(3) Fine-grained image labeling: Using the finalized keywords as a reference, the VLM performs the final annotation for each image, mapping specific visual features to the corresponding keywords.

Tab.~\ref{tab_supp_prob_nmi} presents the Normalized Mutual Information scores (NMI) between semantic category labels and our VLM-derived fine-grained physical labels. Additionally, Fig.~\ref{fig_supp_prob_nmi} visualizes the conditional probability distribution, $P(\text{physical attribute} | \text{category label})$, via a normalized heatmap. The analysis reveals a notable association with an average NMI score of $0.26$, indicating that semantic categories account for approximately 26\% of the variance in fine-grained physical attributes. This observation exemplifies the natural coupling inherent in real-world datasets. For specific classes, fine-grained features, such as background color or unique textures, serve as physical fingerprints that are strongly correlated with semantic identities (\eg a ``polar bear'' is statistically intertwined with a ``snowy background''). 
Consequently, specific embeddings capture these correlated physical priors, which enable them to support category discrimination as effectively as shared embeddings. 
Nevertheless, the heatmap demonstrates that while certain species exhibit strong ties to specific attributes, most physical properties maintain a broad distribution across the entire dataset. 
This suggests that our VLM-derived labels capture critical variance that semantic categories alone cannot account for. 
Through our decoupled architecture, specific embeddings exhibit superior sensitivity in distinguishing these physical properties compared to their shared counterparts, particularly in terms of convergence speed and discriminative precision.

\begin{table}[htbp]
\centering
\small
{
\begin{tabular}{p{0.9\columnwidth}}
\toprule
\texttt{
This video belongs to the category `\{c\_name\}'. Describe the appearance and action of \{c\_name\} in one short sentence. Focus on the segment from \{s\_t\}s to \{e\_t\}s. Use direct subject-verb-object structure. No filler words. Max 15 words. Do not use introductory phrases like `In this video'.
} \\
\bottomrule
\end{tabular}
}
\vspace{-0.1cm}
\caption{
\textbf{Text prompt for VLM-based dense text caption.} Category names (c\_name), and time minutes (s\_t, e\_t) are extracted from AVE dataset annotations.}
\vspace{-0.2cm}
\label{tab_supp_dense_caption}
\end{table}

\paragraph{Multimodal fusion on AVE event classification}
\label{supp: multimodal fusion}
We employ Qwen3-VL~\citep{bai2025qwen3} to generate dense textual captions by conditioning on the event category names and their corresponding video frames, using the prompt template detailed in Tab.~\ref{tab_supp_dense_caption}. Text embeddings are obtained via text encoder by using the category names and text captions, respectively, for multimodal alignment training to align with vision and audio representations. Post-training, the modality encoders and their respective codebooks are frozen. A single-layer MLP classification head is then trained on the fixed embeddings to predict event categories. We evaluate three embedding configurations: (1) shared embeddings in isolation, (2) concatenation of video and audio features, and (3) their summation. For the latter two fusion strategies, modality-specific embeddings are concatenated with the shared embeddings for each respective modality. The classification head is optimized with a learning rate of $5 \times 10^{-3}$ over 10 epochs.

\subsection{Ablation on Codebook Size and Dimension}
\label{sec: codebook ablation supp}

\begin{table}[htbp]
\centering
\small

\begin{tabularx}{\columnwidth}{cc|cc}
\toprule
\multicolumn{2}{c|}{\textbf{Codebook Config.}} & \multicolumn{2}{c}{\textbf{Accuracy (\%)}} \\ 
{\small Number ($K$)} & {\small Dimension ($d$)} & {\small SUN-D} & {\small FLIR\_v2} \\ \midrule
1024 & 1024 & 40.7 & 81.1 \\ 
1024 & 512  & 41.9 & 90.5 \\ 
1024 & 128  & 44.3 & 92.5 \\ 
1024 & 32   & 45.4 & 93.0 \\ 
1024 & 8    & 45.7 & 97.2 \\ \bottomrule
\end{tabularx}

\vspace{-0.1cm}
\caption{
\textbf{Effectiveness of compositional codebook with varying codevector dimensions.} 
Results show improved effectiveness as codevector dimension decreases, indicating enhanced performance with richer combinations. } 
\vspace{-0.2cm}
\label{tab_supp_codebook_diff_dim}
\end{table}

\begin{table}[htbp]
\centering
\small

\newcolumntype{Y}{>{\centering\arraybackslash}X}

\begin{tabularx}{\columnwidth}{c|YY}
\toprule
\textbf{Codebook Config.} & \multicolumn{2}{c}{\textbf{FLIR\_v2 Accuracy (\%)}} \\ \cmidrule{2-3}
\textbf{Number ($K$)} & {\small Conventional (dim=1024)} & {\small Compositional (dim=8)}\\ \midrule
64   & 58.6 & 86.2 \\ 
128  & 65.4 & 91.2 \\ 
256  & 76.6 & 94.0 \\ 
512  & 81.9 & 93.2 \\ 
1024 & 81.1 & 97.2 \\ \bottomrule
\end{tabularx}

\vspace{-0.1cm}
\caption{
\textbf{Impact of varying codevector numbers on accuracy.} Results show decreased accuracy with fewer codevectors. Our approach exhibits greater tolerance to this reduction compared to conventional codebooks due to its compositional structure.
}
\vspace{-0.3cm}
\label{tab_supp_codebook_diff_k}
\end{table}

\paragraph{Dimension of codevectors}
We conduct additional experiments to explore the impact of varying the dimensions of codevectors in our compositional codebook on its effectiveness. 
A series of experiments is carried out using a consistent number of codevectors ($1024$) while varying the dimensions, explicitly using codevector dimensions of $\{8, 32, 128, 512, 1024\}$.
Tab.~\ref{tab_supp_codebook_diff_dim} shows that alignment improves as the dimension of the codevector decreases. Reducing the dimension of codevectors increases the number of combinations required to reconstruct an input embedding, thereby enhancing the combinatorial richness of the quantized output and leading to improved performance.

\paragraph{Number of codevectors}
To accommodate more information across various modalities in standard VQ, the conventional approach expands the codebook by introducing more codevectors, leading to a much larger codebook size. In contrast, our method maintains a compact codebook size by leveraging a compositional codebook design, achieving substantially more capacity even with a limited number of codevectors.
We further explore the impact of reducing the number of codevectors while maintaining the same codevector dimension. 
The cross-modal classification performance using conventional codebooks (dim=1024) and our compositional codebook (dim=8) approaches are compared on FLIR\_v2~\citep{flirv2kaggle} dataset, using different numbers of codevectors $\{64, 128, 256, 512, 1024\}$.
As shown in Tab.~\ref{tab_supp_codebook_diff_k}, decreasing the codevector number results in a drop in accuracy.
However, our approach demonstrates greater robustness to this reduction compared to conventional codebooks, owing to its compositional structure. The largest codebook size is limited to 1024, as in this case our partitioned codebook design (detailed in Sec.~\ref{sec:method_vq} in the main paper) allows for an enormous combinatorial capacity of $128^{1024}$ possible sub-vector compositions, which is sufficiently expressive.

\begin{table}[htbp]
\centering
\footnotesize 
\setlength{\tabcolsep}{3pt} 
\begin{tabularx}{\columnwidth}{l|ccc|ccc}
\toprule
\textbf{Config.} & \multicolumn{3}{c|}{\textbf{Usage (\%) $\uparrow$}} & \multicolumn{3}{c}{\textbf{Perplexity $\downarrow$}} \\
& shared & vision & thermal & shared & vision & thermal \\ \midrule
Conv. & 75.6 & 80.9 & 93.8 & 90.8 & 106.7 & 127.3 \\
Comp. & 100 &  100 & 100   & 661.4  & 87.9   & 73.6 \\ \bottomrule
\end{tabularx}
\vspace{-0.1cm}
\caption{
\textbf{Codebook usage and perplexity} with compositional/conventional configurations on FLIR\_v2. The modality-shared codebook size is set to 1024, while the specific codebooks for vision and thermal modalities are set to 256. Results indicate full utilization of both modality-shared and specific codebooks following the implementation of the compositional design.
}
\vspace{-0.3cm}
\label{tab_supp_codebook_usage}
\end{table}

\paragraph{Codebook collapse prevention}
Benefiting from the reinitialization strategy in CVQ~\citep{zheng2023online} and the EMA update method described in Sec.~\ref{supp: codebook update}, the proposed compositional codebook achieves high usage efficiency. As shown in Tab~\ref{tab_supp_codebook_usage}, the shared codebook usage increases from 75.6\% to 100\%, accompanied by a significant increase in perplexity. Although the perplexity of the specific codebook decreases, this is probably due to the relatively small number of data categories compared to the overall codebook capacity.

\begin{figure}[t]
\centering
\includegraphics[width=1.0\linewidth]{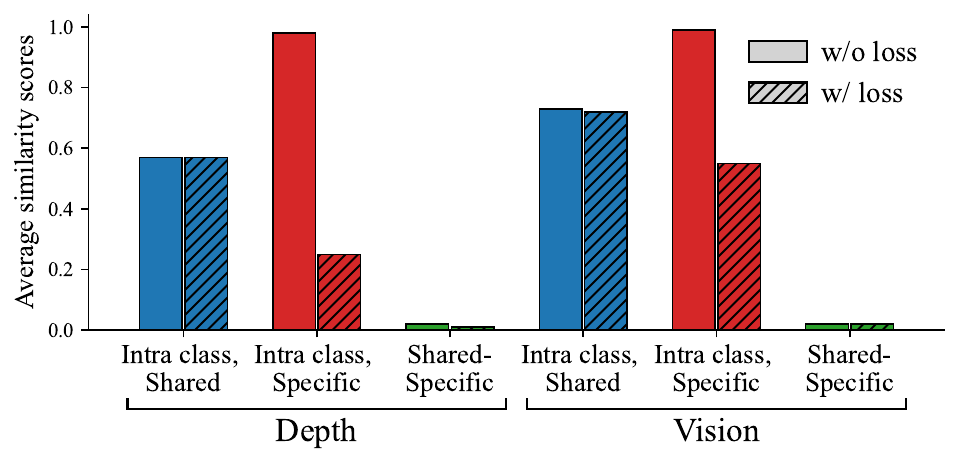}
\vspace{-0.3cm} 
\caption{
\textbf{Visualization of intra-class average similarity scores among shared and specific embeddings.} The intra-class average similarity scores are calculated on SUN-D~\cite{song2015sun}, with and without orthogonal loss $\mathcal{L}_{\mathrm{orth}}$ and uniform loss $\mathcal{L}_{\mathrm{uni}}$. The results demonstrate a substantial reduction in similarity among specific embeddings after applying these losses, indicating that they effectively encourage specific embeddings to encode distinct fine-grained information, while remaining within the same general semantic category, in contrast to shared embeddings.
}
\vspace{-0.3cm}
\label{fig_supp_loss_analysis}
\end{figure}

\subsection{Ablation on Loss Functions}
\label{sec: loss ablation supp}

\paragraph{Orthogonal and uniform loss}
The effectiveness of $\mathcal{L}_{\mathrm{uni}}$ and $\mathcal{L}_{\mathrm{orth}}$ is evaluated on SUN-D~\citep{song2015sun} by analyzing the intra-class average similarity scores of shared and specific embeddings both in image and depth before and after applying these losses. Within each modality, the cosine similarity scores among all shared embeddings (\ie intra-shared embedding similarity), specific embeddings (\ie intra-specific embedding similarity), and paired shared and specific embeddings of the complete modality representation (\ie inter-shared-specific embedding similarity) are computed within each category and then averaged.
In Fig~\ref{fig_supp_loss_analysis}, before applying $\mathcal{L}_{\mathrm{uni}}$, specific embeddings exhibited high similarity within each semantic category, indicating an overlap in encoded information. However, $\mathcal{L}_{\mathrm{uni}}$ significantly reduces the similarity between specific embeddings of depth from 0.98 to 0.25, ensuring that specific embeddings capture fine-grained, modality-specific details rather than aligning with broader semantic categories. Meanwhile, shared embeddings in depth, which encode category-level information aligned with text embeddings, maintain a higher similarity of 0.57 within each class. Additionally, $\mathcal{L}_{\mathrm{orth}}$ further decreases the similarity between shared and specific embeddings in depth from an already low level to 0.02, reinforcing the separation of information, as the two embeddings are initialized in distinct codebook spaces. Similar patterns can be found in the embeddings of vision modality.

\begin{figure*}
\begin{subfigure}{1.0\linewidth}
  \centering
  \includegraphics[width=.85\linewidth]{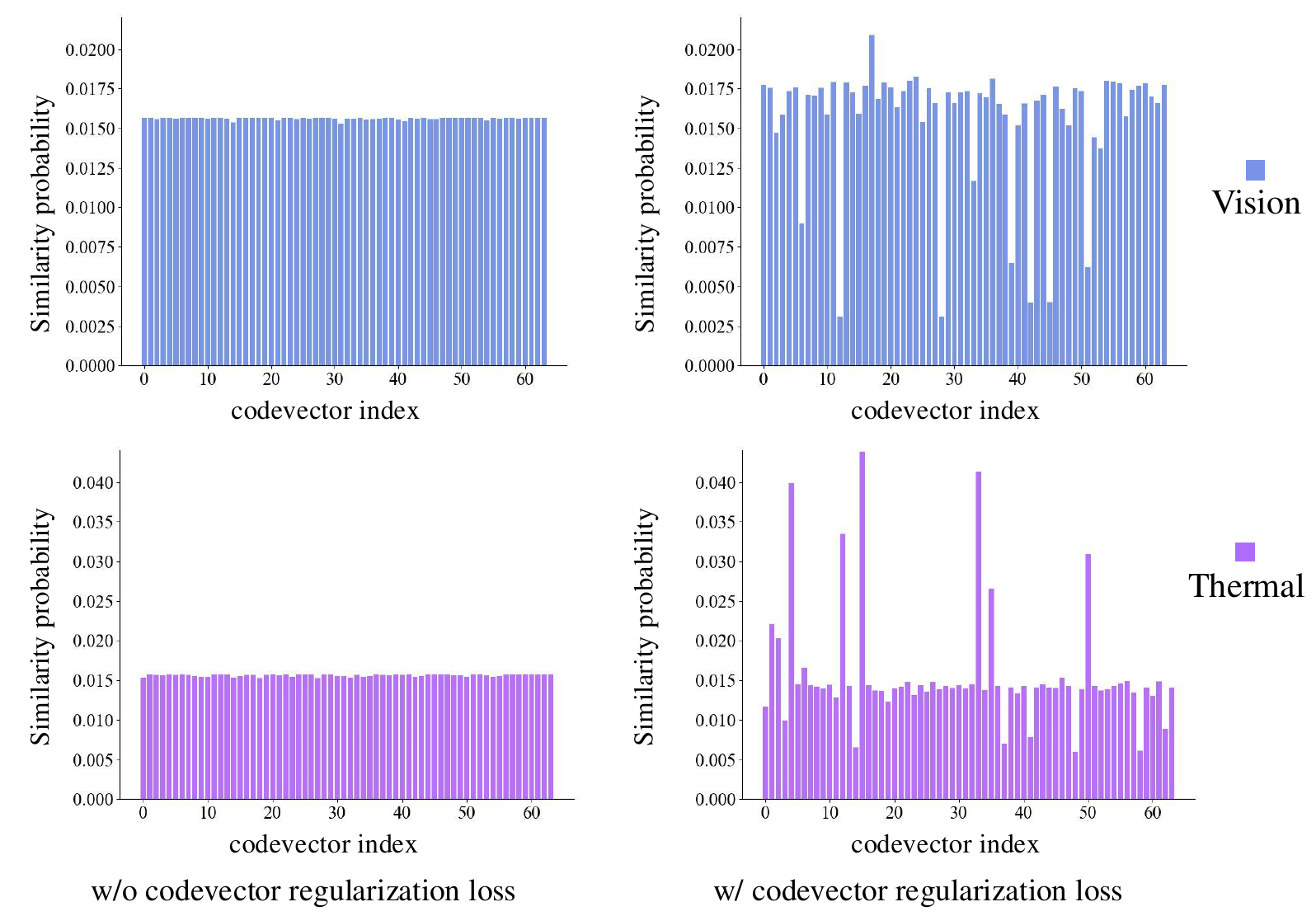}  
  \caption{codevector distribution for specific embeddings selected in FLIR\_v2~\cite{flirv2kaggle}}
\end{subfigure} 

\begin{subfigure}{1.0\linewidth}
  \centering
  \includegraphics[width=0.85
  \linewidth]{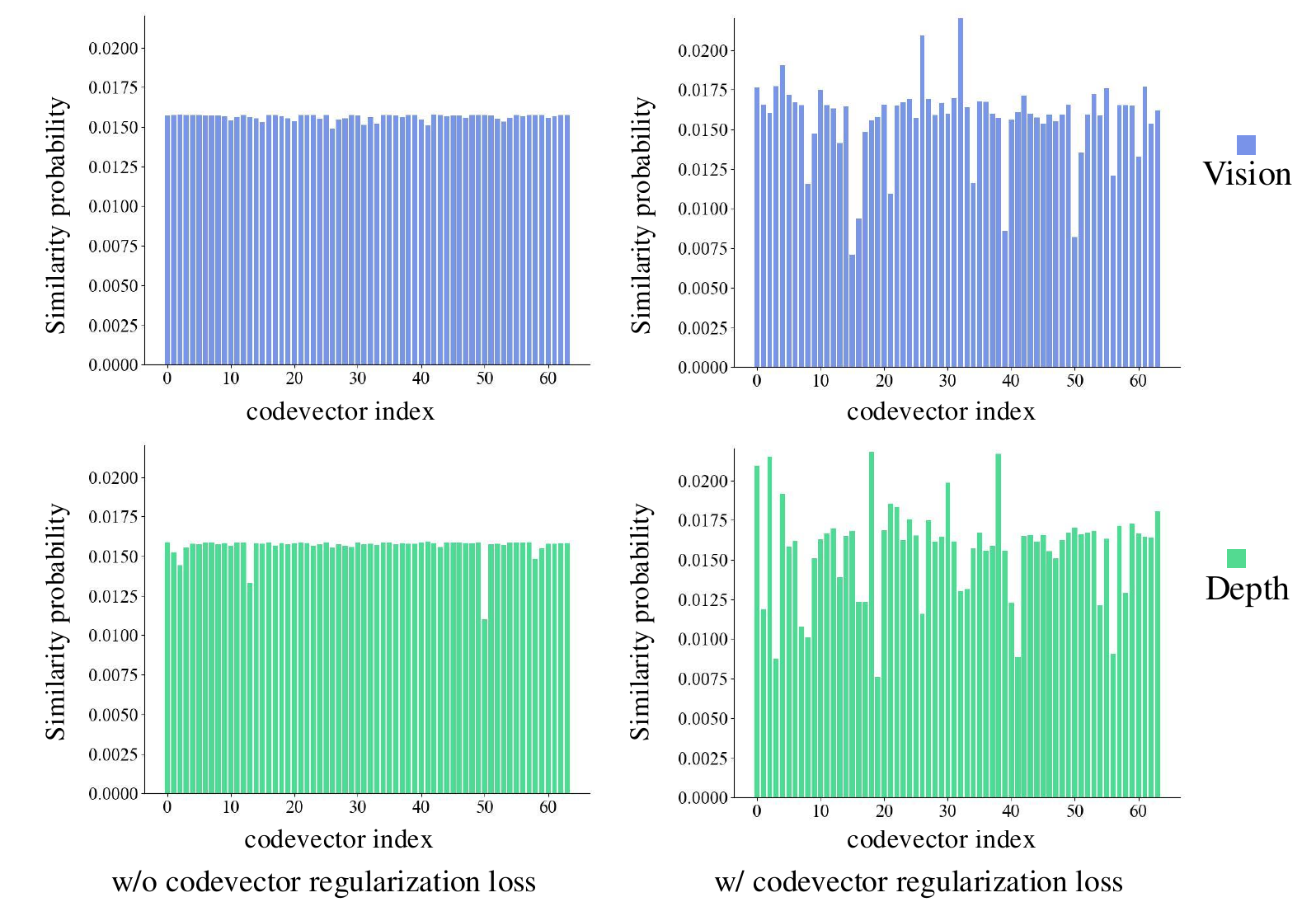}  
  \caption{codevector distribution for specific-embeddings selected in NYU-D~\cite{silberman2012nyu}}
\end{subfigure}
\caption{\textbf{Visualization of codevector similarity distribution.} The left column shows the distribution without codevector regularization loss $\mathcal{L}_{\mathrm{cctr}}$ and $\mathcal{L}_{\mathrm{cuni}}$, while the right column presents the distribution with these losses applied. The codevector regularization loss clearly encourages an uneven distribution of codevectors. This enhancement promotes better discriminativeness of codevectors, ensuring each codevector effectively represents distinct features.}
\label{fig_supp_codedist}
\end{figure*}

\paragraph{Codebook regularization loss}
CMCM loss $\mathcal{L}_{\mathrm{cm}}$, orthogonal loss and uniform loss $\mathcal{L}_{\mathrm{orth}}, \mathcal{L}_{\mathrm{uni}}$ have been discussed in \citealp{liu2022cross} and \citealp{xia2024achieving, jiang2023understanding}, respectively. 
We assess the efficacy of the proposed codevector contrastive loss $\mathcal{L}_{\mathrm{cctr}}$ and codevector uniformity loss $\mathcal{L}_{\mathrm{cuni}}$ in enhancing the distinction capability of codevectors.
To facilitate clear observation, a conventional codebook design is employed without dividing the embeddings, and the size of the shared and specific codebooks is reduced to 256 and 64, respectively. 
The cosine similarity of the specific embeddings from the sampled data in the FLIR\_v2~\citep{flirv2kaggle} and NYU-D~\citep{silberman2012nyu} datasets with all codevectors in the specific codebooks is calculated. A probability distribution of these similarities is then created using the softmax function.
Fig.~\ref{fig_supp_codedist} displays the similarity distributions for paired vision-thermal and paired vision-depth data. 
Without $\mathcal{L}_{\mathrm{cctr}}, \mathcal{L}_{\mathrm{cctr}}$, the distribution is more uniform, suggesting that a specific embedding interacts similarly with most codevectors, indicating less discriminative information.
With $\mathcal{L}_{\mathrm{cctr}}, \mathcal{L}_{\mathrm{cctr}}$, the distribution becomes significantly uneven, showing that the specific embeddings align strongly with specific codevectors while neglecting others. 
This result confirms that the codevectors within a specific codebook are inherently distinct and can offer easily distinguishable information for the specific embeddings. This distinctiveness is further boosted by the application of our codevector regularization loss.

\subsection{Additional CodeBind Applications}
\label{sec: additional downstream supp}

\begin{figure*}[t]
\centering
\includegraphics[width=1.0\linewidth]{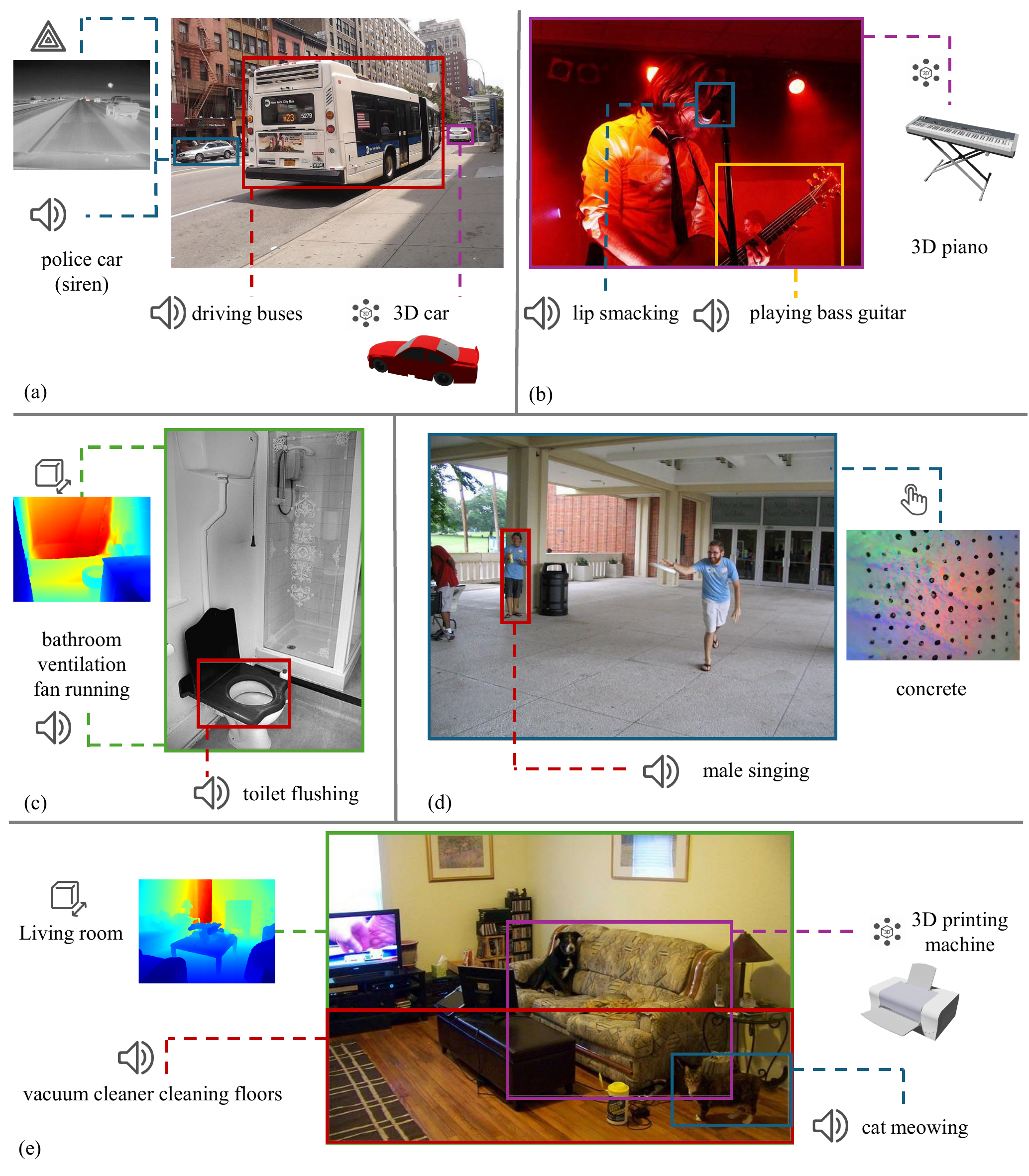}
\caption{
\textbf{Additional results for cross-modal object localization}. Semantically or geometrically related items from audio, depth, thermal, 3D point cloud, and tactile modalities can be effectively retrieved given several visual proposals in the images.
}
\label{fig_supp_app_detection}
\end{figure*}

\paragraph{Cross-modal object localization}
We use the Recognize Anything Model (RAM)~\citep{zhang2024ram} to generate object tags from selected images in MS-COCO~\citep{lin2014mscoco}, and then apply GroundingDINO~\citep{liu2024groundingdino} to obtain the corresponding object bounding boxes, which serve as our vision proposals.
Fig.~\ref{fig_supp_app_detection} displays additional retrieval results from depth, audio, thermal, 3D point cloud, and tactile modalities, based on vision proposals predicted by GroundingDINO~\citep{liu2024groundingdino}. In this process, shared embeddings of the [CLS] token from the proposal image and another modality are used to compute cosine similarity, enabling the selection of the top-1 retrieval data from the other modality for display.
These vision proposals are aligned in our CodeBind space with matching embeddings from depth, audio, and 3D point clouds, sharing similar semantic meanings.

\begin{figure}[t]
\centering
\includegraphics[width=1.0\linewidth]{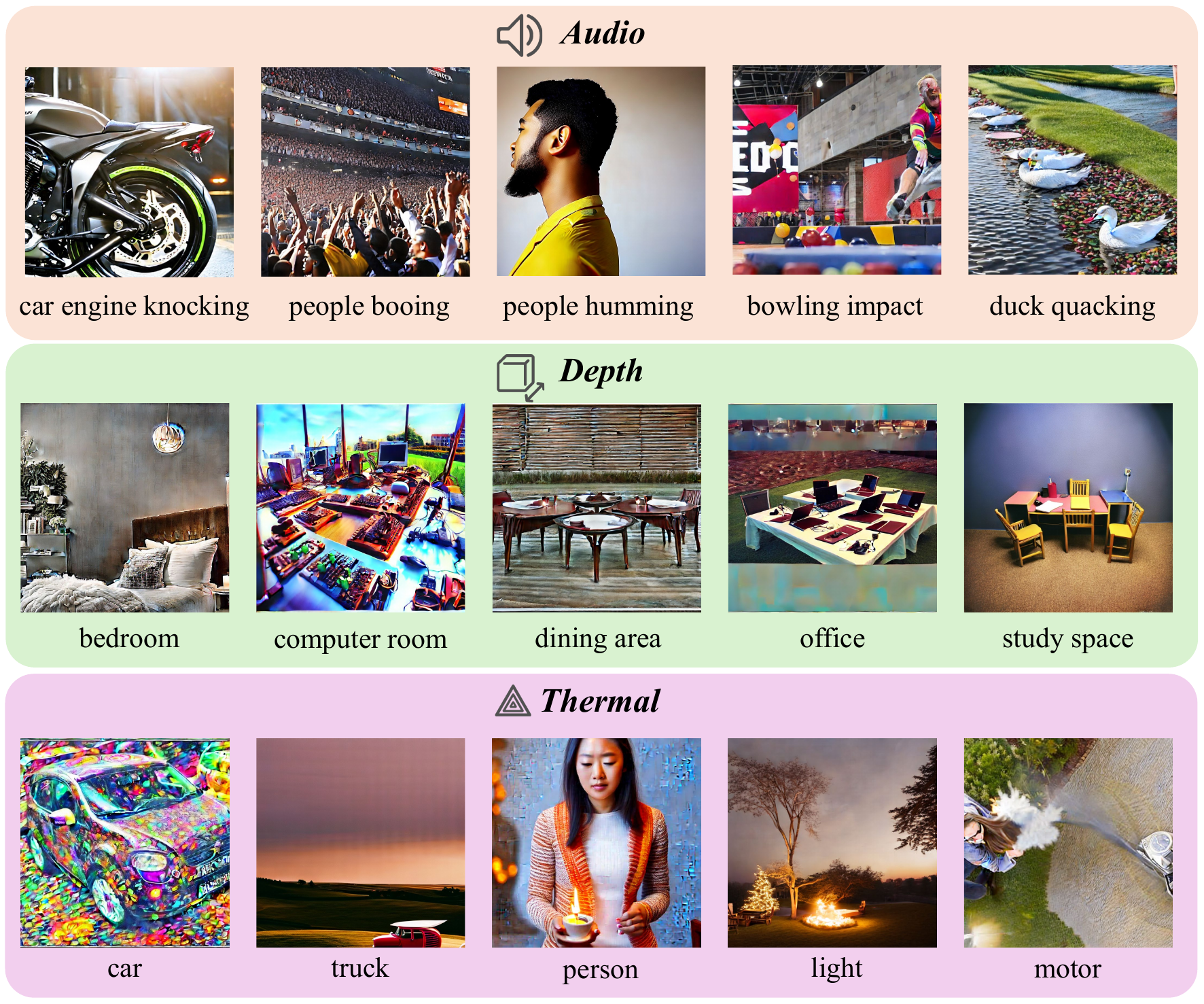}
\caption{
\textbf{Additional results for any-modal to image generation}. Semantically related images can be generated by pretrained diffusion model, using embeddings from audio, depth, and thermal modalities, which are effectively aligned with image and text embeddings through our CodeBind approach.
}
\label{fig_supp_app_generation}
\end{figure}

\newcolumntype{Y}{>{\centering\arraybackslash}X}

\begin{table}[htb]
\centering
\small

\begin{tabularx}{\columnwidth}{lYY}
\toprule
\textbf{Modality} & \textbf{Sim ($e_{\text{gen}}, e_{\text{cond}}$)} & \textbf{Sim ($e_{\text{gen}}, e_{\text{text}}$)} \\ \midrule
Audio & 0.3693 & 0.1749 \\
Depth & 0.4119 & 0.1792 \\ \bottomrule
\end{tabularx}
\vspace{-0.1cm}
\caption{
\textbf{Embedding similarity analysis} between the image embeddings of generated images ($e_{\text{gen}}$), the input modality embeddings ($e_{\text{cond}}$), and text embeddings ($e_{\text{text}}$).
}
\vspace{-0.2cm}
\label{tab_supp_anygen}
\end{table}

\paragraph{Any-modality-to-image generation}
We replace the image encoder in Stable unCLIP~\citep{rombach2022stablediffusion} with our aligned audio, depth, and thermal encoders to unlock its ability to generate semantically related images. 
Specifically, we use the shared embeddings of the [CLS] token from audio, depth, and thermal modalities. Fig.~\ref{fig_supp_app_generation} presents more generated images with their semantic categories from related modalities. To improve generation quality during inference, we apply a negative prompt to reduce unwanted artifacts: \textit{"lowres, text, error, cropped, worst quality, low quality, jpeg artifacts, ugly, duplicate, morbid, mutilated, out of frame, extra fingers, mutated hands, poorly drawn hands, poorly drawn face, mutation, deformed, blurry, dehydrated, bad anatomy, bad proportions, extra limbs, cloned face, disfigured, gross proportions, malformed limbs, missing arms, missing legs, extra arms, extra legs, fused fingers, too many fingers, long neck, username, watermark, signature"}.

Without any retraining of the diffusion model or additional text prompts, semantic concepts from other modalities can be seamlessly integrated into the generated images. Additionally, a quantitative evaluation of the any-to-image generation is conducted. Specifically, we compute the embedding similarity between the vision embeddings of generated images ($e_{\text{gen}}$) and two references: (1) the input modality embeddings (used as diffusion conditions) ($e_{\text{cond}}$), and (2) the text embeddings of their corresponding semantic categories ($e_{\text{text}}$). Based on nearly 3,000 samples across all categories, respectively from the VGGSound and SUN-D datasets, the average similarity scores are summarized in Tab~\ref{tab_supp_anygen}.

\subsection{Visualization of Reconstruction Results}
\begin{figure}[t]
\centering
\includegraphics[width=1.0\linewidth]{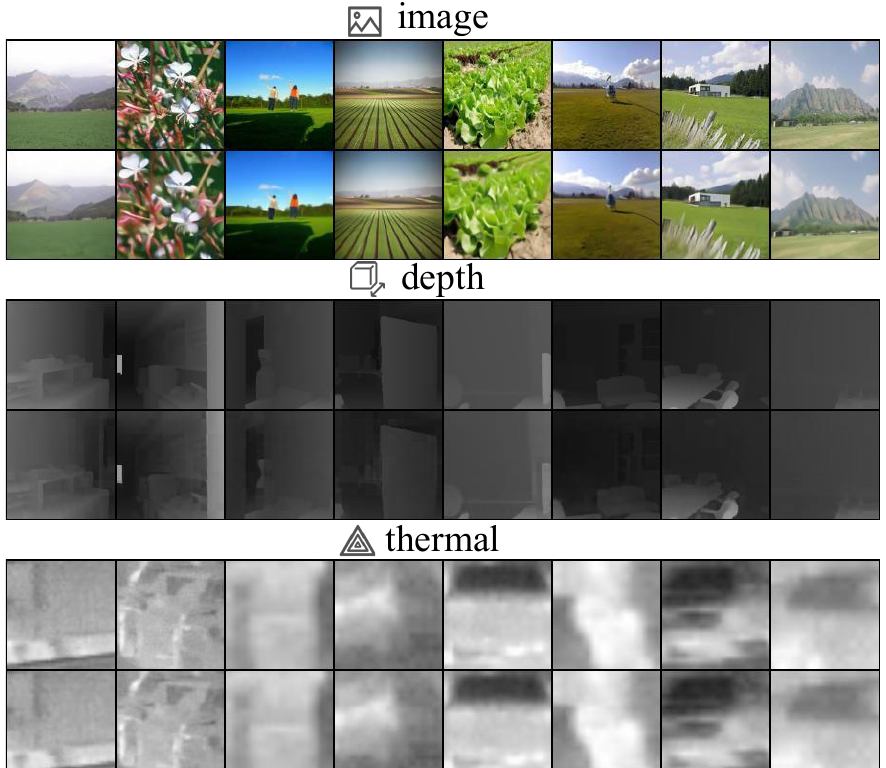}
\vspace{0.1cm} 
\caption{
\textbf{Visualization of reconstructed images, depth, and thermal images.} For each modality, the first row displays the sampled ground truth images, while the second row shows the corresponding reconstructed images.
}
\label{fig_supp_reconstruction}
\end{figure}

We present visualizations of reconstructed RGB images, depth images, and thermal images from the Place365~\citep{zhou2014learning}, NYU-D~\citep{silberman2012nyu}, and FLIR\_v2~\citep{flirv2kaggle} datasets, respectively, in Fig~\ref{fig_supp_reconstruction}. These results demonstrate that the concatenated embeddings effectively capture enough information for the decoder to accurately restore the original modality inputs.

\begin{table*}[tb]
\centering
\footnotesize
\begin{tabularx}{\textwidth}{lX}
\toprule
\textbf{Stage} & \textbf{Prompt Content} \\ \midrule
\textbf{Stage 1} & \texttt{Please provide a comprehensive structural analysis of this image. Your output MUST be a valid JSON object with the following structure:} \\
& \texttt{\{}\\
& \quad \texttt{"class\_name": "The category of the object for reference",}\\
& \quad \texttt{"semantic\_content": "Briefly identify the main objects and their categories.",}\\
& \quad \texttt{"physical\_nuances": \{}\\
& \quad \quad \texttt{"lighting\_illumination": "Describe source direction, intensity, shadows, and contrast.",}\\
& \quad \quad \texttt{"camera\_geometry": "Describe specific viewpoint, lens perspective, and distance.",}\\
& \quad \quad \texttt{"texture\_surface": "Describe visual frequency, graininess, or smoothness (strictly avoiding naming the object).",}\\
& \quad \quad \texttt{"scene\_environment": "Describe background relationship, bokeh depth, or clutter level.",}\\
& \quad \quad \texttt{"color\_tone": "Describe dominant color temperature, saturation, and chroma distribution."}\\
& \quad \texttt{\}}\\
& \texttt{\}} \\
& \texttt{STRICT CONSTRAINT: In the `physical\_nuances' section, do not use words that identify the specific species or category of the object. Focus purely on the imaging properties.} \\ \midrule

\textbf{Stage 2} & \texttt{I will provide you with a collection of physical descriptions of images from various categories. Your task is to analyze these descriptions and synthesize a ``Universal Physical Attribute Taxonomy''.} \\
& \texttt{CORE GOALS:} \\
& (1) \texttt{Disentanglement: The categories must represent `Intra-class variance' (physical imaging properties) rather than `Inter-class semantics' (object identity).} \\
& (2) \texttt{Structure: Define exactly 5 independent categories (e.g., Lighting, Camera Geometry, Scene Context, etc.).} \\
& (3) \texttt{Discreteness: For each category, provide exactly 4--5 mutually exclusive and clear options (labels).} \\
& (4) \texttt{Semantic-free: Ensure these labels do not mention specific object names like `dog' or `fur'.} \\
& \texttt{INPUT DESCRIPTIONS:} \{xxxx\} \\
& \texttt{OUTPUT FORMAT (Strict JSON):} \\
& \texttt{\{ "taxonomy": \{ "category\_name\_1": ["option1", "option2", "option3", "option4"], "category\_name\_2": ["option1", "option2", "option3", "option4"], ... \}, "logic": "Briefly explain why these 5 categories were chosen for representation disentanglement." \}} \\ \midrule

\textbf{Stage 3} & \texttt{You are a high-precision image attribute annotator. For the provided image, you must select exactly ONE option from each category below.} \\
& \texttt{STRICT CONSTRAINTS: 1. DO NOT mention the name of the object. 2. Output MUST be a valid JSON object.} \\
& \texttt{CATEGORIES AND OPTIONS: \{options\_str\}} \\
& \texttt{OUTPUT JSON FORMAT:} \\
& \texttt{\{}\\
& \quad \texttt{"Lighting\_Illumination": "selected\_option",}\\
& \quad \texttt{"Camera\_Geometry": "selected\_option",}\\
& \quad \texttt{"Texture\_Surface": "selected\_option",}\\
& \quad \texttt{"Scene\_Environment": "selected\_option",}\\
& \quad \texttt{"Color\_Tone": "selected\_option"}\\
& \texttt{\}} \\ \bottomrule
\end{tabularx}
\vspace{0.1cm}
\caption{\textbf{Multi-stage prompts for VLM-based fine-grained attribute annotation.}}
\label{tab_supp_linear_prob}
\end{table*}

\label{sec:appendix}

\end{document}